\pdfoutput=1
\documentclass{article}
\usepackage{iclr2027_conference,times}

\usepackage{amsmath,amsfonts,bm}

\def\eqref#1{equation~\ref{#1}}

\def\1{\bm{1}}

\DeclareMathAlphabet{\mathsfit}{\encodingdefault}{\sfdefault}{m}{sl}
\SetMathAlphabet{\mathsfit}{bold}{\encodingdefault}{\sfdefault}{bx}{n}

\newcommand{\Cov}{\mathrm{Cov}}

\usepackage{hyperref}
\usepackage{url}
\usepackage{amsmath}
\usepackage{amssymb}
\usepackage{amsthm}
\usepackage{bm}
\usepackage{graphicx}
\usepackage{tikz}
\usepackage{pgfplots}
\pgfplotsset{compat=1.17}
\usetikzlibrary{arrows.meta, positioning, fit, backgrounds, calc}
\usepackage{booktabs}
\usepackage{multirow}
\usepackage{tabularx}
\usepackage{xcolor}
\usepackage{algorithm}
\usepackage{algpseudocode}
\usepackage{microtype}

\newtheoremstyle{paperthm}%
  {1.0em}{1.0em}%
  {\itshape}{}%
  {\bfseries}{.}%
  { }{\thmname{#1}\thmnumber{ #2}\thmnote{ (#3)}}
\theoremstyle{paperthm}
\newtheorem{proposition}{Proposition}

\newtheorem{lemma}{Lemma}

\newtheorem{conjecture}{Conjecture}
\theoremstyle{definition}

\theoremstyle{remark}
\newtheorem{remark}{Remark}

\newcommand{\manifold}{\mathcal{M}}
\newcommand{\dM}{d_{\manifold}}
\newcommand{\dbot}{d'}
\newcommand{\Dp}{D_p}
\newcommand{\Pbud}{\mathsf{P}}
\newcommand{\ourmodel}{PoD}

\newcommand{\expt}[1]{\mathbb{E}\!\left[#1\right]}
\newcommand{\norm}[1]{\left\|#1\right\|}

\title{Polynomial Denoisers on the Image Manifold:\\
Volterra Networks for Self-Contained Pixel Diffusion}

\author{%
Siddharth Roheda, Rohit Chowdhury, Amit Satish Unde, Aniruddha Bala \\
Samsung Research Institute Bangalore, India
}

\iclrfinalcopy

\begin{document}

\maketitle
\lhead{Preprint}

\begin{abstract}

Natural images are often conjectured to lie on a low-dimensional manifold within the high-dimensional pixel space. Recent work leverages this to build self-contained pixel-space diffusion models without specialized machinery or pretrained components. A plain Transformer on raw image patches suffices, provided it predicts the clean image rather than the noise and compresses each patch through an aggressive bottleneck. The mixing operator, however, has remained softmax attention, inherited from sequence modeling rather than derived from denoising. For Gaussian noise, the optimal denoiser admits a canonical expansion in polynomial functionals of the observation, the Wiener--Volterra series. We prove that softmax attention, and any normalized mixing rule (including linear attention), cannot realize any polynomial map beyond an affine one. In contrast, Volterra layers realize this class directly and, on a data manifold, achieve approximation rates governed by the manifold dimension rather than the ambient pixel dimension. We therefore propose the architecture these results prescribe; raw image patches passed through a bottleneck embedding, followed by cascaded low-rank Volterra layers, which carry no activation function because each layer is nonlinear in its own right. To our knowledge this is the first pixel-space diffusion architecture whose mixing operators are Volterra layers throughout, with no attention, no convolution, no pretrained components, and no activation function on the residual path. On ImageNet $256 \times 256$ and $512 \times 512$, the resulting model is competitive with self-contained pixel-space diffusion at matched parameter and FLOP budgets.

\end{abstract}

\section{Introduction}
\label{sec:introduction}

Diffusion models now dominate high-fidelity image generation, and most
use the Transformer, a sequence operator whose fit to the geometry of
natural images remains unclear. The strongest systems operate in the
latent space of a pretrained tokenizer
\citep{rombach2022ldm,peebles2023dit}, which ties generation to external
components trained with adversarial and perceptual objectives and,
increasingly, aligned to self-supervised encoders
\citep{oquab2023dinov2,yu2025repa}. On the other hand, pixel-space
alternatives avoid the tokenizer but typically add specialized machinery of their own
(Section~\ref{sec:related}).
\citet{li2026jit} showed that a plain Transformer on raw image patches
suffices, provided it predicts the clean image rather than the noise or
velocity and compresses each patch through a low-dimensional bottleneck.
They motivate this with the manifold assumption, that natural images
occupy a low-dimensional subset of pixel space, which clean-image
prediction exploits. This leaves the architectural question open. If
denoising is intrinsically low-dimensional, which \emph{function class}
represents it most efficiently?

We find that the answer is classical. The minimum mean-squared-error
denoiser for a signal in additive Gaussian noise admits a canonical
expansion as a polynomial functional series, the Wiener--Volterra series
\citep{volterra1887sopra,schetzen1980volterra}. We turn this
characterization into two results for pixel-space diffusion. First,
restricted to a $\dM$-dimensional data manifold, the optimal denoiser is
approximated by a polynomial in a low-dimensional embedding, with squared
error decaying as $N^{-2s/(2\dM+1)}$ in the number of coefficients $N$ for
smoothness $s$, a rate set by the intrinsic rather than the ambient
dimension (Proposition~\ref{prop:manifold-approx}). Second, attention lies
structurally outside this class. A softmax attention layer that realizes
a polynomial map at all realizes an affine one, whereas a low-rank
Volterra stack represents the whole class exactly once its width clears
an explicit threshold (Proposition~\ref{prop:separation}).

Our architecture is the explicit construction these results call for.
Each raw patch passes through a bottleneck embedding, followed by a stack
of low-rank Volterra layers, each pairing a cross-token mixer over patch
positions with a channel-wise mixer within a patch. Because each mixer is
nonlinear in its own right, the residual path carries no activation
function, and the model trains end-to-end on pixels with no tokenizer, no
adversarial or perceptual loss, no representation alignment and no
pretrained components. To our knowledge it is the first pixel-space
diffusion architecture derived as the constructive realization of an
approximation-theoretic separation rather than by analogy with sequence
modeling.

On ImageNet $256\!\times\!256$ the resulting model outperforms JiT in FID
at every scale at matched parameters and FLOPs, reaching FID $1.48$ with no
pretrained components, and it does so again at $512\!\times\!512$, where
it reaches $1.55$ (Section~\ref{sec:exp:main}). In a controlled study on a known manifold, the necessity
of $x$-prediction reported by \citet{li2026jit} carries over to our
architecture (Section~\ref{sec:exp:toy}).

\paragraph{Contributions.}
We give an intrinsic-dimension approximation rate for the optimal
denoiser on the data manifold, and a separation showing that softmax
attention realizes no polynomial map beyond affine ones while the
Volterra class attains the rate exactly (Section~\ref{sec:theory}). From
this theory we derive a self-contained, activation-free pixel-space
diffusion backbone built from low-rank Volterra token and channel mixers
(Section~\ref{sec:architecture}). It improves FID over JiT at every scale
on ImageNet at matched compute (Section~\ref{sec:exp:main}), and an
ablation shows that each Volterra sublayer lowers FID, with the
activation-free channel mixer outperforming the feed-forward network it
replaces (Section~\ref{sec:exp:ablation}).

\section{Background and Related Work}
\label{sec:background}
\label{sec:related}

\subsection{Pixel-Space Diffusion}
\label{sec:related:pixel}
\label{sec:bg:flow}

Diffusion models started in pixel space with convolutional U-Nets
\citep{sohldickstein2015thermo,ho2020ddpm,song2019ncsn,dhariwal2021adm}.
Latent diffusion \citep{rombach2022ldm} moved them to autoencoder
latents and became standard for Transformer backbones such as DiT
\citep{peebles2023dit} and SiT \citep{ma2024sit}, at the cost of pretrained tokenizers and alignment to self-supervised
encoders \citep{oquab2023dinov2,yu2025repa}. Pixel-space diffusion remains
 competitive via specialized designs, namely latent-token
interfaces \citep{jabri2023rin}, hourglass Transformers
\citep{crowson2024hdit}, U-ViT hybrids
\citep{hoogeboom2023simplediff,hoogeboom2025sid2}, multi-scale cascades
\citep{chen2025pixelflow}, dual-level pixel Transformers
\citep{yu2025pixeldit}, aligned neural-field decoders
\citep{wang2025pixnerd}, or self-supervised pretraining
\citep{lei2026epg}. \citet{li2026jit} showed that a plain ViT on raw
patches works if it predicts the clean image and aggressively
bottlenecks each patch. We likewise use a self-contained, plain
architecture, but replace attention with Volterra layers,
motivated by approximation theory on the data manifold
(Section~\ref{sec:theory}).

\paragraph{Formulation.}
We use the flow-matching diffusion formulation \citep{lipman2023flow,esser2024scaling}. Given data $x \sim \rho_{\mathrm{data}}$ and noise $\epsilon \sim \mathcal{N}(0,I)$, we define the noised sample $z_t = t\,x + (1-t)\,\epsilon$ for $t\in[0,1]$, with velocity $v = \partial_t z_t = x-\epsilon$.
Following \citet{li2026jit}, a network $\mathrm{net}_\theta(z_t,t,c)$ (with class label $c$) predicts the clean image $\hat{x}_\theta$, and we form the velocity prediction as
\begin{equation}
v_\theta = \frac{\hat{x}_\theta - z_t}{1-t},
\label{eq:bg:vpred}
\end{equation}
with the denominator clipped near $t=1$. We train with the squared velocity loss $\mathcal{L} = \mathbb{E}_{t,x,\epsilon}\,\norm{v_\theta - v}^2$ \citep{salimans2022progressive}, sampling $t$ from a logit-normal distribution \citep{esser2024scaling}. At inference we integrate the learned velocity field from $t=0$ to $t=1$ with an ODE solver. The network output is kept in $x$-space, for the reason given in Section~\ref{sec:arch:tokenization}.

\subsection{The Manifold Assumption}
\label{sec:related:manifold}
\label{sec:bg:manifold}

The hypothesis that high-dimensional natural data lie near a low-dimensional manifold is longstanding \citep{chapelle2006ssl,carlsson2009topology}, and manifold-learning methods \citep{tenenbaum2000isomap, belkin2003laplacian} were designed to recover it. Denoising autoencoders \citep{vincent2008dae} made the link to denoising explicit, anticipating the objective of modern diffusion models.

Generative models leverage the manifold assumption differently. Latent
diffusion \citep{rombach2022ldm} learns it \emph{implicitly}
via a pretrained autoencoder, while \citet{li2026jit} keep it
\emph{ambient}, arguing that clean-image prediction uses
low-dimensional structure without an explicit encoder. We build it into
the architecture, where the bottleneck embedding and Volterra mixers
directly parametrize smooth maps on it. Jackson-type
polynomial rates on compact submanifolds are classical
\citep{ragozin1970polynomial}; our contribution is the observation that
a linear bottleneck into $\mathbb{R}^{2\dM+1}$ makes the rate depend
only on intrinsic dimension, and we provide an architecture that
attains it (Section~\ref{sec:theory}).
We work with image patches, so the ambient dimension is the per-patch pixel dimension $D := \Dp = 3p^2$ (hundreds to low thousands). We assume $\rho_{\mathrm{data}}$ is supported on a compact smooth submanifold $\manifold \subset \mathbb{R}^D$, the per-patch data manifold, of intrinsic dimension $\dM \ll D$.

Real image distributions concentrate near rather than exactly on such a manifold; we state results for exact support, which the proofs use through compactness and reach, and near-manifold data behave as this case at noise levels well above the data's thickness. That $\dM \ll D$ is an empirical and conjectural premise, supported by the success of strong bottleneck embeddings in \citep{li2026jit} and by intrinsic-dimension estimates \citep{pope2021intrinsic,levina2004maximum}.

\subsection{Volterra Series and Polynomial Architectures}
\label{sec:related:polynomial}
\label{sec:bg:volterra}

Polynomial and multiplicative interactions are longstanding alternatives to linear maps with pointwise nonlinearities, though largely justified empirically rather than by target-function structure \citep{lin2015bilinear, chrysos2020pinets}. Among attention-free token mixers, MLP-Mixer \citep{tolstikhin2021mlpmixer} mixes tokens with a learned linear map, while gMLP \citep{liu2021gmlp} uses multiplicative gating. Its spatial gating unit is a special case of our token mixer with one identity projection.

\paragraph{The discrete Volterra series.}
For $u \in \mathbb{R}^d$, the order-$K$ discrete Volterra expansion is expressed as,
\begin{equation}
y_c \;=\; \sum_{k=1}^{K} \bigg[ \;\sum_{j_1= 1}^{d} \dots \sum_{j_k= 1}^{d}
\; \big[W^{(k)}_{c,\, j_1, \dots, j_k}\, \prod_{m=1}^{k} u_{j_m}\big] \bigg],
\label{eq:bg:volterra}
\end{equation}
where $y_c$ is the $c$-th output and the kernel $W^{(k)} \in \mathbb{R}^{d_{\mathrm{out}} \times d \times \cdots \times d}$, a $(k+1)$-way tensor, weights every degree-$k$ monomial $u_{j_1}\cdots u_{j_k}$ of the input. Introduced as a functional series for smooth nonlinear systems \citep{volterra1887sopra,schetzen1980volterra}, the expansion is, by Stone--Weierstrass \citep{stone1948}, dense in continuous functions on compact domains as $K\to\infty$.

\paragraph{Volterra neural networks.}
Cascaded low-rank Volterra filtering \citep{roheda2020conquering,roheda2024vnn} first made the discrete Volterra series a competitive deep backbone, reaching state-of-the-art action recognition with far fewer parameters than comparable convolutional networks. Volterra networks have since proved competitive at low parameter budgets in classification, multimodal fusion, restoration and signal processing \citep{roheda2025vnode,ghanem2023latent,roheda2024mrvnet,bai2025wavenetvolterra}. \citet{banerjee2021volterranet} define Volterra convolutions on \emph{known} homogeneous manifolds such as the sphere, where the manifold is the signal domain; here it is the \emph{unknown} support of the data. \citet{bala2026galaxyedit} add Volterra layers to an adapter of a pretrained latent diffusion model, whereas we use them as the entire backbone, on raw pixels with no pretrained component.

\section{Theoretical Motivation}
\label{sec:theory}

\subsection{The Denoising Map on the Manifold}
\label{sec:theory:denoiser}

Throughout, $\sigma := 1 - t$ and $t \in (0,1)$, so $\sigma > 0$; notation is summarized in Table~\ref{tab:notation}. The
optimal denoiser at time $t$ is the conditional mean
$f_t(z) = \expt{x \mid z_t = z}$, which is both the minimum mean-squared-error (MMSE)
estimate of the clean signal $x$ from the noised observation $z_t = z$,
and the population minimizer of the training objective. Its second role holds because, at each $t$, the velocity loss of
Section~\ref{sec:bg:flow} equals the $x$-prediction error multiplied by
$\sigma^{-2}$, so both losses share the minimizer $f_t$
(Appendix~\ref{app:extended:targets}). Asking \emph{which architecture
approximates $f_t$ efficiently} is therefore asking which architecture is
matched to the denoising task. Because the
forward process adds Gaussian noise,
$z_t \mid x \sim \mathcal{N}(tx, \sigma^2 I)$, Bayes' rule turns this
mean into an average over the manifold, each clean point $x$ weighted by
how closely $tx$ matches $z$, and its Jacobian is the scaled posterior
covariance,
\begin{equation}
f_t(z) \;=\;
\frac{\int_{\manifold} x\,
e^{-\norm{z - tx}^2 / 2\sigma^2}\, d\rho_{\mathrm{data}}(x)}
{\int_{\manifold}
e^{-\norm{z - tx}^2 / 2\sigma^2}\, d\rho_{\mathrm{data}}(x)},
\qquad
\nabla_z f_t(z) \;=\; \frac{t}{\sigma^2}\,
\Cov_{\,x \sim p(\cdot \mid z_t = z)}(x).
\label{eq:theory:mmse}
\end{equation}
\makeatletter\edef\@currentlabel{\theequation}\makeatother\label{eq:theory:jacobian}%
So $f_t(z)$ is a soft projection of the observation onto the data
manifold. Two structural facts drive the analysis. First, because
$\sigma > 0$ the Gaussian weight is strictly positive and real-analytic,
so $f_t$ is real-analytic on all of $\mathbb{R}^D$
(Appendix~\ref{app:proof}) and its smoothness order $s$ may be taken
arbitrarily large. Second, the Jacobian of $f_t$ is
exactly the scaled posterior covariance (right
of~\eqref{eq:theory:mmse}), a denoising analogue of Tweedie's formula
\citep{robbins1956empirical,efron2011tweedie}. As the posterior is
supported on the $\dM$-dimensional manifold, this covariance is
concentrated on about $\dM$ tangent directions, so $f_t$ varies quickly
along $\manifold$ and slowly transverse to it. The first fact governs
the approximation \emph{rate}, the second its \emph{cost}.

\paragraph{Why a polynomial class.}
For a signal in Gaussian noise the MMSE estimator is classically a
series of polynomial functionals of the observation, the Wiener--Volterra
series \citep{wiener1958nonlinear,schetzen1980volterra}, and here that
expansion can be written explicitly. Write $\tilde{\rho}$ for the data distribution $\rho_{\mathrm{data}}$
reweighted by $e^{-t^2\norm{x}^2/2\sigma^2}$ and renormalized, a fixed
measure that does not depend on the observation. The denoiser is the
mean of $\tilde{\rho}$ under the exponential tilt $e^{\theta^\top x}$
with natural parameter $\theta := tz/\sigma^2$, and expanding that mean
in $\theta$ about $\theta = 0$ gives
\begin{equation}
f_t(z) \;=\; \kappa_1 \;+\; \tfrac{t}{\sigma^2}\, \kappa_2\, z
\;+\; \tfrac{1}{2}\big(\tfrac{t}{\sigma^2}\big)^{2} \kappa_3[z, z]
\;+\; \cdots,
\label{eq:theory:cumulants}
\end{equation}
a Volterra series whose order-$n$ kernel is the $(n{+}1)$-st cumulant
$\kappa_{n+1}$ of $\tilde{\rho}$, scaled by $(t/\sigma^2)^n/n!$
(Appendix~\ref{app:extended:cgf}). The series is local, converging only
for $\norm{z}$ below a radius set by $\tilde{\rho}$, so it identifies the
polynomial class, and Proposition~\ref{prop:manifold-approx} supplies
the approximation on all of $t\manifold$.

\begin{proposition}[Polynomial approximation of the denoising map]
\label{prop:manifold-approx}
Let $\manifold \subset \mathbb{R}^D$ be a compact $C^\infty$ submanifold of
intrinsic dimension $\dM$, $\rho_{\mathrm{data}}$ a probability measure
on $\manifold$, and $f_t$ the optimal denoiser at $t \in (0, 1)$.
\emph{(a) Regularity.} $f_t$ is real-analytic on $\mathbb{R}^D$.
\emph{(b) Approximation.} There is a linear projection
$\pi : \mathbb{R}^D \to \mathbb{R}^{\dbot}$, $\dbot = 2\dM + 1$, whose
restriction to the scaled manifold $t\manifold$, on which the noised
inputs concentrate, is a smooth embedding for every $t$, such that for
every $s > 0$ and degree $K$ there is a polynomial
$P_K : \mathbb{R}^{\dbot} \to \mathbb{R}^D$ with
\begin{equation*}
\sup_{z \in t\manifold} \norm{f_t(z) - P_K\!\big(\pi(z)\big)}^2
\;\leq\; C \cdot K^{-2s}.
\end{equation*}
Since $P_K$ has $N = \binom{\dbot+K}{\dbot} \asymp K^{2\dM+1}$
coefficients per output coordinate, the error is
$O(N^{-2s/(2\dM+1)})$, set by $\dM$ rather than by $D$, whereas a
polynomial on the raw patch would need $\asymp K^{D}$. The constant $C = C(s,t)$ depends on $t$, the
reach and curvature of $t\manifold$ and $\norm{f_t}_{C^s}$, but not on
$D$.
\end{proposition}

The rate depends on the intrinsic dimension $\dM$ and, by part~(a), is unbounded in the smoothness $s$. On $t\manifold$ the four targets are affine reparametrizations of one another, so the rate holds for $\epsilon$-, $v$- and score prediction too; off $t\manifold$ only $x$ survives the bottleneck (Appendix~\ref{app:extended:targets}). Proposition~\ref{prop:manifold-approx} studies $f_t$ on the scaled manifold $t\manifold$, but the network is evaluated at the noised input $z_t$ in a radius-$\delta$ tube $\mathcal{T}_\delta$ around it. A compressing linear map $\pi$ discards directions normal to $t\manifold$, and the Jacobian identity (right of~\eqref{eq:theory:mmse}) shows these directions contribute little. In a unit direction $u$, the sensitivity of $f_t$ is $\tfrac{t}{\sigma^2}\, u^\top \Cov(x)\, u$ (the posterior variance along $u$), which is small for normal directions once the noise falls below the manifold reach $\tau$. Lemma~\ref{lem:concentration} (Appendix~\ref{app:lemma}) formalizes this and bounds the extra error from replacing the nonlinear retraction onto $t\manifold$ with the linear bottleneck $\pi$.

Combined with Proposition~\ref{prop:manifold-approx}, the lemma gives the
bound on $\mathcal{T}_\delta$ that the network needs,
\begin{equation}
\sup_{z \in \mathcal{T}_\delta}
\norm{f_t(z) - P_K\!\big(\pi(z)\big)}^2
\;\leq\;
\underbrace{C \cdot K^{-2s}}_{\text{polynomial approximation}}
\;+\;
\underbrace{C' \cdot \big(\eta(t, \sigma, \tau) + c_\pi/t\big)^2\, \delta^2}_{\text{compression cost}}.
\label{eq:theory:combined}
\end{equation}
The first term is the rate of Proposition~\ref{prop:manifold-approx}.
The second, the price of compression, has two sources. The factor
$\eta(t,\sigma,\tau) = C_\eta \min\{1/t,\ \sigma^2/(t^3\tau^2)\}$ bounds
how much $f_t$ changes along the normal directions that $\pi$ discards,
and is small once $\sigma \ll t\tau$. The constant $c_\pi$ covers the
tangential component in the kernel of $\pi$ and the error of evaluating
$P_K$ slightly off the manifold, and it shrinks as the bottleneck widens,
one reason the trained $\dbot$ exceeds $2\dM + 1$. Neither depends on
$\sigma$, $\delta$, or $D$. The normal component of the noise has length
about $\sigma\sqrt{D-\dM}$, so in the worst case the compression term is
$O((D-\dM)\,\sigma^2/t^2)$, which vanishes late in sampling but is not
free of $D$. Averaged over the noise, its slippage part depends on
$\dbot$ rather than $D$ (Remark~\ref{rem:typical-slippage}). All proofs
are in Appendix~\ref{app:proof}.
\paragraph{Behaviour across the sampling schedule.}
These results hold at a fixed $t$, and
Appendix~\ref{app:extended:schedule} checks that the bound stays
controlled as sampling runs $t$ from $0$ to $1$. Early in sampling $f_t$
is nearly affine, and late in sampling the posterior is concentrated and
the compression cost vanishes. The hard case is intermediate $t$, where
the posterior spreads over a curved patch whose tangent spaces together
span more than $\dM$ directions. The bottleneck must cover that span, a
second reason $\dbot$ exceeds $2\dM + 1$. The required degree is also
highest there but finite, and depth supplies it. None of this involves
$D$.

\begin{proposition}[Attainment, and the polynomial content of attention]
\label{prop:separation}
With the setting of Proposition~\ref{prop:manifold-approx}, suppose
$\rho_{\mathrm{data}}$ is not a point mass and fix $L \geq 1$. Write
$K_L := 2^L$ and $\mathcal{P}_{K}$ for the polynomial maps
$\mathbb{R}^{\dbot} \to \mathbb{R}^{D}$ of degree at most $K$.
\emph{(i) Attainment.} Let $\mathcal{V}_{L, \Pbud}$ be the depth-$L$
stacks of the low-rank Volterra mixers of Section~\ref{sec:architecture}
(degree $2$ per layer, gated residual, affine embedding, linear head, no
normalization) applied to $\pi(z)$, with at most $\Pbud$ parameters per
layer. For
$\Pbud \geq \Pbud_0(L, \dbot) \asymp \binom{\dbot + K_L}{\dbot}^{2}$,
\begin{equation*}
\big\{\, Q \circ \pi \;:\; Q \in \mathcal{P}_{K_L} \,\big\}
\;\subseteq\; \mathcal{V}_{L, \Pbud},
\qquad\text{hence}\qquad
\inf_{g \in \mathcal{V}_{L, \Pbud}} \sup_{z \in t\manifold}
\norm{f_t(z) - g(z)}^2 \;\leq\; C \cdot K_L^{-2s},
\end{equation*}
independently of $D$, and on $\mathcal{T}_\delta$ up to the compression
term of~\eqref{eq:theory:combined}.
\emph{(ii) Attention.} A softmax self-attention layer with residual
connection (any heads and projections) that agrees with a polynomial map
on a nonempty open set is affine, and $f_t$ is not affine. So the class
$\{Q \circ \pi : Q \in \mathcal{P}_{K_L}\}$, which $\mathcal{V}_{L,\Pbud}$
represents exactly, meets the attention class only in its affine part,
at every budget.
\end{proposition}

Part (i) shows that the low-rank factorization loses nothing. Above the
threshold the class equals $\mathcal{P}_{K_L}$, since it contains every polynomial of degree
$K_L$ and a depth-$L$ stack of degree-$2$ layers has no higher degree.
The proof (Appendix~\ref{app:proof-attainment}) is constructive, since
iterated squaring raises an affine form to the power $2^L$ in $L$ layers
and such powers span the polynomials. The threshold $\Pbud_0$ certifies
representability and is not a practical width. For part (ii)
(Appendix~\ref{app:proof-attention}), the infinite Taylor series of the
softmax is not extra capacity. An analytic function whose power series
does not terminate agrees with no polynomial, which puts attention
outside the class rather than above it. That attention also attains a
slower rate at matched budget is natural to expect, and we state it as
Conjecture~\ref{conj:rate-gap}.

\subsection{Beyond Softmax}
\label{sec:theory:normalized}

Proposition~\ref{prop:separation}(ii) is specific to the softmax, but the
separation is not. Proposition~\ref{prop:normalized-attention}
(Appendix~\ref{app:proof-normalized}) extends its global form to any
mixing layer whose weights are nonnegative and sum to one, however
computed. Such a layer averages values that are linear in the input,
so its output grows at most linearly, and a polynomial on the whole input
space that does so is affine; for softmax, analyticity makes an open set
enough. The result covers linear attention with a nonnegative
feature map \citep{katharopoulos2020transformers,xie2024sana}, kernel
approximations of the softmax \citep{choromanski2021rethinking}, and any
other rule that averages. Appendix~\ref{app:extended:boundary} treats
unnormalized attention, where it fails.

\paragraph{Where the nonlinearity comes from.}
A Transformer block splits the work between two components. The
attention sublayer mixes tokens, but its polynomial content is only
affine (Propositions~\ref{prop:separation}(ii)
and~\ref{prop:normalized-attention}). The feed-forward sublayer supplies
the nonlinearity but acts on each token separately. Applied to the
attention output, its activation does produce cross-token polynomial
terms, but not a free parametrization of them, since every slot draws on
the same attention weights, the activation fixes the coefficients across
orders, and a wider sublayer buys channel rank but no new token
structure (Table~\ref{tab:kernels}, Appendix~\ref{app:ffn}). The
Volterra token mixer mixes tokens and raises the degree in the same step
by multiplying two tokens, and the channel mixer does the same within a
token. The claim is therefore not that a Transformer cannot approximate
$f_t$, which it can, but that its mixing operator contributes no
polynomial content beyond affine, so the polynomial structure of the
target must be recovered through a pointwise nonlinearity whose orders
are coupled rather than free.

\section{Polynomial Denoisers on the Image Manifold}
\label{sec:architecture}

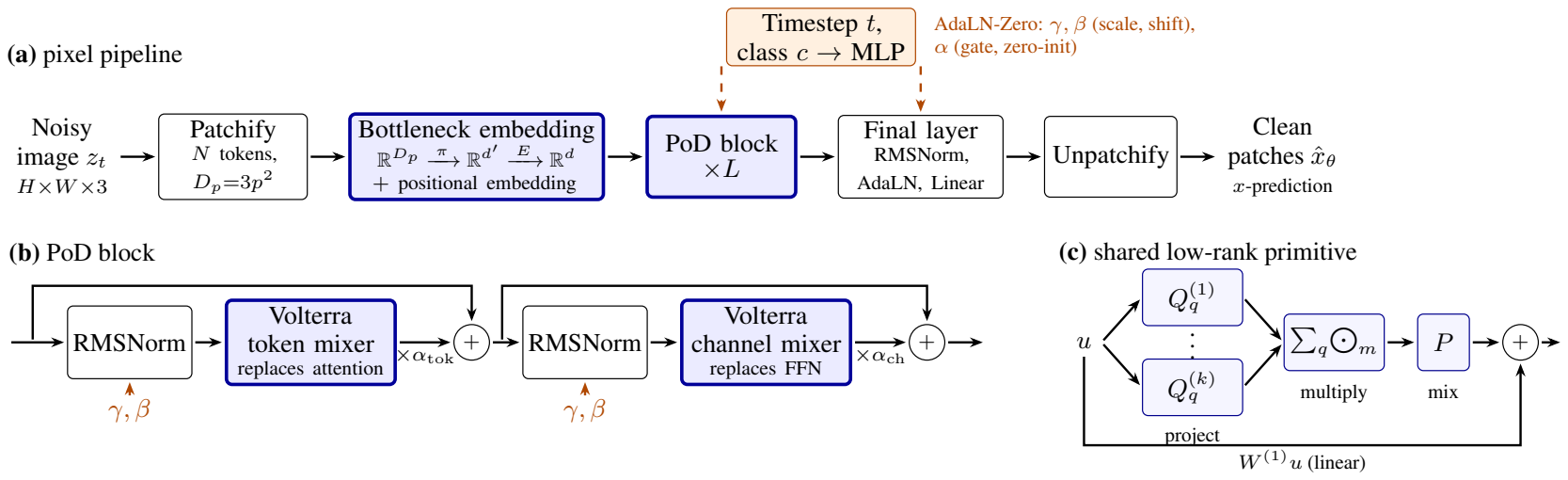
\begin{figure}[t]
\centering
\resizebox{0.8\textwidth}{!}{%
\begin{tikzpicture}[
  font=\small,
  /utils/exec={\definecolor{condc}{RGB}{176,72,0}},
  >={Stealth[length=1.6mm]},
  box/.style={draw, rounded corners=2pt, align=center, inner sep=2pt,
              minimum height=10mm, text width=19mm},
  emph/.style={box, fill=blue!8, draw=blue!60!black, very thick},
  cond/.style={draw, rounded corners=2pt, align=center, inner sep=2pt,
               minimum height=7mm, text width=22mm, fill=orange!12,
               draw=condc},
  sbox/.style={draw, rounded corners=2pt, align=center, inner sep=2pt,
              minimum height=9mm, text width=14mm, font=\footnotesize},
  semph/.style={sbox, fill=blue!8, draw=blue!60!black, very thick, text width=19.5mm},
  prim/.style={draw, rounded corners=2pt, align=center, inner sep=2pt,
               minimum height=7mm, text width=11mm, fill=blue!4,
               draw=blue!55!black, font=\footnotesize},
  op/.style={draw, circle, inner sep=0pt, minimum size=4.4mm, font=\scriptsize},
  fl/.style={->, thick, shorten >=1pt, shorten <=1pt},
  zoom/.style={dashed, gray!50},
  ada/.style={->, condc, thick, dashed, shorten >=1pt},
]
\node[align=center, font=\footnotesize] (img) {Noisy\\image $z_t$\\{\scriptsize $H{\times}W{\times}3$}};
\node[box, right=5mm of img, text width=17mm] (patch) {Patchify\\[-2pt]{\scriptsize $N$ tokens, $\Dp{=}3p^2$}};
\node[emph, right=5mm of patch, text width=30mm] (embed)
  {Bottleneck embedding\\[-1pt]{\scriptsize $\mathbb{R}^{\Dp}\!\xrightarrow{\,\pi\,}\!\mathbb{R}^{\dbot}\!\xrightarrow{\,E\,}\!\mathbb{R}^{d}$}\\[-2pt]{\scriptsize $+$ positional embedding}};
\node[emph, right=5mm of embed, text width=17mm] (blocks) {\ourmodel{} block\\[-1pt]{\footnotesize $\times L$}};
\node[box, right=5mm of blocks] (head) {Final layer\\[-2pt]{\scriptsize RMSNorm, AdaLN, Linear}};
\node[box, right=5mm of head, text width=15mm] (unpatch) {Unpatchify};
\node[align=center, font=\footnotesize, right=5mm of unpatch] (out) {Clean\\patches $\hat{x}_\theta$\\{\scriptsize $x$-prediction}};
\draw[fl] (img)--(patch); \draw[fl] (patch)--(embed); \draw[fl] (embed)--(blocks);
\draw[fl] (blocks)--(head); \draw[fl] (head)--(unpatch); \draw[fl] (unpatch)--(out);
\node[cond, above=6mm of $(blocks.north)!0.5!(head.north)$] (cmlp) {Timestep $t$, class $c$ $\to$ MLP};
\node[anchor=west, text=condc, font=\scriptsize, align=left] at ($(cmlp.east)+(1mm,0)$)
  {AdaLN-Zero: $\gamma,\beta$ (scale, shift),\\$\alpha$ (gate, zero-init)};
\draw[ada] (cmlp.south -| blocks.north) -- (blocks.north);
\draw[ada] (cmlp.south -| head.north) -- (head.north);
\node[anchor=south west, font=\footnotesize, inner sep=0pt] at ($(img.north west)+(0,5mm)$) {\textbf{(a)}\ pixel pipeline};
\coordinate (brow) at ($(img.south west)+(0,-17mm)$);
\node[anchor=west, inner sep=0pt] (bin) at (brow) {};
\node[sbox, right=7mm of bin] (n1) {RMSNorm};
\node[semph, right=4mm of n1] (tok) {Volterra token mixer\\[-2pt]{\scriptsize replaces attention}};
\node[op, right=7mm of tok] (g1) {$+$};
\node[sbox, right=4mm of g1] (n2) {RMSNorm};
\node[semph, right=4mm of n2] (ch) {Volterra channel mixer\\[-2pt]{\scriptsize replaces FFN}};
\node[op, right=7mm of ch] (g2) {$+$};
\coordinate (bout) at ($(g2.east)+(5mm,0)$);
\draw[fl] (bin.center)--(n1); \draw[fl] (n1)--(tok);
\draw[fl] (tok)-- node[below,font=\scriptsize,inner sep=1.5pt]{$\times\alpha_{\mathrm{tok}}$} (g1);
\draw[fl] (g1)--(n2); \draw[fl] (n2)--(ch);
\draw[fl] (ch)-- node[below,font=\scriptsize,inner sep=1.5pt]{$\times\alpha_{\mathrm{ch}}$} (g2);
\draw[fl] (g2)--(bout);
\coordinate (r1) at ($(bin.center)+(3mm,0)$);
\draw[fl] (r1) -- ++(0,7mm) -| (g1.north);
\coordinate (r2) at ($(g1.east)+(1.5mm,0)$);
\draw[fl] (r2) -- ++(0,7mm) -| (g2.north);
\node[text=condc, font=\footnotesize, below=2mm of n1, inner sep=1pt] (m1) {$\gamma,\beta$};
\node[text=condc, font=\footnotesize, below=2mm of n2, inner sep=1pt] (m2) {$\gamma,\beta$};
\draw[ada] (m1.north) -- (n1.south); \draw[ada] (m2.north) -- (n2.south);
\node[anchor=south west, font=\footnotesize, inner sep=0pt] at ($(bin)+(0,9.5mm)$) {\textbf{(b)}\ \ourmodel{} block};
\node[anchor=west, font=\small] (pu) at ($(bout)+(10mm,0)$) {$u$};
\node[prim, above right=2mm and 5mm of pu.east] (pq1) {$Q^{(1)}_q$};
\node[prim, below right=2mm and 5mm of pu.east] (pqk) {$Q^{(k)}_q$};
\node[font=\small, inner sep=0pt] at ($(pq1.south)!0.5!(pqk.north)+(0,0.6mm)$) {$\vdots$};
\coordinate (pmid) at ($(pq1.east)!0.5!(pqk.east)$);
\node[prim, right=5mm of pmid, text width=11mm] (pp) {$\sum_{q}\!\bigodot_{m}$};
\node[prim, right=4mm of pp, text width=5mm] (pm) {$P$};
\node[op, right=4mm of pm] (psum) {$+$};
\node[anchor=west, font=\small, right=3mm of psum] (py) {$y$};
\draw[fl] (pu.east)-- (pq1.west); \draw[fl] (pu.east)-- (pqk.west);
\draw[fl] (pq1.east)-- (pp.west); \draw[fl] (pqk.east)-- (pp.west);
\draw[fl] (pp)--(pm); \draw[fl] (pm)--(psum); \draw[fl] (psum)--(py);
\coordinate (lin) at ($(pqk.south)+(0,-3.6mm)$);
\draw[fl] ($(pu.south)+(0,0.5mm)$) |- (lin -| psum) node[pos=0.75,below,font=\scriptsize,inner sep=1pt]{$W^{(1)}u$ (linear)} -- (psum.south);
\node[font=\scriptsize, below=0.2mm of pqk.south] {project};
\node[font=\scriptsize, below=0.4mm of pp.south] {multiply};
\node[font=\scriptsize, below=0.4mm of pm.south] {mix};
\node[anchor=south west, font=\footnotesize, inner sep=0pt] at ($(pu.west)+(-1mm,9.5mm)$) {\textbf{(c)}\ shared low-rank primitive};
\end{tikzpicture}%
}%
\vspace{-11pt}
\caption{\textbf{The \ourmodel{} architecture.} (a) pixel pipeline; (b) block,
both Transformer sublayers replaced by Volterra mixers; (c) the
primitive they share,~\eqref{eq:arch:eval}.}
\label{fig:architecture}
\end{figure}

The architecture realizes the construction of
Proposition~\ref{prop:manifold-approx}. Its projection $\pi$ is a
bottleneck embedding (Section~\ref{sec:arch:tokenization}), and the
degree-$K$ polynomial $P_K$ is a stack of low-rank Volterra layers
alternating a cross-token and a channel-wise mixer
(Section~\ref{sec:arch:mixers}), as summarized in
Figure~\ref{fig:architecture}.

\subsection{Patch Tokenization and Manifold Embedding}
\label{sec:arch:tokenization}

Given an image of shape $H \times W \times 3$, we partition it into
non-overlapping $p \times p$ patches, giving $N = HW/p^2$ tokens of
ambient dimension $\Dp = 3p^2$ ($N = 256$ in all our models).

The patch embedding is a low-rank linear map factored through a
bottleneck,
$\mathbb{R}^{\Dp} \xrightarrow{\pi} \mathbb{R}^{\dbot}
\xrightarrow{E} \mathbb{R}^{d}$, where $\dbot \ll \Dp$ is the bottleneck
dimension and $d \ge \dbot$ the hidden width of the mixers that follow.
Sinusoidal positional embeddings \citep{vaswani2017attention} are added
after the embedding. Such bottlenecks improve sample quality in
pixel-space diffusion \citep{li2026jit}, and
Proposition~\ref{prop:manifold-approx} offers a theoretical
explanation.

The bottleneck is also why the network predicts the clean image. The four
prediction targets share one function class, but only the clean image
lies on the manifold, where the compressed representation can produce
it, whereas a noise or velocity output must span the ambient space
(Appendix~\ref{app:extended:targets}).

\subsection{The Low-Rank Volterra Primitive}
\label{sec:arch:primitive}

Both mixers in this section are instances of a single low-rank Volterra
primitive (Figure~\ref{fig:architecture}c).
Rather than storing a dense order-$k$ kernel, the primitive performs a
\emph{project--multiply--mix} operation. Write $n$ for the size of the
axis being mixed, $n = N$ for the cross-token mixer and $n = d$ for the
channel mixer. The input $u \in \mathbb{R}^{n}$ is projected to a hidden
width $h$ by $R$ groups of $k$ independent factors
$Q^{(1)}_q, \dots, Q^{(k)}_q \in \mathbb{R}^{h \times n}$, the $k$
projections of each group are multiplied elementwise, and the $R$
products are summed and mixed back through a factor
$P \in \mathbb{R}^{d_{\text{out}} \times h}$,
\begin{equation}
y = W^{(1)} u + P \sum_{q=1}^{R} \bigodot_{m=1}^{k} \big(Q^{(m)}_q u\big),
\qquad
W^{(k)}_{i, j_1, \dots, j_k} = \sum_{c=1}^{h} \sum_{q=1}^{R} P_{ic}
\prod_{m=1}^{k} Q^{(m)}_{q,\, c j_m}.
\label{eq:arch:eval}
\end{equation}
\makeatletter\edef\@currentlabel{\theequation}\makeatother\label{eq:arch:lowrank}%
Each coordinate $(Q^{(m)}_q u)_c$ is a learned mixture of \emph{all}
inputs, so the product produces every degree-$k$ monomial, cross-terms
included. The product term is therefore the order-$k$ Volterra
term~\eqref{eq:bg:volterra}, and $W^{(k)}$ becomes its order-$k$ kernel,
a canonical-polyadic factorization of rank at most $hR$ with an
independent (untied) factor per slot, which constrains the kernel but not
the set of interactions it represents (Appendix~\ref{app:arch}).
Evaluation costs $\mathcal{O}(k\, nhR)$ per kernel against
$\mathcal{O}(n^{k+1})$ for a dense one. We use $k = 2$ throughout, since
depth already supplies degree $2^L$ (Proposition~\ref{prop:separation}(i)).

\subsection{Cross-Token and Channel-Wise Mixing}
\label{sec:arch:mixers}
\begin{figure}[t]
\centering
\begin{tikzpicture}[font=\scriptsize, inner sep=1pt]
\definecolor{tyred}{RGB}{212,78,94}
\definecolor{tyorange}{RGB}{235,164,64}
\definecolor{tyblue}{RGB}{44,100,148}
\node[anchor=south west, inner sep=0] at (0,0)
  {\includegraphics[width=214pt]{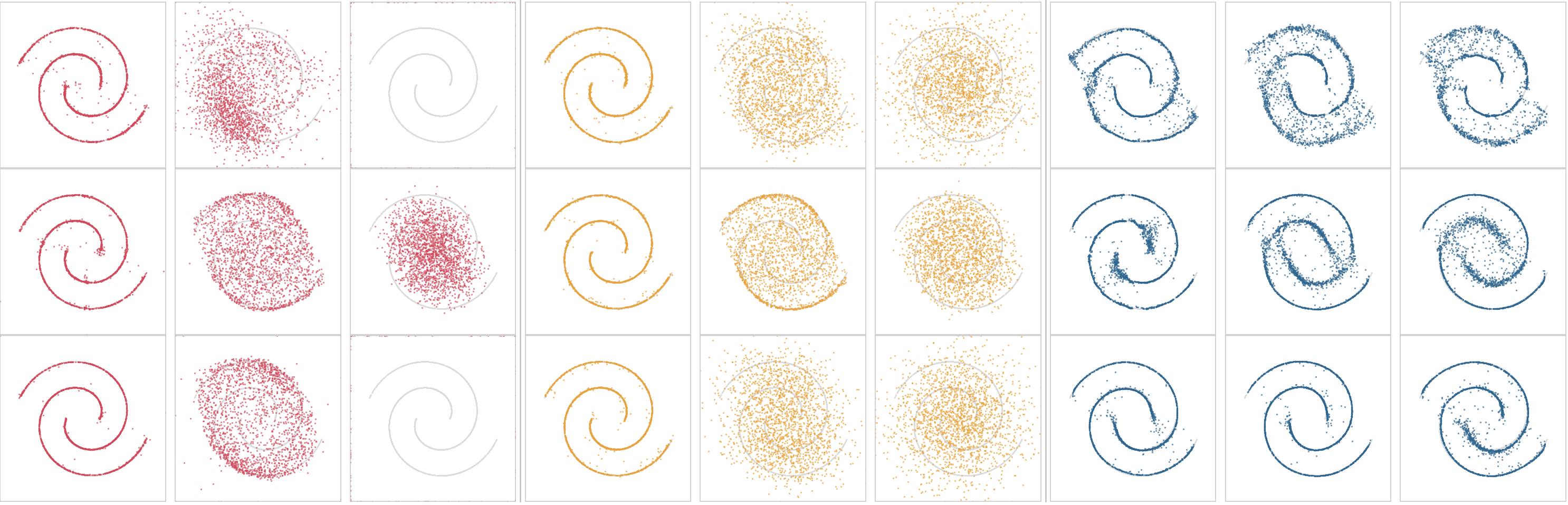}};
\begin{scope}[x={(214pt,0)}, y={(0,68.43pt)}]
  \node[text=tyred,    anchor=south] at (0.1647,1.15) {$\epsilon$-prediction};
  \node[text=tyorange, anchor=south] at (0.4981,1.15) {$v$-prediction};
  \node[text=tyblue,   anchor=south] at (0.8314,1.15) {$x$-prediction};
  \foreach \c/\d in {0.0533/2,0.1647/32,0.2752/512,0.3866/2,0.4981/32,
                     0.6085/512,0.7200/2,0.8314/32,0.9428/512}
    \node[anchor=south] at (\c,1.005) {$\d$};
  \node[anchor=south east] at (-0.012,1.005) {$D$};
  \node[anchor=east, align=right] at (-0.012,0.8318) {MLP};
  \node[anchor=east, align=right] at (-0.012,0.5015) {Softmax attention\\+ FFN};
  \node[anchor=east, align=right] at (-0.012,0.1712) {Volterra token\\+ channel mixer};
\end{scope}
\node[font=\small\bfseries, anchor=south west] at (-54pt,80pt) {(a)};
\begin{scope}[shift={(247pt,0)}]
  \node[anchor=south west, inner sep=0] at (0,0)
    {\includegraphics[width=87.7pt]{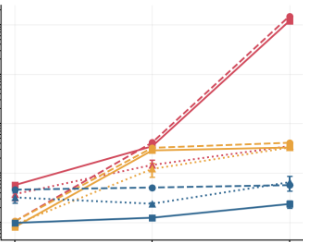}};
  \begin{scope}[x={(87.7pt,0)}, y={(0,68.46pt)}]
    \foreach \y/\e in {0.9008/2,0.6942/1,0.4917/0,0.2872/-1,0.0826/-2} {
      \draw (0,\y) -- (-0.035,\y);
      \node[anchor=east] at (-0.04,\y) {$10^{\e}$};
    }
    \foreach \x/\d in {0.0468/2,0.4887/32,0.9323/512} {
      \draw (\x,0.012) -- (\x,-0.035);
      \node[anchor=north] at (\x,-0.04) {$\d$};
    }
    \node[anchor=north] at (0.5,-0.17) {ambient dimension $D$};
    \node[rotate=90, anchor=south] at (-0.27,0.5) {$d(\manifold)$};
  \end{scope}
  \node[font=\small\bfseries, anchor=south west] at (-30pt,80pt) {(b)};
\end{scope}
\node[anchor=north] at (80pt,-2.74pt) {%
  \tikz[baseline=-0.6ex]\draw[tyred, line width=0.9pt] (0,0) -- (12pt,0);~$\epsilon$-pred\hspace{7pt}%
  \tikz[baseline=-0.6ex]\draw[tyorange, line width=0.9pt] (0,0) -- (12pt,0);~$v$-pred\hspace{7pt}%
  \tikz[baseline=-0.6ex]\draw[tyblue, line width=0.9pt] (0,0) -- (12pt,0);~$x$-pred};
\node[anchor=north] at (80pt,-11.64pt) {%
  \tikz[baseline=-0.6ex]{\draw[black!75, line width=0.9pt, dash pattern=on 2.4pt off 1.4pt] (0,0) -- (14pt,0);
                         \fill[black!75] (7pt,0) circle (1.5pt);}~MLP\hspace{7pt}%
  \tikz[baseline=-0.6ex]{\draw[black!75, line width=0.9pt, dash pattern=on 0.8pt off 1.2pt] (0,0) -- (14pt,0);
                         \fill[black!75] (7pt,1.7pt) -- (5.2pt,-1.2pt) -- (8.8pt,-1.2pt) -- cycle;}~Softmax attention + FFN\hspace{7pt}%
  \tikz[baseline=-0.6ex]{\draw[black!75, line width=0.9pt] (0,0) -- (14pt,0);
                         \fill[black!75] (5.5pt,-1.4pt) rectangle (8.5pt,1.4pt);}~Volterra token + channel mixer};
\end{tikzpicture}
\caption{\textbf{The manifold picture in a controlled setting.} (a)~Samples
from a $\dM=1$ spiral in $\mathbb{R}^D$ (grey) for three
targets and three architectures. (b)~Mean distance to the manifold
$d(\manifold)$ against $D$.}
\label{fig:toy}
\end{figure}
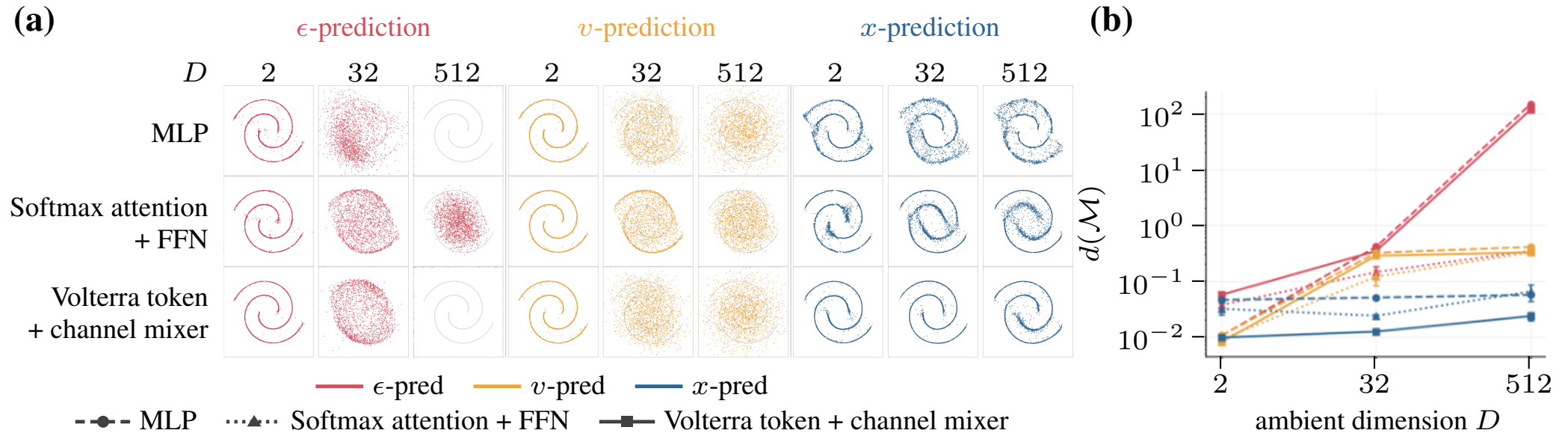

The block instantiates the primitive twice, once along each axis. Write
$u_{ia}$ for channel $a$ of token $i$, with $i = 1, \dots, N$ and
$a = 1, \dots, d$. The token mixer takes the $N$ positions as slots at
fixed channel, and the channel mixer takes the $d$ channels of one token
as slots at fixed position,
\begin{equation}
y_{ia} = \sum_{j} W^{(1)}_{\mathrm{tok},ij}\, u_{ja}
 + \sum_{j,j'} W^{(2)}_{\mathrm{tok},ijj'}\, u_{ja} u_{j'a},
\qquad
y_{ia} = \sum_{b} W^{(1)}_{\mathrm{ch},ab}\, u_{ib}
 + \sum_{b,c} W^{(2)}_{\mathrm{ch},abc}\, u_{ib} u_{ic}.
\label{eq:token-mixer}
\end{equation}
\makeatletter\edef\@currentlabel{\theequation}\makeatother\label{eq:channel-mixer}%
Each mixer shares its kernels along the axis it does not mix
($W_{\mathrm{tok}}$ carries no channel index, $W_{\mathrm{ch}}$ no
position index), and each is an instance of~\eqref{eq:arch:eval}
(Appendix~\ref{app:arch}). The token mixer has $h = N$, $P = I$ and
dense $Q^{(m)}_q \in \mathbb{R}^{N \times N}$, so it holds
$(2R+1)N^2$ parameters per block for any width $d$, against $4d^2$ for
self-attention, and costs $\mathcal{O}(RN^2 d)$ against
$\mathcal{O}(N^2 d + N d^2)$. The channel mixer has hidden width $h$ and
rank $R'$, and costs $\mathcal{O}(dhR')$ per token.

\subsection{Block Structure and Conditioning}
\label{sec:arch:block}

A polynomial denoiser block applies, in order: RMSNorm
\citep{zhang2019rmsnorm} with affine modulation, cross-token polynomial
mixer, residual connection, RMSNorm with modulation, channel-wise
polynomial mixer, residual connection
(Figure~\ref{fig:architecture}b). The pre-norm residual structure
follows standard practice for stably training deep residual networks.

Diffusion time $t$ and class label $c$ enter through unmodified
AdaLN-Zero conditioning \citep{peebles2023dit}. A small MLP of $(t, c)$
produces, for each sublayer, a scale $\gamma$ and shift $\beta$ applied
as $\gamma \odot \text{RMSNorm}(\cdot) + \beta$, and a zero-initialized
gate $\alpha$ on the sublayer output, so that each block starts as the
identity.

\paragraph{No activation function.}
The residual path contains no activation. The only nonlinearity is the
degree-$2$ product, which takes the place of the
expansion--activation--contraction of a feed-forward block, and
GELU-type functions appear only in the conditioning MLP. This is
required rather than stylistic, since a pointwise activation on the
residual path would leave the polynomial class and void
Proposition~\ref{prop:separation}(i). In the ablation of
Section~\ref{sec:exp:ablation}, replacing the feed-forward channel
sublayer with the activation-free Volterra mixer lowers FID.

The RMSNorm and the conditioning branch remain outside the polynomial
class, but both are shared with the attention baseline and neither mixes
tokens. At fixed $(t, c)$ the conditioning acts as a diagonal affine map,
and RMSNorm is a positive per-token rescaling, so
Proposition~\ref{prop:separation}(i) holds exactly for the
normalization-free class and the rate of
Proposition~\ref{prop:manifold-approx} is unaffected
(Appendix~\ref{app:extended:blocknorm}).

\section{Experiments}
\label{sec:experiments}

\subsection{The Manifold Picture in a Controlled Setting}
\label{sec:exp:toy}

Before ImageNet, we test the theory on a known manifold with adjustable
ambient dimension. We embed a one-dimensional spiral ($\dM = 1$) in
$\mathbb{R}^D$, $D \in \{2, 32, 512\}$, and compare the MLP of
\citet{li2026jit}, a softmax attention block and a Volterra block at
matched parameter count. We report the mean distance of generated points
to the manifold, $d(\manifold)$, which measures whether they lie on the
manifold at all.
Figure~\ref{fig:toy} reproduces the failure pattern of
\citet{li2026jit}. Under $\epsilon$- and $v$-prediction, $d(\manifold)$
grows by one to more than three orders of magnitude as $D$ increases and
the samples lose the spiral, whereas under $x$-prediction every denoiser
stays within a factor of about three of its $D = 2$ value. Attention degrades rather than collapses
under $\epsilon$-prediction, its output being confined to the hull of its
values, the boundedness that caps its polynomial content at affine
(Proposition~\ref{prop:normalized-attention}). Under $x$-prediction the
Volterra block attains the lowest $d(\manifold)$ at every $D$, by a
factor of roughly two to five over the MLP and attention, across a
$256$-fold increase in $D$. The polynomial class is matched to the
geometry of $f_t$, and what it buys is fidelity to the manifold.

\subsection{Training Details}
\label{sec:exp:setup}
\begin{figure}[t]
\centering
\includegraphics[width=0.85\textwidth]{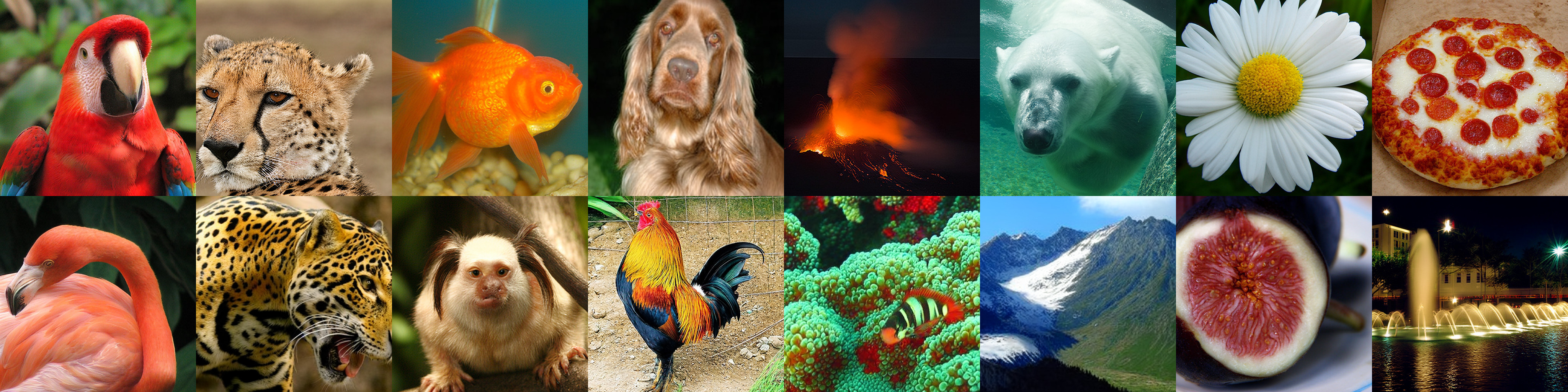}
\caption{\textbf{Class-conditional samples} from \ourmodel-G/16 on
ImageNet $256\times256$.}
\label{fig:samples}
\end{figure}
We train class-conditional models on ImageNet-1K at $256\times256$ and
$512\times512$ with patch sizes $16$ and $32$, so $N = 256$ in both, and
report FID and Inception Score on 50K samples against the ADM reference
statistics \citep{dhariwal2021adm}. Four scales, B, L, H and G, match the
parameters and FLOPs of \citet{li2026jit}, with $\dbot = 128$ for B, L
and $256$ for H, G, token rank $R = 10$, and a channel mixer of hidden
width $h = 2d$ and rank $R' = 1$ (Appendix~\ref{app:arch}). The $x$-prediction head is trained
under the velocity loss of Section~\ref{sec:bg:flow} for 800 epochs with
classifier-free guidance \citep{ho2021cfg} and sampled with a 50-step
Heun solver \citep{heun1900,karras2022edm}; the time distribution, label
dropout and sampler settings are in Appendix~\ref{app:algorithms}. Each
quadratic branch carries a learned per-channel scale initialized to
$0.1$, which can be absorbed into $W^{(2)}$ and leaves the function class
unchanged. Our primary comparison is JiT \citep{li2026jit}, which shares
our recipe and differs only in the mixing operator, with other models
given for context.

\subsection{Main Results on ImageNet}
\label{sec:exp:main}

\begin{table}[t]
\centering
\small
\setlength{\tabcolsep}{4pt}
\caption{ImageNet $256\times256$, FID and IS on 50K samples. Pre-training
columns list the external models each method needs, and parameters
include the tokenizer decoder where used. }
\label{tab:main256}
{%
\footnotesize\setlength{\tabcolsep}{3pt}
\begin{tabular*}{\textwidth}{@{\extracolsep{\fill}}lcccrrcc@{}}
\toprule
& \multicolumn{3}{c}{pre-training} & & & & \\
\cmidrule(lr){2-4}
Method & token & perc. & self-sup. & params & GFLOPs & FID $\downarrow$ & IS $\uparrow$ \\
\midrule
\multicolumn{8}{l}{\emph{Latent-space diffusion}} \\
DiT-XL/2 \citep{peebles2023dit}          & SD-VAE & VGG & --      & 675+49M  & 119 & 2.27 & 278.2 \\
REPA, SiT-XL/2 \citep{yu2025repa}        & SD-VAE & VGG & DINOv2  & 675+49M  & 119 & 1.42 & 305.7 \\
LightningDiT-XL/2 \citep{yao2025lightningdit} & VA-VAE & VGG & DINOv2 & 675+49M & 119 & 1.35 & 295.3 \\
DDT-XL/2 \citep{wang2025ddt}             & SD-VAE & VGG & DINOv2  & 675+49M  & 119 & 1.26 & 310.6 \\
RAE, DiT$^{\text{DH}}$-XL/2 \citep{zheng2025rae} & RAE & VGG & DINOv2 & 839+415M & 146 & 1.13 & 262.6 \\
\midrule
\multicolumn{8}{l}{\emph{Pixel-space, non-diffusion}} \\
JetFormer \citep{tschannen2025jetformer} & -- & -- & -- & 2.8B  & --  & 6.64 & --    \\
FractalMAR-H \citep{li2025fractal}       & -- & -- & -- & 848M  & --  & 6.15 & 348.9 \\
\midrule
\multicolumn{8}{l}{\emph{Pixel-space diffusion}} \\
ADM-G \citep{dhariwal2021adm}            & -- & -- & --     & 554M & 1120 & 4.59 & 186.7 \\
RIN \citep{jabri2023rin}                 & -- & -- & --     & 410M & 334  & 3.42 & 182.0 \\
SiD, UViT/2 \citep{hoogeboom2023simplediff} & -- & -- & --  & 2B   & 555  & 2.44 & 256.3 \\
VDM++, UViT/2 \citep{kingma2023vdmpp}    & -- & -- & --     & 2B   & 555  & 2.12 & 267.7 \\
SiD2, UViT/2 \citep{hoogeboom2025sid2}   & -- & -- & --     & N/A  & 137  & 1.73 & --    \\
SiD2, UViT/1 \citep{hoogeboom2025sid2}   & -- & -- & --     & N/A  & 653  & 1.38 & --    \\
PixelFlow, XL/4 \citep{chen2025pixelflow}& -- & -- & --     & 677M & 2909 & 1.98 & 282.1 \\
PixNerd, XL/16 \citep{wang2025pixnerd}   & -- & -- & DINOv2 & 700M & 134  & 2.15 & 297.0 \\
JiT-B/16 \citep{li2026jit}               & -- & -- & --     & 131M & 25   & 3.66 & 275.1 \\
JiT-L/16 \citep{li2026jit}               & -- & -- & --     & 459M & 88   & 2.36 & 298.5 \\
JiT-H/16 \citep{li2026jit}               & -- & -- & --     & 953M & 182  & 1.86 & 303.4 \\
JiT-G/16 \citep{li2026jit}               & -- & -- & --     & 2B   & 383  & 1.82 & 292.6 \\
\midrule
\multicolumn{8}{l}{\emph{Pixel-space diffusion, polynomial (ours)}} \\
\ourmodel-B/16 & -- & -- & -- & 119M & $27$  & $3.41$ & $282.1$ \\
\ourmodel-L/16 & -- & -- & -- & 391M & $86$  & $2.14$ & $301.6$ \\
\ourmodel-H/16 & -- & -- & -- & 787M & $164$ & $1.63$ & $307.9$ \\
\ourmodel-G/16 & -- & -- & -- & 1.6B & $319$ & $1.48$ & $290.2$ \\
\bottomrule
\end{tabular*}}

\end{table}

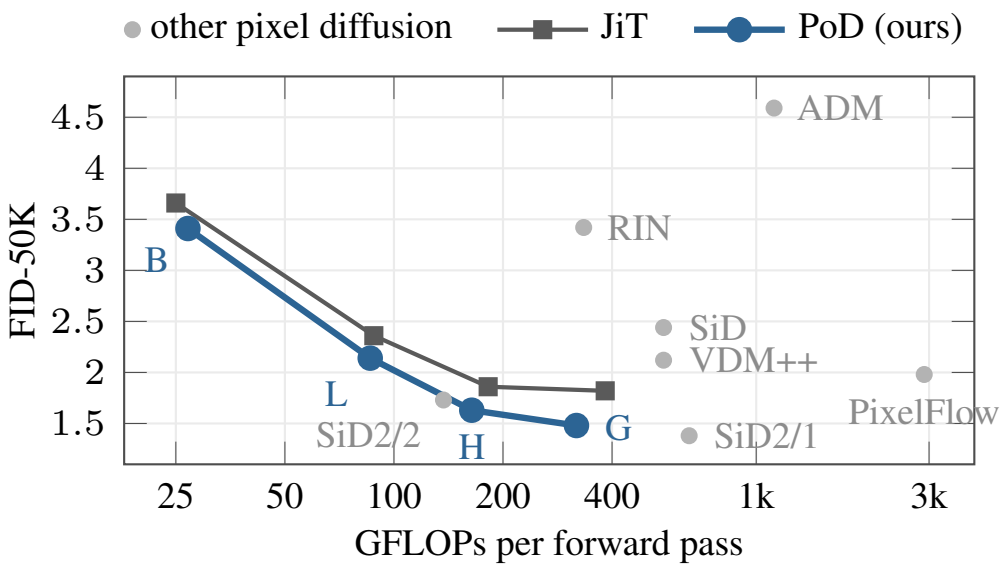
\begin{figure}[t]
\centering
\begin{minipage}[c]{0.60\textwidth}
\centering
\begin{tikzpicture}[font=\scriptsize]
\definecolor{podblue}{RGB}{44,100,148}
\definecolor{jitgray}{RGB}{90,90,90}
\begin{axis}[
  width=205pt, height=118pt,
  xmode=log, xmin=18, xmax=4000, ymin=1.1, ymax=4.9,
  xtick={25,50,100,200,400,1000,3000},
  xticklabels={25,50,100,200,400,1k,3k},
  ytick={1.5,2,2.5,3,3.5,4,4.5},
  xlabel={GFLOPs per forward pass}, ylabel={FID-50K},
  tick label style={font=\scriptsize}, label style={font=\scriptsize},
  xlabel style={yshift=3pt}, ylabel style={yshift=-4pt},
  axis line style={black!70}, tick style={black!70},
  grid=major, grid style={black!8},
  legend columns=3,
  legend style={font=\scriptsize, draw=none, at={(0.5,1.02)}, anchor=south,
                /tikz/every even column/.append style={column sep=6pt}},
  clip=false,
]
\addplot[only marks, mark=*, mark size=1.4pt, black!30] coordinates
  {(1120,4.59) (334,3.42) (555,2.44) (555,2.12) (137,1.73) (653,1.38) (2909,1.98)};
\addlegendentry{other pixel diffusion}
\addplot[jitgray, thick, mark=square*, mark size=1.4pt] coordinates
  {(25,3.66) (88,2.36) (182,1.86) (383,1.82)};
\addlegendentry{JiT}
\addplot[podblue, very thick, mark=*, mark size=1.8pt] coordinates
  {(27,3.41) (86,2.14) (164,1.63) (319,1.48)};
\addlegendentry{\ourmodel{} (ours)}
\node[podblue, anchor=north east] at (axis cs:27,3.41) {B};
\node[podblue, anchor=north east] at (axis cs:84,2.10) {L};
\node[podblue, anchor=north] at (axis cs:164,1.58) {H};
\node[podblue, anchor=west]       at (axis cs:340,1.46) {G};
\node[black!45, anchor=west] at (axis cs:1150,4.59) {ADM};
\node[black!45, anchor=west] at (axis cs:345,3.42) {RIN};
\node[black!45, anchor=west] at (axis cs:580,2.44) {SiD};
\node[black!45, anchor=west] at (axis cs:580,2.12) {VDM++};
\node[black!45, anchor=north east] at (axis cs:134,1.70) {SiD2/2};
\node[black!45, anchor=west] at (axis cs:680,1.38) {SiD2/1};
\node[black!45, anchor=north] at (axis cs:2909,1.93) {PixelFlow};
\end{axis}
\end{tikzpicture}
\end{minipage}\hfill
\begin{minipage}[c]{0.3\textwidth}
\caption{\textbf{Quality against compute on ImageNet $256\times256$.}
\ourmodel{} and JiT at scales B--G, which differ only in the mixing
operator, and other pixel-space diffusion models.}
\label{fig:fidflops}
\end{minipage}
\end{figure}

\ourmodel{} is ahead of JiT in FID at every scale at matched parameters
and FLOPs, most at G ($1.48$ against $1.82$; Table~\ref{tab:main256},
Figure~\ref{fig:fidflops}), and the ordering holds at $512\times512$,
where \ourmodel-G/32 reaches $1.55$ against $1.78$
(Table~\ref{tab:main512}, Appendix~\ref{app:results512}). Among
pixel-space models without pretrained components, \ourmodel-G/16 is the
strongest at comparable compute. SiD2-UViT/1 reaches $1.38$ but at $653$
GFLOPs against our $319$, and SiD2-UViT/2 lies on the PoD curve between L and H. The best latent-space model,
RAE at $1.13$ \citep{zheng2025rae}, remains ahead but relies on a
pretrained tokenizer, a perceptual loss and self-supervised features.
Figure~\ref{fig:samples} shows samples from \ourmodel-G/32. More samples at both resolutions are in
Appendix~\ref{app:samples}.

\subsection{Architecture Ablation}
\label{sec:exp:ablation}

\begin{table}[tb]
\centering
\begin{minipage}[c]{0.31\textwidth}
\caption{Architecture ablation on ImageNet $256 \times 256$ at B/16
scale. 
}
\label{tab:ablation}
\end{minipage}\hfill
\begin{minipage}[c]{0.67\textwidth}
\centering
\footnotesize
\setlength{\tabcolsep}{3pt}
\begin{tabular}{lccc}
\toprule
Configuration              & Token mixer        & Channel mixer       & FID-50K $\downarrow$ \\
\midrule
JiT baseline               & Self-attention     & FFN                  & 3.66 \\
Polynomial token only      & Volterra $k\!=\!2$ & FFN                  & 3.54 \\
Full polynomial & Volterra $k\!=\!2$ & Volterra $k\!=\!2$  & 3.41 \\
\bottomrule
\end{tabular}
\end{minipage}
\end{table}

Table~\ref{tab:ablation} replaces the two Transformer sublayers one at a
time. The Volterra token mixer in place of self-attention lowers FID from
$3.66$ to $3.54$, and the activation-free Volterra channel mixer in place
of the feed-forward network lowers it further to $3.41$, so the gain over
JiT is not carried by a single component. The second step removes the
only activation in the block and still helps, as
Section~\ref{sec:theory:normalized} predicts. A Transformer block needs
its activation because attention contributes no polynomial nonlinearity,
whereas a Volterra mixer supplies its own.

\section{Conclusion}
\label{sec:conclusion}
\label{sec:discussion}
The optimal denoiser is locally a Wiener--Volterra series, approximated on the
data manifold by a polynomial of a low-dimensional linear projection at a
rate set by the intrinsic dimension. Low-rank Volterra stacks represent
that class exactly once their width clears an explicit threshold,
whereas softmax attention, and any mixing rule that averages, meets it
only in its affine part. The resulting architecture, a bottleneck
embedding followed by activation-free Volterra mixers, trains on raw
pixels without pretrained components and improves on JiT in FID at every
scale at $256\times256$ and $512\times512$ at matched parameters and
FLOPs. The analysis assumes an exact smooth manifold, its worst-case
compression term can grow with the ambient dimension, and a rate gap for
attention remains a conjecture. Natural extensions are polynomial
conditioning, where the diffusion time gates the Volterra kernels, and
video and scientific data, where pretrained tokenizers are hard to
obtain.

\subsection*{AI use statement}
\label{app:llm}
In this work, we used generative AI tools to assist in the writing of
proofs, in revising the statements and steps of proofs developed by the
authors. We have not used generative AI tools to help develop theoretical
models or conceptual frameworks, formulate mathematical claims, provide
critical ingredients for proving mathematical claims, propose or refine
hypotheses, design or provide feedback on research methodology or
experiments, implement methods, or interpret results, and generating
synthetic data sets, cleaning and reformatting datasets, qualitative and
thematic data analysis, and translation are not applicable to this work.
Additionally, we used generative AI tools to polish and aid the writing
of the paper, including editing for readability, for the retrieval and
discovery of related work, including searching for and identifying
relevant literature, and to create or modify scientific figures. We have
reviewed all AI-assisted work. Every reference surfaced with AI
assistance was checked against the original source, and all AI-assisted
text, proofs and figures were read and revised by the authors. We take
responsibility for the final content of this work, including text,
claims or artifacts produced with the aid of generative AI.

\subsection*{Reproducibility statement}
All theoretical claims are stated with their assumptions in
Section~\ref{sec:theory}, and complete proofs are given in
Appendix~\ref{app:proof}, with supporting derivations in
Appendix~\ref{app:extended}. The architecture is specified in
Section~\ref{sec:architecture}, with the derivations behind the
primitive and the mixers in Appendix~\ref{app:arch}. The training
objective, schedule and sampler are given in Section~\ref{sec:exp:setup},
with pseudocode in Appendix~\ref{app:algorithms}. All experiments use the public ImageNet-1K
dataset with the standard ADM evaluation protocol.

\bibliography{references}
\bibliographystyle{iclr2027_conference}

\appendix

\clearpage
\begin{center}
{\LARGE\sc Appendix}
\end{center}
\vspace{0.5em}

\section{Notation}
\label{app:notation}

\begin{table}[H]
\centering
\small
\caption{Summary of notation.}
\label{tab:notation}
\begin{tabularx}{\linewidth}{@{}lX@{}}
\toprule
Symbol & Meaning \\
\midrule
\multicolumn{2}{@{}l}{\emph{Dimensions and indices}}\\
$D = \Dp$      & ambient dimension of the theory, the per-patch pixel dimension \\
$\Dp = 3p^2$   & ambient dimension of one $p\times p$ RGB patch ($\Dp \in \{768, 3072\}$) \\
$p$            & patch side length ($p = 16$ at $256\times256$, $p = 32$ at $512\times512$) \\
$\dM$          & intrinsic dimension of the data manifold $\manifold$, with $\dM \ll D$ \\
$\dbot$        & per-token bottleneck dimension ($\dbot \in \{128, 256\}$); $2\dM + 1$ suffices in Proposition~\ref{prop:manifold-approx}, and the trained value is deliberately larger (Appendix~\ref{app:proof}) \\
$d$            & hidden width of the network, the working dimension of all token and channel mixers; $d \ge \dbot$ \\
$N$            & number of patch tokens per image ($N = 256$) \\
$N$            & number of coefficients of a degree-$K$ polynomial on $\mathbb{R}^{\dbot}$, per output coordinate; $N = \binom{\dbot+K}{\dbot} \asymp K^{2\dM+1}$ \\
$R,\ R'$       & rank of the Volterra primitive~\eqref{eq:arch:eval} in the token and channel mixers, the number of products summed per hidden unit ($R = 10$, $R' = 1$) \\
$n_{\mathrm{h}}$  & number of attention heads (Appendices~\ref{app:ffn} and~\ref{app:proof-attention}) \\
\midrule
\multicolumn{2}{@{}l}{\emph{Diffusion and denoising}}\\
$x,\ \epsilon$ & clean image $x \sim \rho_{\mathrm{data}}$ and noise $\epsilon \sim \mathcal{N}(0, I)$ \\
$t,\ \sigma$   & flow time $t \in (0,1)$ and noise level $\sigma := 1 - t$ \\
$z_t$          & noised observation, $z_t = tx + (1-t)\epsilon$ \\
$\rho_{\mathrm{data}}$ & data distribution; $d\rho_{\mathrm{data}}(x)$ denotes integration against it \\
$f_t(z)$       & optimal (MMSE) denoiser, $f_t(z) = \expt{x \mid z_t = z}$ \\
$\hat{x}_\theta$ & the network's prediction of the clean image $x$ \\
$\tilde{\rho}$ & $\rho_{\mathrm{data}}$ reweighted by $e^{-t^2\norm{x}^2/2\sigma^2}$ and renormalized \\
$\theta$       & natural parameter $\theta := tz/\sigma^2$ of the tilt of $\tilde{\rho}$ (unrelated to the network parameters in $\hat{x}_\theta$) \\
$\Lambda$      & cumulant generating function of $\tilde{\rho}$, $\Lambda(\theta) = \log\int e^{\theta^\top x}\,d\tilde{\rho}(x)$ \\
$\kappa_n$     & $n$-th cumulant tensor of $\tilde{\rho}$; the order-$n$ Volterra kernel of $f_t$ is $\kappa_{n+1}$ scaled by $(t/\sigma^2)^n/n!$ \\
\midrule
\multicolumn{2}{@{}l}{\emph{Manifold and approximation}}\\
$\manifold \subset \mathbb{R}^D$ & data manifold; $\tau$ its reach \\
$\pi$          & linear bottleneck projection $\mathbb{R}^D \to \mathbb{R}^{\dbot}$ (Proposition~\ref{prop:manifold-approx}) \\
$\rho,\ \rho_{\min}$ & density of $\rho_{\mathrm{data}}$ with respect to the volume measure of $\manifold$, and its lower bound (Lemma~\ref{lem:concentration}) \\
$\tau_\pi$     & reach of the projected manifold $\pi(\manifold)$ (Lemma~\ref{lem:concentration}(b)) \\
$\delta,\ \mathcal{T}_\delta$ & tube radius, and the tube $\{z : \operatorname{dist}(z, t\manifold) \le \delta\}$ about the scaled manifold \\
$r,\ \mathrm{np}$ & nearest-point retraction onto $t\manifold$ in $\mathbb{R}^D$, and nearest-point projection onto $\pi(t\manifold)$ in $\mathbb{R}^{\dbot}$ \\
$\eta,\ C_\eta$ & normal-sensitivity factor of Lemma~\ref{lem:concentration}, $\eta = C_\eta \min\{1/t,\ \sigma^2/(t^3\tau^2)\}$, and its constant \\
$c_\pi$        & slippage constant of Lemma~\ref{lem:concentration}(b), set by $\pi$ and $\manifold$ \\
$s,\ K$        & smoothness order of $f_t$; polynomial degree \\
$P_K$          & degree-$K$ polynomial approximant $\mathbb{R}^{\dbot} \to \mathbb{R}^D$ of Proposition~\ref{prop:manifold-approx} (distinct from the factor $P$) \\
$\mathcal{P}_K$ & space of polynomial maps $\mathbb{R}^{\dbot} \to \mathbb{R}^D$ of total degree at most $K$ \\
$K_L$          & $K_L = 2^L$, the degree reached by a depth-$L$ stack of degree-$2$ layers \\
\midrule
\multicolumn{2}{@{}l}{\emph{Architecture}}\\
$L$            & number of stacked \ourmodel{} blocks \\
$k$            & Volterra order (degree) per layer, default $k = 2$ \\
$n$            & size of the axis a mixer acts on: $n = N$ for the token mixer, $n = d$ for the channel mixer \\
$h$            & hidden width of the primitive ($h = N$ in the token mixer, $h = 2d$ in the channel mixer) \\
$P,\ Q^{(m)}_q$ & factors of the primitive, $Q^{(m)}_q \in \mathbb{R}^{h \times n}$ projects input slot $m$ of group $q$ and $P \in \mathbb{R}^{d_{\text{out}} \times h}$ mixes back (tied variant, $Q^{(m)}_q$ equal across $m$) \\
$W^{(k)}_{\mathrm{tok}},\ W^{(k)}_{\mathrm{ch}}$ & order-$k$ Volterra kernels of the token and channel mixers \\
$\alpha_{\mathrm{tok}}, \alpha_{\mathrm{ch}}$ & per-channel, zero-initialized residual gates \\
$\gamma,\ \beta$ & AdaLN scale and shift computed from the time $t$ and class label $c$, applied as $\gamma \odot \mathrm{RMSNorm}(\cdot) + \beta$ \\
\bottomrule
\end{tabularx}
\end{table}

\begin{table}[H]
\centering
\small
\begin{tabularx}{\linewidth}{@{}lX@{}}
\multicolumn{2}{@{}l}{Table~\ref{tab:notation}, continued.}\\[2pt]
\toprule
Symbol & Meaning \\
\midrule
\multicolumn{2}{@{}l}{\emph{Architecture}}\\
$\Pbud$        & per-layer parameter budget in the function-class statements of Section~\ref{sec:theory} (distinct from the factor $P$) \\
$\mathcal{V}_{L, \Pbud}$ & depth-$L$ stacks of low-rank Volterra mixers with at most $\Pbud$ parameters per layer (Proposition~\ref{prop:separation}); $\Pbud_0$ is the attainment threshold \\
\bottomrule
\end{tabularx}
\end{table}

\section{Extended Derivations and Discussion}
\label{app:extended}

This appendix holds material referenced from Section~\ref{sec:theory} in
summary form: the derivation of the cumulant expansion and the
condition under which it terminates, the all-order Tweedie identity, the
prediction-target discussion, the covering argument across the sampling
schedule, the extended comparison against attention-shaped operators,
the role of normalization and conditioning in our block, and the
feed-forward objection. It gives derivations and discussion; the proofs themselves
are in Appendix~\ref{app:proof}.

\subsection{The cumulant expansion, in full}
\label{app:extended:cgf}

This subsection derives the expansion quoted as
Equation~\eqref{eq:theory:cumulants} in Section~\ref{sec:theory},
verifies it against the Jacobian identity, and extends it to every order
of differentiation.

\paragraph{The three-term split.}
Expanding the square in the exponent of the denoiser formula (left of~\eqref{eq:theory:mmse}) separates
the observation from the signal,
\begin{equation}
-\frac{\norm{z - tx}^2}{2\sigma^2}
\;=\;
\underbrace{-\frac{\norm{z}^2}{2\sigma^2}}_{\text{$z$ alone}}
\;+\;
\underbrace{\frac{t}{\sigma^2}\, z^\top x}_{\text{the coupling}}
\;-\;
\underbrace{\frac{t^2}{2\sigma^2}\norm{x}^2}_{\text{$x$ alone}},
\label{eq:theory:split}
\end{equation}
and the three terms play different roles. The first depends on $z$
alone, so it factors out of both integrals on the left of~\eqref{eq:theory:mmse}
and cancels between them. The third depends on $x$ alone, so rather
than being carried through the calculation it can be absorbed into the
measure one integrates against: write $\tilde{\rho}$ for the data
distribution reweighted by $e^{-t^2\norm{x}^2/2\sigma^2}$ and
renormalized, a fixed measure that mildly down-weights images of large
norm and, crucially, does not depend on the observation. Only the
middle term couples $z$ to $x$, and it does so \emph{linearly}, through
the single vector $\theta := t z/\sigma^2$. Substituting
into the left of~\eqref{eq:theory:mmse} and cancelling the $z$-only factor and the
normalizing constant of $\tilde{\rho}$ between numerator and denominator
leaves
\[
f_t(z)
\;=\;
\frac{\int x\, e^{\theta^\top x}\, d\tilde{\rho}(x)}
     {\int e^{\theta^\top x}\, d\tilde{\rho}(x)}.
\]

\paragraph{The tilt is a gradient.}
The right-hand side is the mean of $\tilde{\rho}$ reweighted by
$e^{\theta^\top x}$, an \emph{exponential tilt} of $\tilde{\rho}$ with
natural parameter $\theta$. Such a mean is always a gradient. Setting
$\Lambda(\theta) := \log\int e^{\theta^\top x}\, d\tilde{\rho}(x)$ and
differentiating under the integral,
\[
\nabla_\theta \Lambda(\theta)
\;=\;
\frac{\int x\, e^{\theta^\top x}\, d\tilde{\rho}(x)}
     {\int e^{\theta^\top x}\, d\tilde{\rho}(x)},
\]
which is exactly the mean of $\tilde{\rho}$ under the tilt. Hence
\begin{equation}
f_t(z) \;=\; \nabla_\theta \Lambda(\theta)\big|_{\theta = \frac{t}{\sigma^2} z},
\qquad
\Lambda(\theta) := \log \int_{\manifold} e^{\theta^\top x}\, d\tilde{\rho}(x),
\qquad
d\tilde{\rho} \;\propto\; e^{-\frac{t^2}{2\sigma^2}\norm{x}^2}\, d\rho_{\mathrm{data}}.
\label{eq:theory:cgf}
\end{equation}
Taking the logarithm is what turns the unnormalized integral into a
mean: with $M(\theta) := \int e^{\theta^\top x} d\tilde{\rho}$ we have
$\nabla_\theta \Lambda = \nabla_\theta M / M$, and it is the denominator
that normalizes the weights. By definition $\Lambda$ is then the
\emph{cumulant generating function} of $\tilde{\rho}$, and its Taylor
coefficients at $\theta = 0$ are the cumulant tensors $\kappa_n$. $\kappa_1$ is the mean, $\kappa_2$ the covariance, and the higher $\kappa_n$ are 
$n$-way arrays, symmetric in their indices. Differentiating that Taylor expansion once and
substituting $\theta = t z/\sigma^2$ gives~\eqref{eq:theory:cumulants},
whose kernels are therefore named
rather than left abstract. 

\paragraph{The expansion is local.}
The Taylor series of $\Lambda$ at $\theta = 0$, and with
it~\eqref{eq:theory:cumulants}, has a finite radius of convergence in
general. Since $\tilde{\rho}$ has compact support, $M$ extends to an
entire function of $\theta \in \mathbb{C}^D$ that is positive on
$\mathbb{R}^D$, so $\Lambda$ is real-analytic on all of $\mathbb{R}^D$.
But $\Lambda = \log M$ is singular wherever $M$ vanishes, and $M$
generally has complex zeros, so along each real direction the series in
$\theta$ converges only up to the nearest zero on the complex line
through that direction. For the
uniform measure on a circle of radius $a$ about the origin (a case in
which $\tilde{\rho} = \rho_{\mathrm{data}}$, since every point has the
same norm), $M(\theta) = I_0(a\norm{\theta})$ for real $\theta$ in the
plane of the circle, and
its nearest complex zeros lie at $\norm{\theta} = j_{0,1}/a$, with
$j_{0,1} \approx 2.405$ the first zero of the Bessel function $J_0$. With
$\theta = tz/\sigma^2$ and $\norm{z} = ta$ on $t\manifold$, the series
converges on $t\manifold$ only while $t^2 a^2/\sigma^2 < j_{0,1}$, and
diverges there late in sampling. The expansion therefore identifies the
polynomial class and names its kernels, but it is not the approximant.
Proposition~\ref{prop:manifold-approx} rests instead on the
real-analyticity of $f_t$ on all of $\mathbb{R}^D$, which holds at every
$t$, and its polynomial $P_K$ is not a truncation of the series
(see the remark after its proof in Appendix~\ref{app:proof}).

\paragraph{When the expansion terminates.}
The expansion terminates exactly when $\tilde{\rho}$ is Gaussian, since
a Gaussian is characterized by the vanishing of all cumulants beyond the
second. For $\tilde{\rho} = \mathcal{N}(\mu, S)$ one gets
$\Lambda(\theta) = \theta^\top \mu + \tfrac{1}{2}\theta^\top S \theta$
and hence $f_t(z) = \mu + \tfrac{t}{\sigma^2} S z$ with no truncation.
Substituting the precision addition
$S^{-1} = \Sigma_\rho^{-1} + \tfrac{t^2}{\sigma^2} I$ recovers the
classical Wiener filter, the linear MMSE estimator that the
Wiener--Volterra series generalizes. This case sits outside our setting, which requires compact support
(Section~\ref{sec:bg:manifold}), and the contrast is informative. With
compact support, $f_t$ is affine only if
$\rho_{\mathrm{data}}$ is a point mass (Lemma~\ref{lem:not-affine}), so
within our setting the expansion never terminates, i.e. the optimal denoiser
lies in the closure of the polynomial class rather than in the class
itself, approximated to arbitrary accuracy by polynomials of sufficient
degree but never equal to one. This is why
Proposition~\ref{prop:manifold-approx} is an approximation result.
What classical theory leaves open is
the degree required, and this is what
Proposition~\ref{prop:manifold-approx} settles. The
real-analyticity of $f_t$ makes the best degree-$K$ approximation error
decay faster than any fixed power of $1/K$, while the low intrinsic
dimension of $\manifold$ makes the degree, and hence the parameter
count, depend on $\dM$ rather than the ambient $D$.

\paragraph{An all-order Tweedie identity.}
The cumulant form~\eqref{eq:theory:cgf} sharpens the Jacobian
identity (right of~\eqref{eq:theory:mmse}) to every order of differentiation.
Since $f_t(z) = \nabla_\theta \Lambda(\theta)$ at
$\theta = tz/\sigma^2$, each derivative in $z$ brings down a factor
$t/\sigma^2$ by the chain rule, and the higher derivatives of a
cumulant generating function at parameter $\theta$ are, by definition,
the cumulants of the measure tilted by $e^{\theta^\top x}$, which at
$\theta = tz/\sigma^2$ is exactly the posterior $p(\cdot \mid z_t = z)$.
Hence
\begin{equation}
\nabla_z^{k} f_t(z) \;=\;
\Big(\tfrac{t}{\sigma^2}\Big)^{k}\,
\kappa_{k+1}\big(p(\cdot \mid z_t = z)\big),
\qquad k \ge 1,
\label{eq:theory:tweedie}
\end{equation}
of which the right of~\eqref{eq:theory:mmse} is the case $k = 1$, the second
cumulant being the covariance. Both identities are proved formally in
Appendix~\ref{app:proof}; the derivation here is the informal route.
The identity
makes the $t$-dependence of the approximation theory explicit rather
than opaque: the norm $\norm{f_t}_{C^s}$, which controls the constant
$C(s,t)$ of Proposition~\ref{prop:manifold-approx}(b), is
by~\eqref{eq:theory:tweedie} a statement about the posterior cumulants
up to order $s{+}1$, scaled by powers of $t/\sigma^2$, so uniform-in-$t$
control of the constant is equivalent to uniform-in-$t$ bounds on
posterior cumulants. Both ends of the schedule are benign on this
reading. Late in sampling the posterior concentrates at scale
$\sigma/t$ and near-Gaussianity leaves
$\kappa_{k+1} = O\big((\sigma^2/t^2)^{k}\,\tau^{-(k-1)}\big)$, with
$\tau$ the reach of $\manifold$ introduced with
Lemma~\ref{lem:concentration}, so the powers of $\sigma$ cancel exactly
and $\nabla_z^k f_t = O\big(t^{-k}\tau^{-(k-1)}\big)$; early in
sampling the posterior relaxes to $\rho_{\mathrm{data}}$, whose
cumulants are bounded on the compact $\manifold$, and
$\nabla_z^k f_t = O(t^k)$. Any growth of $C(s,t)$ is therefore
confined to intermediate $t$, consistent with the covering argument of
Appendix~\ref{app:extended:schedule}.

\paragraph{The denominator.}
The denominator on the left of~\eqref{eq:theory:mmse} is the marginal density
$p_{z_t}(z) = \int_{\manifold} p(z \mid x)\, d\rho_{\mathrm{data}}(x)$,
which normalizes the weights. The Gaussian prefactor
$(2\pi\sigma^2)^{-D/2}$ is common to numerator and denominator and
cancels, so the exponential written there is proportional to the
likelihood rather than equal to it.

\subsection{Prediction targets and learnability}
\label{app:extended:targets}

Section~\ref{sec:theory} works throughout with the clean-image denoiser
$f_t$, whereas a diffusion model may be trained against any of four
interchangeable targets (clean image, score, noise, velocity), and
the choice among them is usually treated as consequential. Two questions
follow, and they have opposite answers. The approximation theory does not
depend on the choice: the four targets are affine reparametrizations of
one another at fixed $t$, so they span the same polynomial class and
Proposition~\ref{prop:manifold-approx} applies verbatim to each. The
architecture does depend on it, but the equivalence locates where. It
separates two decisions that are ordinarily conflated, namely which
quantity the network's output layer emits and in which coordinates the
training error is measured, and only the first is constrained by the manifold
hypothesis. Section~\ref{sec:bg:flow} accordingly settles them
separately, with an $x$-prediction head and a $v$-space loss.

\paragraph{The four targets span one class.}
The score of the noised marginal is a rescaling of the denoiser. Since
$p_t(z) \propto \int e^{-\norm{z-tx}^2/2\sigma^2}\,d\rho_{\mathrm{data}}(x)$,
differentiating the logarithm brings $(tx - z)/\sigma^2$ down inside the
integral and divides by the integral itself, leaving a posterior
expectation. This is the same differentiation under the integral sign that
identifies $f_t$ with $\nabla_\theta \Lambda$ in
Appendix~\ref{app:extended:cgf}:
\begin{equation}
\nabla_z \log p_t(z) \;=\; \frac{t\, f_t(z) - z}{\sigma^2},
\qquad\text{equivalently}\qquad
f_t(z) \;=\; \frac{z + \sigma^2\, \nabla_z \log p_t(z)}{t}.
\label{eq:theory:score}
\end{equation}
The noise and velocity targets follow from $z_t = t x + \sigma \epsilon$
and $v = x - \epsilon$, using $t + \sigma = 1$:
\begin{equation*}
\expt{\epsilon \mid z_t = z} \;=\; \frac{z - t f_t(z)}{\sigma},
\qquad\qquad
\expt{v \mid z_t = z} \;=\; \frac{f_t(z) - z}{\sigma}.
\end{equation*}
Each of the three is an invertible affine function of $f_t(z)$ and $z$
whose coefficients depend on $t$ alone. At fixed $t$ such a map carries a
degree-$K$ polynomial in $z$ to a degree-$K$ polynomial in $z$, so the
four targets generate the same class and
Proposition~\ref{prop:manifold-approx} transfers to each with only the
constant changed. The approximation theory of Section~\ref{sec:theory} is
in this sense a statement about diffusion denoisers in general rather
than about one parametrization.

\paragraph{The loss coordinates are a reweighting.}
Section~\ref{sec:theory:denoiser} states that the flow-matching loss is
the $x$-prediction error rescaled by $\sigma^{-2}$; here is the
substitution. Putting $v_\theta = (\hat{x}_\theta - z_t)/\sigma$ and
$v = x - \epsilon$ into the velocity loss of Section~\ref{sec:bg:flow}, and using
$z_t = t x + \sigma\epsilon$ so that the noise cancels and $t+\sigma = 1$
so that the signal collapses as well,
\begin{equation*}
v_\theta - v \;=\; \frac{\hat{x}_\theta - z_t}{\sigma} - (x - \epsilon)
\;=\; \frac{\hat{x}_\theta - (t+\sigma)\,x}{\sigma}
\;=\; \frac{\hat{x}_\theta - x}{\sigma}.
\end{equation*}
The factor $\sigma^{-2}$ is positive and independent of $\theta$, so
minimizing the loss at each $t$ minimizes
$\expt{\norm{\hat{x}_\theta - x}^2}$, whose minimizer over measurable
functions of $z_t$ is the conditional mean $f_t$; training drives
$\hat{x}_\theta(\cdot, t)$ toward it. The factor is not constant in $t$,
so the $t$-averaged objective weights low-noise times heavily, which is
why the denominator is clipped near $t = 1$
(Section~\ref{sec:bg:flow}). The same substitution
applied to the noise target gives
$\norm{\hat{\epsilon}_\theta - \epsilon}^2 =
(t^2/\sigma^2)\,\norm{\hat{x}_\theta - x}^2$. Measuring the error in
$\epsilon$- or $v$-coordinates therefore alters the weight each noise
level carries in the $t$-averaged objective and nothing else, and both
up-weight the low-noise end. This is a choice about
optimization, and it is made freely. It also cannot explain the failure
of $\epsilon$- and $v$-prediction in Figure~\ref{fig:toy}. The weights
depend on $t$ alone and not on the ambient dimension $D$, whereas that
failure grows with $D$ at fixed $\dM$.

\paragraph{The output space is not.}
What the final layer must emit is a different matter, and here the
targets are not interchangeable. The patch embedding of
Section~\ref{sec:arch:tokenization} passes each patch through the
rank-$\dbot$ projection $\pi$ before anything else, so everything the
network computes depends on a patch $z$ only through $\pi(z)$. The
noised input lies at distance about $\sigma\sqrt{D-\dM}$ from
$t\manifold$, and the question for each head is how much of its target
survives this compression. For the clean image, most of it does. Late in
sampling the posterior concentrates (Step~2 of the proof of
Lemma~\ref{lem:concentration}), so $f_t(z)$ lies within
$O(\sigma^2/(t^2\tau))$ of $\manifold$, and by
Lemma~\ref{lem:concentration}(b) it is a function of $\pi(z)$ up to the
compression term of~\eqref{eq:theory:combined}. Once the lemma applies to
a typical input (Appendix~\ref{app:extended:schedule}), that term is at
most $O((D-\dM)\,\sigma^2/t^2)$ and vanishes as $\sigma \to 0$. For the
noise, little of it does. Late in sampling the noise target
$\expt{\epsilon \mid z_t = z} = (z - t f_t(z))/\sigma$ is essentially the
component of $\epsilon$ normal to the manifold, a vector of length about
$\sqrt{D-\dM}$ spread evenly over the normal directions. The part of
$\epsilon$ in the $D - \dbot$ directions that $\pi$ discards is
independent of $\pi(z)$ and of the other patches, so no function of them
recovers it, and an $\epsilon$-head incurs a squared error of at least
$D - \dbot$ at every noise level. The same holds for a head emitting
$v = x - \epsilon$. The prediction is therefore about the ambient
dimension. At fixed $\dM$, the error an $\epsilon$- or $v$-head incurs
through the bottleneck grows with $D$ at every noise level, while that of
an $x$-head is bounded, late in sampling, by a term that vanishes as
$\sigma \to 0$.

A second and softer effect compounds the first. When $\sigma$ is small
the posterior concentrates, so $z \approx t f_t(z)$ and the noise target
is a difference of two nearly equal quantities divided by $\sigma$;
obtaining it to fixed relative accuracy demands relative accuracy of
order $\sigma$ in the manifold quantity. At matched loss weighting this
makes no difference to the population optimum, which is why it is stated
as a claim about capacity allocation under a finite budget rather than
about function classes.

The empirical necessity of $x$-prediction reported by \citet{li2026jit},
and reproduced in the controlled study of Section~\ref{sec:exp:toy}, is on
this reading a statement about the output layer and not about function
classes; the classes coincide. That study separates the two accounts by
varying $D$ at fixed $\dM$, and the distinction it draws is between heads
rather than losses: the runs labelled $v$-prediction there emit $v$
directly, whereas the architecture of Section~\ref{sec:architecture}
emits $\hat{x}_\theta$ and forms $v_\theta$ from it afterwards by an
explicit formula in $z_t$, and is therefore an $x$-head trained under a
$v$-space loss, which is not the configuration that fails. The affine maps
above are invertible and computable in closed form from $z$, so they can
be applied outside the network, which leaves the bottleneck responsible
for the manifold-valued factor alone.

\subsection{Behaviour across the sampling schedule}
\label{app:extended:schedule}

Proposition~\ref{prop:manifold-approx}, Lemma~\ref{lem:concentration}
and the combined bound~\eqref{eq:theory:combined} hold at a fixed $t$,
while sampling runs $t$ from $0$ to $1$. This subsection checks that the
bound stays controlled along the whole path. What changes along the path
is the noise-to-signal ratio $\sigma/t = (1-t)/t$, which falls from
$\infty$ to $0$. We compare it with the reach $\tau$ of the manifold, the
largest distance from $\manifold$ within which every point has a unique
nearest point on $\manifold$ (for a circle, the reach is its radius).
Given a noised input, the posterior over clean images spreads by about
$\sigma/t$ along the manifold (Step~1 of the proof of
Lemma~\ref{lem:concentration}), so comparing $\sigma/t$ with $\tau$ asks
whether the posterior sees a nearly flat piece of the manifold or a
curved one. This gives three regimes, summarized in
Table~\ref{tab:schedule} and taken below in the order sampling visits
them. A final paragraph explains where Lemma~\ref{lem:concentration}
itself applies, which is a separate question.

\begin{table}[h]
\centering
\small
\caption{The three regimes of the sampling schedule, in the order
sampling visits them.}
\label{tab:schedule}
\begin{tabular}{@{}llll@{}}
\toprule
Regime & $\sigma/t$ & Why the bound is controlled & Constrains \\
\midrule
Early        & $\gg \tau$  & target nearly affine & --- \\
Intermediate & $\sim \tau$ & tangent span; analyticity and depth & width, degree \\
Late         & $\ll \tau$  & posterior concentrated; compression cost vanishes & --- \\
\bottomrule
\end{tabular}
\end{table}

\paragraph{Early, $\sigma/t \gg \tau$.}
The observation carries almost no information about $x$, so the
posterior stays close to $\rho_{\mathrm{data}}$ and its covariance
relaxes to $\Sigma_\rho = \Cov_{\rho_{\mathrm{data}}}(x)$. By the
Jacobian identity (right of~\eqref{eq:theory:mmse}),
$\nabla_z f_t \approx \tfrac{t}{\sigma^2}\Sigma_\rho$. What matters is
not that this Jacobian is small but that it is nearly the same at every
$z$, and a map with a constant Jacobian is affine. Truncating the
expansion~\eqref{eq:theory:cumulants} after its linear term gives
$f_t(z) \approx \bar{x} + \tfrac{t}{\sigma^2}\Sigma_\rho\, z$, with
$\bar{x}$ the data mean, and the higher-order terms carry higher powers
of $t$. A degree-one polynomial therefore nearly suffices, and
Lemma~\ref{lem:concentration} is not needed.

\paragraph{Intermediate, $\sigma/t \sim \tau$.}
This is the hard regime. The posterior spreads over a patch of the
manifold comparable in size to the reach, so the patch is curved and
$f_t$ is genuinely nonlinear. In the proof of
Lemma~\ref{lem:concentration} (Step~2), the spread of the posterior away
from the manifold is smaller than its spread along it by the factor
$(\sigma/t\tau)^2$, and here that factor is of order one, so the
posterior no longer lies close to a single tangent plane. Two facts keep
the bound controlled, one for each design parameter. For the
\emph{width}, the bottleneck must retain the span of the tangent spaces
across the patch. This is larger than a single tangent space, but it is
set by $\dM$ and the curvature, not by $D$
(Remark~\ref{rem:dbot-width}). For the \emph{degree}, $f_t$ is
real-analytic at every $t$ by
Proposition~\ref{prop:manifold-approx}(a), so the degree needed for a
given accuracy is largest here but finite, and depth supplies it, since
an $L$-layer stack reaches degree $2^L$. Neither fact involves $D$.

\paragraph{Late, $\sigma/t \ll \tau$.}
The posterior is concentrated near a single point of the manifold, on a
patch small enough to be nearly flat, and the target is as simple as it
gets. Once Lemma~\ref{lem:concentration} applies to a typical input (see
below), the compression term of~\eqref{eq:theory:combined} is
$O((D-\dM)\,\sigma^2/t^2)$ and vanishes as $\sigma \to 0$ at any fixed
$D$. The shrinking distance does the work, since $\eta \to 0$ but
$c_\pi/t \to c_\pi$ does not. Averaged over the noise, the slippage part
of the term scales with $\dbot$ instead of $D$
(Remark~\ref{rem:typical-slippage}).

\paragraph{Where Lemma~\ref{lem:concentration} applies.}
The lemma bounds the worst case over all inputs within distance
$\delta < t\tau$ of the scaled manifold $t\manifold$. A typical noised
input is farther away than $\sigma$ suggests. Write $z = tx + \sigma\epsilon$
with $x \in \manifold$ and $\epsilon \sim \mathcal{N}(0, I_D)$. Of the $D$
directions of $\epsilon$, $\dM$ run along the manifold at $x$ and
$D - \dM$ are perpendicular to it. The perpendicular part has length
about $\sqrt{D-\dM}$, so $z$ lies at distance about
$\sigma\sqrt{D-\dM}$ from $t\manifold$. The lemma therefore applies to a
typical input only once $\sigma\sqrt{D-\dM} < t\tau$, that is
$\sigma/t < \tau/\sqrt{D-\dM}$, a threshold well below $\tau$ when $D$
is large (about $27$ times smaller for a $16\times16$ RGB patch).
Part~(b) needs in addition $\norm{\pi}\,\sigma\sqrt{D-\dM} \le t\tau_\pi/2$,
the same condition with $\tau$ replaced by $\tau_\pi/(2\norm{\pi})$.
This factor $\sqrt{D-\dM}$ measures how far the \emph{input} $z$ lies
from $t\manifold$, not the target $f_t(z)$, which late in sampling stays
within $O(\sigma^2/(t^2\tau))$ of $\manifold$
(Appendix~\ref{app:extended:targets}). It limits where the lemma's
worst-case argument reaches, not how hard $f_t$ is to approximate.

This does not make the posterior less concentrated between
$\tau/\sqrt{D-\dM}$ and $\tau$. By the left of~\eqref{eq:theory:mmse}, the posterior
gives each candidate clean image $x' \in \manifold$ the weight
$e^{-\norm{z - tx'}^2/2\sigma^2}\, d\rho_{\mathrm{data}}(x')$, so comparing
two candidates $x'$ and $x$ depends only on the difference of their
exponents and on the data density at the two points. Substituting $z = tx + \sigma\epsilon$,
\[
\norm{z - tx'}^2 - \norm{z - tx}^2
\;=\; \norm{t(x - x') + \sigma\epsilon}^2 - \norm{\sigma\epsilon}^2
\;=\; t^2\norm{x - x'}^2 + 2t\sigma\,\epsilon^\top(x - x').
\]
The term $\sigma^2\norm{\epsilon}^2$ is the squared distance of $z$ from
the true point $tx$. It is the same for every candidate, so it cancels,
and with it the large distance $\sigma\sqrt{D-\dM}$ of $z$ from the
manifold drops out. What remains depends on the noise only through
$\epsilon^\top(x - x')$, which is $\norm{x - x'}$ times the component of
$\epsilon$ along the direction from $x'$ to $x$. That component is a
single standard Gaussian number, not a sum over $D$ coordinates. As $x'$
ranges over the manifold, these directions form a family of dimension
$\dM$, so the largest such component grows like $\sqrt{\dM}$ rather
than $\sqrt{D}$. The posterior therefore concentrates at scale about
$\sigma\sqrt{\dM}/t$, as if the noise acted only along the manifold,
and it is sharp once $\sigma/t$ is well below $\tau$ up to factors of
$\dM$. Between the two thresholds the target thus behaves as in the late
regime, and only the worst-case bound of the lemma does not yet apply.
We give this argument as a sketch for a typical input rather than as a
bound over the tube.

\paragraph{Summary.}
Only the intermediate regime constrains the architecture. The two ends
are easy for opposite reasons. Early in sampling the target is nearly
affine even though the input is nowhere near the manifold, and late in
sampling the geometry is clean. The ambient dimension does not enter how
hard the target is in any regime. It enters only the worst-case bound,
through where Lemma~\ref{lem:concentration} begins to apply and through
the size of the compression term late in sampling. Finally,
Proposition~\ref{prop:separation} and Conjecture~\ref{conj:rate-gap}
compare Volterra and attention stacks on the same $f_t$, so the
comparison does not depend on how hard any individual $t$ is.

\subsection{The unnormalized boundary, and what distinguishes the
Volterra parametrization}
\label{app:extended:boundary}

Propositions~\ref{prop:separation}(ii)
and~\ref{prop:normalized-attention} are complementary rather than
redundant. The first is sharper in hypothesis, needing agreement only on
an open set, and is available for softmax because analyticity propagates
that agreement globally; the second is far broader in reach, applying to
weightings that are not analytic at all. For real-analytic weightings the
two coincide, since agreement on an open set extends to
$\mathbb{R}^{N \times d}$.

Proposition~\ref{prop:normalized-attention} is also sharp: normalization
is not a technical convenience in the proof but the whole of it. Remove it
and the conclusion genuinely fails. Unnormalized linear attention with
the identity feature map,
\begin{equation*}
h_i(X) \;=\; x_i \;+\; \sum_{j=1}^{N}
\big(x_i^\top W_Q^\top W_K x_j\big)\, W_O W_V x_j,
\end{equation*}
is a cubic polynomial map of $X$, and a stack of such layers lies inside
the polynomial class. This is the one attention-shaped operator our
results do not exclude, and we regard it as a corroboration rather than
a counterexample: the claim of this section is about the function class
the denoising task inhabits, not about a proprietary operator, so a
polynomial attention variant performing well on pixel diffusion is what
the theory predicts. What the operator forfeits is the property that
motivated attention in the first place. The convex-combination bound
that Proposition~\ref{prop:normalized-attention} turns against
normalized attention is also what keeps its outputs inside the hull of
the values; dropping the denominator to gain polynomiality gives up that
guarantee, which is the known stability difficulty with unnormalized
attention.

That difficulty follows from leaving the normalized class \emph{without
replacing the bound}, not from leaving it. The softmax discharges two
roles at once: it makes the mixing weights a convex combination, which is
the hypothesis Proposition~\ref{prop:normalized-attention} turns against
it, and it confines the output to the hull of the values, which is what
keeps activations from growing with sequence length. The two are
separable, and the block of Section~\ref{sec:arch:block} keeps the second
while discarding the first, by moving the bound off the mixing weights
and onto the mixer's input. The RMSNorm preceding each mixer fixes the
norm of its output at $\sqrt{d}$, and the modulation
$u = \gamma \odot \mathrm{RMSNorm}(\cdot) + \beta$ that follows keeps
$\norm{u} \le b := \norm{\gamma}_\infty\sqrt{d} + \norm{\beta}$, where
$\gamma$ and $\beta$ depend on $(t, c)$ alone. The bilinear term
therefore obeys $\norm{W^{(2)}(u,u)} \le b^2\,\norm{W^{(2)}}$, a bound in
the parameters and the conditioning with no dependence on the data or
on $N$. Unnormalized
attention discards both roles together: its weights stay data-dependent
and are no longer confined, so nothing bounds the layer, and the output
grows with the $N$ unnormalized terms in its sum.

The derivatives separate the two as well. The Jacobian of a polynomial
mixer, $W^{(1)} + W^{(2)}(u, \cdot) + W^{(2)}(\cdot, u)$, which is
$W^{(1)} + 2W^{(2)}(u, \cdot)$ when $W^{(2)}$ is symmetric, is affine in
$u$, with no
saturating and no diverging regime; the mixer contains neither an
exponential nor a reciprocal, so neither the gradient growth that
accompanies large logits in exponential attention nor the vanishing
gradient of a saturated softmax has an analogue here. The one
unnormalized attention variant trained at scale corroborates the
separation: the ReLU attention of \citet{wortsman2023relu} reintroduces
an explicit factor $1/N$ in front of the scores. That factor is
data-independent, so the weights still do not sum to one and
Proposition~\ref{prop:normalized-attention} still does not apply; what it
restores is boundedness in sequence length, not convexity. It is the same
split we make, arrived at from the other side.

Being inside the polynomial class is necessary, not sufficient, and
four things separate the architecture of Section~\ref{sec:architecture}
from unnormalized attention within that class. First, its kernels are
the ones the theory names: by~\eqref{eq:theory:cumulants} the order-$n$
Volterra kernel of the optimal denoiser is the $(n{+}1)$-st cumulant of
the reweighted data measure $\tilde{\rho}$, and $W^{(2)}_{\mathrm{tok}}$ is a free low-rank kernel of
exactly that order and type, learned rather than set to the cumulant
value, whereas the cubic tensor of unnormalized attention
is constrained to the query--key--value form and realizes a structured
slice of it. Second, degree is a per-layer choice: the
project--multiply--mix primitive delivers order $k$ from a single layer
through the $k$-fold product, so degree and depth are independent knobs,
while attention's degree is pinned at three by its own factorization.
Third, the mixing kernels $W^{(1)}_{\mathrm{tok}}$ and
$W^{(2)}_{\mathrm{tok}}$ are indexed by absolute
token position rather than derived from content, so spatial structure on
a fixed image grid is native to the operator rather than supplied only
by positional encodings or by convolutional components reintroduced
alongside it. Finally, Proposition~\ref{prop:separation}(i) certifies
that our class contains $\mathcal{P}_{K_L}$ exactly at finite budget; the
corresponding question for unnormalized attention is open, and we claim
no answer to it.

\subsection{Normalization and conditioning in the \ourmodel{} block}
\label{app:extended:blocknorm}

Our own use of normalization is of a different kind, and bears on
Proposition~\ref{prop:separation}(i) in a way worth stating precisely.
The RMSNorm preceding each mixer rescales a token by a positive scalar,
$u \mapsto c(u)\, u$. Were $c$ constant the kernels would absorb it
outright: $W^{(1)}$ and $W^{(2)}$ are learned independently, so the
linear term takes $c$ and the quadratic $c^2$ at no cost to the class.
Being input-dependent, $c$ instead divides each homogeneous component by
the matching power of the token norm, so what the trained network
realizes is a degree-graded reweighting of the polynomial rather than an
unstructured departure from it. We do not claim that the reweighted class
still contains $\mathcal{P}_{K_L}$: recovering the discarded radial
dependence from the residual stream is plausible, and we have not shown
it. The rate of Proposition~\ref{prop:manifold-approx} is untouched
either way, being a statement about the target denoiser rather than about
the network that approximates it.

The conditioning branch is the block's other non-polynomial component,
and it is harmless for a different reason. At fixed $(t, c)$ it emits
constants, so modulation acts on the data path as a diagonal affine map
and the branch's own nonlinearities never reach the data; this is the
regime in which Propositions~\ref{prop:manifold-approx}
and~\ref{prop:separation} are stated, so the fixed-$t$ form of the theory
and the harmlessness of the branch are one fact rather than two
concessions. That the kernels vary with $t$ at all is required rather
than tolerated, since by~\eqref{eq:theory:cumulants} they are cumulants
of a $t$-dependent tilt. A conditioning branch that is itself polynomial
in $t$ is available and we do not use one. Nothing makes the dependence
on $t$ polynomial, and Lemma~\ref{lem:concentration} points the other
way, its binding quantities behaving like $1/t$ and
$\sigma^2/(t^3\tau^2)$, which a low-degree polynomial in $t$ models
poorly where the schedule spends its resolution. This is distinct from
letting $t$ gate the Volterra kernels directly, which changes how the
conditioning acts on the data path rather than how it is computed, and
which Section~\ref{sec:conclusion} lists as a next step. Keeping the branch identical to the baseline's is also what
makes every comparison we report isolate the mixing operator.

\subsection{The feed-forward objection}
\label{app:ffn}

A Transformer block is not attention alone. Its feed-forward sublayer applies a pointwise
nonlinearity, and a block so equipped is a universal approximator that
certainly approximates $f_t$. We first take a single attention head
with attention matrix $A$, held fixed at the point of expansion, and
return to several heads below. Expanding the nonlinearity does produce
cross-token polynomial terms, since the $k$-th power of
$W_1 \sum_j A_{ij} x_j$ contains every degree-$k$ monomial across
tokens.\footnote{The expansion presumes an activation with a
nonvanishing Taylor series, which covers GELU and SiLU. For ReLU there
is none, and the conclusion is more direct: the sublayer is then
piecewise affine and realizes no polynomial of degree $\ge 2$ exactly on
any open set.} What it does not produce is a \emph{free} parametrization
of them, and Table~\ref{tab:kernels} sets the two operators side by
side. The constraint worth drawing out is that width does not relax the
others: the token factor $A_{i\cdot}^{\otimes k}$ carries no hidden-unit
index, so it survives the sum over hidden units untouched and extra
width buys rank only in the channel factor. Only depth, composing blocks
with different $A$, changes the token-mixing structure at all. With $n_{\mathrm{h}}$
heads the argument of the activation becomes
$\sum_{e=1}^{n_{\mathrm{h}}} W_1 W_O^{e} W_V^{e} \sum_j A^{e}_{ij} x_j$, so each slot
of the order-$k$ term draws its token factor from the same $n_{\mathrm{h}}$ matrices
$A^{1}, \dots, A^{n_{\mathrm{h}}}$. The token rank per slot is then at most $n_{\mathrm{h}}$
rather than one, the coefficients across orders are still fixed by the
activation, and width still adds no token structure, which stays capped
by the number of heads. A further constraint is recorded in
Remark~\ref{rem:ffn-rigidity}. The
claim is therefore not that a Transformer cannot approximate $f_t$,
which it can, but that its \emph{mixing} operator contributes no
polynomial content beyond affine, so the polynomial structure of the
target must be recovered through the pointwise nonlinearity, where the
orders are coupled rather than free.

\begin{table}[t]
\centering
\small
\caption{\textbf{What each operator can parametrize at order $k$.} The
feed-forward sublayer does produce cross-token polynomial terms, but as
a $k$-fold power of the attention matrix, scaled by a coefficient the
architecture cannot choose. Shown for one head. With $n_{\mathrm{h}}$ heads each slot
draws from the same $n_{\mathrm{h}}$ matrices.}
\label{tab:kernels}
\begin{tabularx}{\linewidth}{@{}l>{\raggedright\arraybackslash}X>{\raggedright\arraybackslash}X@{}}
\toprule
At order $k$ & Transformer block (one head), smooth $\varphi$ & Low-rank Volterra layer \\
\midrule
Kernel
  & $c_k \big[\sum_a (W_2)_{ca} (W_1)_{a\cdot}^{\otimes k}\big] \otimes A_{i\cdot}^{\otimes k}$
  & $W^{(k)}$ of~\eqref{eq:arch:lowrank}, rank at most $hR$ \\
Slots
  & the same $A$ in every slot; token rank one ($\le n_{\mathrm{h}}$ with $n_{\mathrm{h}}$ heads)
  & independent $Q^{(m)}_q$ per slot \\
Across orders
  & one shared $W_1 A$; $c_k$ fixed by $\varphi$
  & independent kernels \\
Wider sublayer
  & buys channel rank, never token rank
  & --- \\
\bottomrule
\end{tabularx}
\end{table}

\begin{remark}[A further constraint on the feed-forward kernel]
\label{rem:ffn-rigidity}
The effective order-$k$ kernel of a feed-forward sublayer after a single head,
$c_k \big[\sum_a (W_2)_{ca} (W_1)_{a\cdot}^{\otimes k}\big] \otimes
A_{i\cdot}^{\otimes k}$, retains one genuine freedom beyond those noted
above, and it is worth pricing because it is the strongest version of
the objection. Rescaling
$W_1 \mapsto \lambda W_1$ sends the order-$k$ term to
$c_k \lambda^{k}$, so the network can shift weight between orders after
all. But it does so along the fixed one-parameter curve
$\{(c_k \lambda^{k})_{k} : \lambda > 0\}$, whose shape is determined by
the activation. A degree-$K$ Volterra layer chooses $K$ kernels
independently. The comparison is one degree of freedom against $K$, not
none against $K$.
\end{remark}

\section{Proofs for Section~\ref{sec:theory}}
\label{app:proof}

Throughout this appendix, $t \in (0,1)$ is the flow time and
$\sigma := 1 - t \in (0,1)$ the corresponding noise level;
$\manifold \subset \mathbb{R}^D$ is the compact data manifold inside the
ambient pixel space of dimension $D$, and $\rho := \rho_{\mathrm{data}}$
is the data distribution, a probability measure supported on $\manifold$.
For a noised observation $z \in \mathbb{R}^D$ and a clean image
$x \in \manifold$, the \emph{Gaussian likelihood weight}
\[
\psi(z,x) := \exp\!\big(-\norm{z - tx}^2 / 2\sigma^2\big)
\]
is the relative weight the observation $z$ assigns to the clean point $x$
at noise level $\sigma$. The optimal denoiser is the posterior-weighted
average of $x$,
\[
f_t(z) \;=\; \frac{Y(z)}{Z(z)}, \qquad
Y(z) := \int_{\manifold} x\, \psi(z,x)\, d\rho(x), \qquad
Z(z) := \int_{\manifold} \psi(z,x)\, d\rho(x),
\]
where $Z(z)$ is the normalizer (partition function) and $Y(z)$ the
weight-averaged numerator. We write $p(x \mid z) \propto \psi(z,x)\,\rho(x)$
for the posterior distribution over clean images given $z$, and
$\mathbb{E}_z[\,\cdot\,]$ and $\Cov_z(\cdot)$ for its mean and covariance,
so that $f_t(z) = \mathbb{E}_z[x]$.

\subsection{Regularity (Proposition~\ref{prop:manifold-approx}(a))}

We prove that $f_t$ is \emph{real-analytic} on $\mathbb{R}^D$, that is,
locally equal to a convergent power series, and in particular
$C^\infty$, with derivatives of every order. The strategy is standard: a
function that extends to a holomorphic (complex-differentiable) function
of a complex argument $z \in \mathbb{C}^D$ is real-analytic on the real
slice $\mathbb{R}^D$, so it suffices to show that the numerator $Y$ and
the normalizer $Z$ each admit such an extension, and then to divide.

\begin{proof}
\emph{Step 1 (the normalizer is strictly positive).} For every $z$ the
integrand $\psi(z,x)$ is strictly positive, an exponential never vanishing,
and $\rho$ is a probability measure, so
$Z(z) = \int_{\manifold} \psi(z,x)\, d\rho(x) > 0$. The ratio $f_t = Y/Z$ is
therefore well-defined for every $z$.

\emph{Step 2 (each weight extends to an entire function).} Fix
$x \in \manifold$. Viewed as a function of $z$, the weight
$\psi(z,x) = \exp(-\norm{z - tx}^2 / 2\sigma^2)$ is the exponential of a
polynomial in $z$, so it extends from $\mathbb{R}^D$ to an entire
function of $z \in \mathbb{C}^D$, one holomorphic on all of
$\mathbb{C}^D$.

\emph{Step 3 (bounds uniform in $x$).} For a multi-index
$\alpha = (\alpha_1, \dots, \alpha_D) \in \mathbb{Z}_{\geq 0}^D$, a tuple
of non-negative integers recording how many times to differentiate in
each coordinate, write
$\partial_z^\alpha := \partial_{z_1}^{\alpha_1} \cdots
\partial_{z_D}^{\alpha_D}$ for the corresponding mixed partial derivative
in $z$, of total order $|\alpha| := \alpha_1 + \dots + \alpha_D$. On any
compact set $\Omega \subset \mathbb{C}^D$, both $|\psi(z,x)|$ and every
derivative $|\partial_z^\alpha \psi(z,x)|$ are bounded by a constant that
does not depend on $x$. Indeed, differentiating $\psi$ in $z$ brings down,
by the chain rule, a fixed polynomial in $(z - tx)/\sigma$ multiplying
$\psi$ itself; as $x$ ranges over the compact manifold $\manifold$ and $z$
over the compact set $\Omega$, both that polynomial factor and $\psi$ remain
bounded, uniformly in $x$.

\emph{Step 4 (holomorphy passes through the integral).} The integrands
are holomorphic in $z$ with the locally uniform, $x$-independent bounds
of Step 3, so dominated convergence justifies differentiating under the
integral sign: the difference quotients of $Y$ and $Z$ converge to the
integrals of the corresponding $z$-derivatives of $\psi$. Hence $Y$ and $Z$
are holomorphic on $\mathbb{C}^D$, and therefore real-analytic on
$\mathbb{R}^D$.

\emph{Step 5 (the quotient).} Since $Z$ is real-analytic and nowhere zero
(Step 1), the quotient $f_t = Y/Z$ is real-analytic on $\mathbb{R}^D$. In
particular $f_t$ is $C^\infty$, with derivatives of every order, so the
smoothness order $s$ in Proposition~\ref{prop:manifold-approx}(b) may be
taken arbitrarily large.
\end{proof}

\begin{remark}
The smoothness comes from the noise, not from the data: adding noise at
level $\sigma > 0$ convolves the data distribution with a Gaussian, an
infinitely smoothing operation, so $f_t$ is real-analytic for every
$t \in (0,1)$ however irregular $\rho_{\mathrm{data}}$ may be. The
regularity of $f_t$ is thus never the binding constraint in the rate of
Proposition~\ref{prop:manifold-approx}(b); the polynomial degree one is
willing to spend is.
\end{remark}

\subsection{The Jacobian identity (right of~\eqref{eq:theory:mmse})}

\begin{proof}
From $\nabla_z \psi = -\sigma^{-2}(z - tx)\,\psi$,
\begin{align*}
\nabla_z Z &= -\sigma^{-2}\big(z Z - t Y\big), \\
\nabla_z Y &= -\sigma^{-2}\!\int_{\manifold} x\,(z - tx)^\top \psi\, d\rho
= -\sigma^{-2}\big(\mathbb{E}_z[x]\,z^\top - t\,\mathbb{E}_z[xx^\top]\big) Z.
\end{align*}
Using $\nabla_z f_t = \nabla_z Y / Z - (Y/Z)\,(\nabla_z Z / Z)^\top$ and
$Y/Z = \mathbb{E}_z[x]$, the two terms are
\begin{align*}
\nabla_z Y / Z &= -\sigma^{-2}\big(\mathbb{E}_z[x]\,z^\top - t\,\mathbb{E}_z[xx^\top]\big),\\
(Y/Z)(\nabla_z Z/Z)^\top &= -\sigma^{-2}\big(\mathbb{E}_z[x]\,z^\top - t\,\mathbb{E}_z[x]\mathbb{E}_z[x]^\top\big).
\end{align*}
Subtracting, the $\mathbb{E}_z[x]\,z^\top$ terms cancel and
\[
\nabla_z f_t(z) = \frac{t}{\sigma^2}\big(\mathbb{E}_z[xx^\top] - \mathbb{E}_z[x]\mathbb{E}_z[x]^\top\big)
= \frac{t}{\sigma^2}\,\Cov_z(x). \qedhere
\]
\end{proof}

\noindent\emph{The all-order identity~\eqref{eq:theory:tweedie}.}
By~\eqref{eq:theory:cgf}, $f_t(z) = \nabla_\theta \Lambda(\theta)$ at
$\theta = tz/\sigma^2$, so the chain rule gives
\[
\nabla_z^{k} f_t(z)
\;=\; \Big(\tfrac{t}{\sigma^2}\Big)^{k}\,
\nabla_\theta^{\,k+1} \Lambda(\theta)\Big|_{\theta = tz/\sigma^2}.
\]
The derivatives of a cumulant generating function at parameter $\theta$
are the cumulants of the measure tilted by $e^{\theta^\top x}$, and
tilting $\tilde{\rho}$ by $e^{\theta^\top x}$ with
$\theta = tz/\sigma^2$ reconstitutes exactly the posterior: by the
derivation of~\eqref{eq:theory:cgf}, the posterior density is
proportional to $e^{\theta^\top x}\, d\tilde{\rho}(x)$, the
$z$-only Gaussian factor having cancelled between numerator and
denominator. Hence
$\nabla_z^{k} f_t(z) = (t/\sigma^2)^{k}\,
\kappa_{k+1}(p(\cdot \mid z_t = z))$, which
is~\eqref{eq:theory:tweedie}; at $k = 1$ the second cumulant is the
covariance, recovering the computation above. \hfill$\square$

\subsection{Approximation (Proposition~\ref{prop:manifold-approx}(b))}

We prove that the best polynomial of total degree at most $K$
approximates $f_t$ on $t\manifold$ with squared error of order $K^{-2s}$,
equivalently $N^{-2s/(2\dM+1)}$ in the number $N$ of coefficients per
output coordinate, with every constant free of the ambient dimension $D$. Here
\[
\norm{h}_{C^s(\Omega)} \;:=\; \max_{|\alpha| \le s}\ \sup_{w \in \Omega}\
\big|\partial^\alpha h(w)\big|
\]
is the $C^s$ norm, the largest size attained by a function $h$ and its
partial derivatives up to order $s$ over a domain $\Omega$ (with $\alpha$
a multi-index, as in the regularity proof above). The argument has three
stages: embed $t\manifold$ into a space whose dimension is set by $\dM$,
run a classical polynomial approximation rate there, and convert the
degree $K$ into a parameter count.

\begin{proof}
\emph{Step 1 (linear embedding into the intrinsic dimension).} Since
$t\manifold$ is a compact $C^\infty$ manifold of dimension $\dM$,
Whitney's projection theorem provides a residual (hence dense) set of
linear maps $\mathbb{R}^D \to \mathbb{R}^{\dbot}$ with $\dbot = 2\dM + 1$
whose restriction to $t\manifold$ is an injective immersion, and
therefore, by compactness, a smooth embedding; the value $2\dM + 1$ is
the least dimension at which a generic linear projection avoids the
double points of $t\manifold$, whose offending directions form a set of
dimension $2\dM + 1$. Fix such a map $\pi$ (the bottleneck of
Section~\ref{sec:architecture}). Its image $\manifold_\pi := \pi(t\manifold)$ is a
compact $C^\infty$ submanifold of $\mathbb{R}^{\dbot}$ of dimension
$\dM$, and $\pi|_{t\manifold}$ is a diffeomorphism onto $\manifold_\pi$ that is
\emph{bi-Lipschitz}: for some $L_\pi \ge 1$,
\[
L_\pi^{-1}\,\norm{z - z'} \;\le\; \norm{\pi(z) - \pi(z')} \;\le\;
L_\pi\,\norm{z - z'}, \qquad z, z' \in t\manifold,
\]
with $L_\pi$, like every constant below, set by the reach and curvature of
$t\manifold$ and by $\pi$, not by $D$. Because $\pi|_{t\manifold}$ is
injective, each $w \in \manifold_\pi$ has a unique preimage, so
\[
g \;:=\; f_t \circ (\pi|_{t\manifold})^{-1} \;:\; \manifold_\pi \to \mathbb{R}^D
\]
is well-defined; it is the target the polynomial must match on $\manifold_\pi$,
the clean answer $f_t(z)$ indexed by the projected point $\pi(z)$. As a
composition of the real-analytic $f_t$ (part~(a)) with the smooth inverse
$(\pi|_{t\manifold})^{-1}$, $g$ is $C^\infty$, and the chain rule through
this fixed smooth map on a compact set gives
\[
\norm{g}_{C^s(\manifold_\pi)} \;\le\; c_1\,\norm{f_t}_{C^s(t\manifold)},
\]
the constant $c_1$ depending on the derivatives of
$(\pi|_{t\manifold})^{-1}$, hence on $L_\pi$ and the curvature of $t\manifold$,
not on $D$. Approximating $f_t$ on $t\manifold$ is thereby reduced to
approximating the equally smooth $g$ on the $\dM$-dimensional $\manifold_\pi$.

\emph{Step 2 (polynomial rate by Jackson's theorem).} A Jackson-type
theorem on a compact submanifold of Euclidean space is not new:
\citet{ragozin1970polynomial} relates the smoothness of a function on
such a submanifold to its rate of best polynomial approximation, with a
constructive companion for spheres and projective spaces
\citep{ragozin1971constructive}. We give the argument here in the form
we need, via Whitney extension to a cube, because the textbook
statement of Jackson's theorem is on a solid domain and the extension
route keeps every constant explicitly attached to $\dbot$; the step
should be read as a rederivation for self-containedness, not as a new
result. What it does \emph{not} make redundant is Step~1: the constants
in any such theorem track the dimension of the ambient space in which
the submanifold sits, so applying one directly to
$t\manifold \subset \mathbb{R}^{D}$ would reintroduce $D$ into the
bound. The projection into $\mathbb{R}^{2\dM+1}$ is what makes a
Jackson-type rate usable here, and it is the load-bearing step of the
proof.

Treat $g$ one
output coordinate at a time; fix $a \in \{1, \dots, D\}$, a scalar $g^a$
on $\manifold_\pi$. The Whitney extension theorem produces a compactly supported
$G^a \in C^s(\mathbb{R}^{\dbot})$ with $G^a|_{\manifold_\pi} = g^a$ and
$\norm{G^a}_{C^s(\mathbb{R}^{\dbot})} \le c_2\,\norm{g^a}_{C^s(\manifold_\pi)}$,
extending the target from the curved $\manifold_\pi$ to the flat space on which a
polynomial is defined. The multivariate Jackson theorem on a cube
$\Omega \supset \operatorname{supp} G^a$ then yields a polynomial $P_K^a$ of
total degree at most $K$ with
\[
\sup_{w \in \Omega} \big|G^a - P_K^a\big| \;\le\; c_3(\dbot)\, K^{-s}\,
\norm{G^a}_{C^s},
\]
the classical statement that a $C^s$ function on a $\dbot$-dimensional
cube is approximated to order $K^{-s}$ by polynomials of degree $K$, the
constant depending on the cube dimension $\dbot$, not on $D$. Assembling
$P_K = (P_K^1, \dots, P_K^D)$ and restricting to $\manifold_\pi$,
\[
\sup_{w \in \manifold_\pi} \norm{g(w) - P_K(w)}^2
\;\le\; \sum_{a=1}^{D} \big(c_3 K^{-s} \norm{G^a}_{C^s}\big)^2
\;\le\; c_4\, K^{-2s}\, \norm{g}_{C^s(\manifold_\pi, \mathbb{R}^D)}^2,
\]
where squaring $K^{-s}$ reflects the squared error and the output
dimension $D$ enters \emph{only} through
$\norm{g}_{C^s}^2 = \sum_a \norm{g^a}_{C^s}^2$, a fixed quantity for
fixed $(t\manifold, f_t)$ that touches neither the exponent nor the rate.

\emph{Step 3 (assembly, and the parameter form of the rate).} For
$z \in t\manifold$ we have $(\pi|_{t\manifold})^{-1}(\pi(z)) = z$, so
$f_t(z) = g(\pi(z))$ and $P_K(\pi(z)) = (P_K \circ \pi)(z)$; the inverse
cancels and is never evaluated. Hence
\[
\sup_{z \in t\manifold} \norm{f_t(z) - P_K(\pi(z))}^2
\;=\; \sup_{w \in \manifold_\pi} \norm{g(w) - P_K(w)}^2
\;\le\; C\, K^{-2s}, \qquad C = c_4\,\norm{g}_{C^s}^2,
\]
the degree form of the rate. For the parameter form, a polynomial of
total degree at most $K$ on $\mathbb{R}^{\dbot}$ has
$N = \binom{\dbot+K}{\dbot} \asymp K^{\dbot}$ coefficients per output
coordinate, so with $\dbot = 2\dM + 1$, substituting
$K \asymp N^{1/(2\dM+1)}$ gives
\[
K^{-2s} \;\asymp\; N^{-2s/(2\dM+1)}.
\]
The exponent is set by $\dM$ and not by $D$, whereas a polynomial of the
raw patch would need $\binom{D+K}{D} \asymp K^{D}$ coefficients. The
dimension of the restrictions $P|_{\manifold_\pi}$ can be smaller than $N$
when $\manifold_\pi$ lies on a lower-dimensional algebraic variety, but
for a generic smooth manifold no nonzero polynomial vanishes on
$\manifold_\pi$ and it equals $N$, so the count above is the one we use.
By part~(a) the smoothness order $s$ is unbounded, so the rate exceeds
any fixed polynomial order in $K$. Every constant
($c_1, \dots, c_4$ and $C$) tracks $\dbot = 2\dM + 1$, the reach and
curvature of $t\manifold$, and $\norm{f_t}_{C^s}$; none carries an
explicit $D$, which is the precise sense in which the rate escapes the
ambient dimension.
\end{proof}

\begin{remark}
The polynomial $P_K$ above is produced by Whitney extension and
approximation on a cube, not by truncating the cumulant expansion
of~\eqref{eq:theory:cumulants}; the proposition does not assert that the
truncation attains the rate, and the network never computes it. Wiener's
orthogonalization, which does make the degree-$K$ truncation optimal, is
constructed against a Gaussian input measure, whereas $z_t$ here is a
Gaussian smoothing of a manifold-supported law. Neither property is
needed: what is claimed is the existence of \emph{some} degree-$K$
polynomial attaining squared error $O(K^{-2s})$ uniformly on
$t\manifold$, a sup-norm statement rather than an $L^2$ one, and
Proposition~\ref{prop:separation}(i) makes the existential form
sufficient by showing the architecture contains the whole class.
\end{remark}

\begin{remark}
\label{rem:dbot-width}
The embedding dimension $\dbot = 2\dM + 1$ is the least value at which a
generic linear projection is guaranteed faithful, and is the dimension
used in the bound, not a prescription for the trained width. Two points
relate it to the architecture, and they have different status. First,
the trained bottleneck should in fact exceed $2\dM + 1$: at intermediate
noise levels the posterior $p(x \mid z)$ spreads over a curved patch of
$\manifold$, so the embedding must cover the span of tangent spaces
across that patch rather than the manifold at a single point
(Appendix~\ref{app:extended:schedule}),
and this surplus is still governed by $\dM$ and the curvature of
$\manifold$, never by $D$. Second, the practical width
$\dbot \in \{128, 256\}$ lies below the ambient dimension by
construction (here $D = \Dp \in \{768, 3072\}$, a factor of six to
twenty-four). Whether it
also exceeds $2\dM + 1$ depends on the intrinsic dimension $\dM$, which we
do not measure; under the manifold hypothesis, intrinsic-dimension
estimates place natural images at a few dozen dimensions
\citep{pope2021intrinsic, levina2004maximum}, comfortably within the
chosen $\dbot$, but this is an empirical premise rather than a quantity
established here.
\end{remark}

\subsection{Low-dimensional dependence (Lemma~\ref{lem:concentration})}
\label{app:lemma}

The lemma uses two pieces of geometry. The \emph{reach} $\tau > 0$ of
$\manifold$ is the largest radius such that every point within distance
$\tau$ of $\manifold$ has a unique nearest point on $\manifold$; it
bounds the curvature, the second fundamental form satisfying
$\norm{\mathrm{II}} \le 1/\tau$. The scaled manifold $t\manifold$ then
has reach $t\tau$.

\begin{lemma}[Low-dimensional dependence of the denoiser]
\label{lem:concentration}
Let $\manifold$ have reach $\tau > 0$, let $\rho_{\mathrm{data}}$ have a
density $\rho$ with respect to the volume measure of $\manifold$ that is
$C^3$ and bounded below by some $\rho_{\min} > 0$, and fix
$t \in (0,1)$, $\sigma = 1-t$. On the tube
$\mathcal{T}_\delta = \{ z : \operatorname{dist}(z, t\manifold) \le \delta \}$
with $\delta < t\tau$, let $r(z) := t x^\star(z)$ denote the
nearest-point retraction onto the scaled manifold $t\manifold$.

\emph{(a) (Normal insensitivity.)}
\begin{equation*}
\sup_{z \in \mathcal{T}_\delta} \norm{f_t(z) - f_t\!\big(r(z)\big)}
\;\leq\; \eta(t, \sigma, \tau)\,\delta,
\qquad
\eta(t, \sigma, \tau) \;=\; C_\eta \min\Big\{
\underbrace{\tfrac{1}{t}}_{\text{uniform}},\;
\underbrace{\tfrac{\sigma^2}{t^3 \tau^2}}_{\text{small-noise}} \Big\},
\end{equation*}
valid for every $t \in (0,1)$, with $C_\eta$ depending on $\manifold$,
$\rho_{\min}$ and the $C^3$ norm of $\log\rho$, but not on $\sigma$,
$\delta$, or $D$; the
small-noise branch is the active one once $\sigma \ll t\tau$,
where $\eta = O(\sigma^2/(t^3\tau^2))$.

\emph{(b) (Linear compression.)} Let $\pi$ be the projection of
Proposition~\ref{prop:manifold-approx}, let $\tau_\pi > 0$ be the reach of
$\pi(\manifold)$, and suppose in addition that
$\norm{\pi}\,\delta \le t\tau_\pi/2$. Then there is a map
$F : \mathbb{R}^{\dbot} \to \mathbb{R}^D$ with
\begin{equation*}
\sup_{z \in \mathcal{T}_\delta} \norm{f_t(z) - F\!\big(\pi(z)\big)}
\;\leq\; \Big( \eta(t, \sigma, \tau) + \tfrac{c_\pi}{t} \Big)\,\delta,
\end{equation*}
where $c_\pi \geq 0$ is a constant of the pair $(\pi, \manifold)$,
independent of $\sigma$, $\delta$, and $D$. The retraction $r$ is not a
function of the linear image $\pi(z)$, and the $c_\pi/t$ term prices the
exchange: it reflects the tangential component that vectors of
$\ker \pi$ carry, vanishes when $\ker \pi$ is orthogonal to every
tangent space of $t\manifold$, and cannot be removed by any choice of
$F$ (Remark~\ref{rem:cpi-sharp}).
\end{lemma}

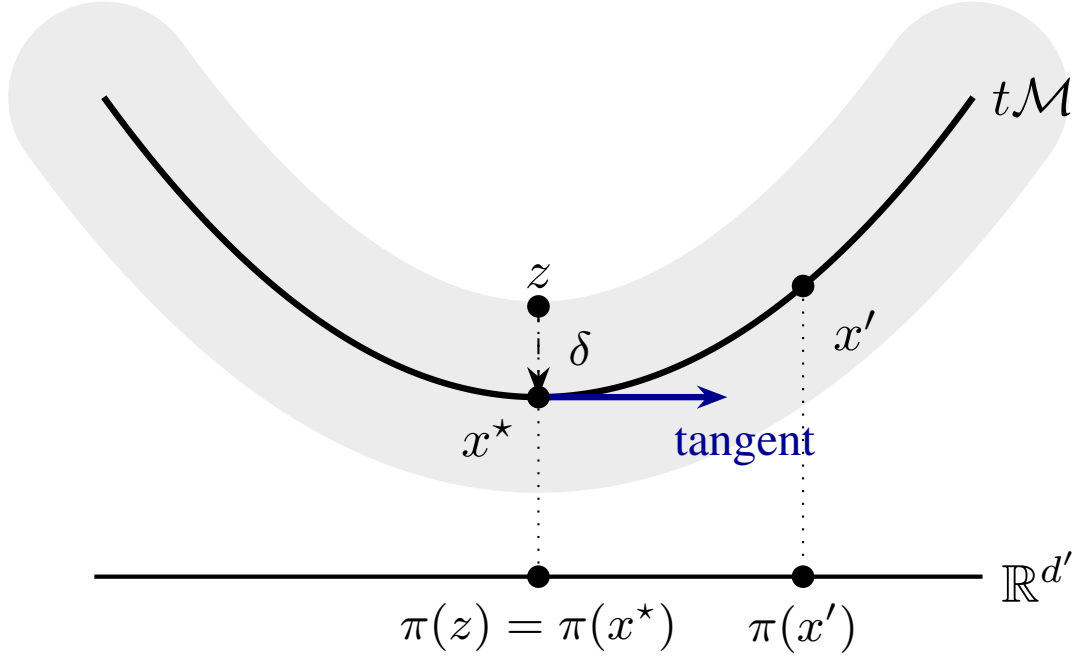
\begin{figure}[t]
\centering
\begin{tikzpicture}[>={Stealth[length=2mm]}, scale=1.25]
  \draw[line width=36pt, gray!15, line cap=round]
        plot[smooth, domain=-2.3:2.3, samples=80] (\x, {0.30*\x*\x});
  \draw[very thick]
        plot[smooth, domain=-2.3:2.3, samples=80] (\x, {0.30*\x*\x});
  \node[anchor=west] at (2.30,1.59) {$t\manifold$};

  \coordinate (xs) at (0,0);
  \coordinate (z)  at (0,0.48);
  \coordinate (xp) at (1.40,0.588);

  \draw[dashed,->] (z) -- (xs);
  \node[anchor=west, font=\small] at (0.06,0.26) {$\delta$};

  \draw[->, very thick, blue!55!black] (xs) -- (1.00,0);
  \node[anchor=north west, font=\small, blue!55!black]
        at (0.62,-0.06) {tangent};

  \fill (z)  circle (1.7pt); \node[anchor=south] at (z) {$z$};
  \fill (xs) circle (1.7pt); \node[anchor=north east] at (-0.02,-0.02) {$x^\star$};
  \fill (xp) circle (1.7pt);
  \node[anchor=north west] at (1.46,0.56) {$x'$};

  \def\ay{-0.95}
  \draw[thick] (-2.35,\ay) -- (2.35,\ay);
  \node[anchor=west] at (2.35,\ay) {$\mathbb{R}^{\dbot}$};
  \draw[dotted] (z) -- (0,\ay);
  \draw[dotted] (xp) -- (1.40,\ay);
  \fill (0,\ay) circle (1.7pt);
  \node[anchor=north, font=\small] at (0,{\ay-0.04}) {$\pi(z)=\pi(x^\star)$};
  \fill (1.40,\ay) circle (1.7pt);
  \node[anchor=north, font=\small] at (1.40,{\ay-0.04}) {$\pi(x')$};
\end{tikzpicture}
\caption{\textbf{Why a low-dimensional projection suffices} (schematic).
Noised inputs $z$ lie in a tube of radius $\delta$ about the scaled data
manifold $t\manifold$. The clean signals consistent with $z$ differ by
moving \emph{along} the manifold (tangent direction), not by leaving it,
so the denoiser $f_t$ varies along the tangent direction and is nearly
flat along the normal one. The bottleneck $\pi$ keeps a $\dbot$-dimensional image of the input. It
sends $z$ and $x^\star$ to the same point; this costs little, since
$f_t$ takes nearly the same value at both. Further, it keeps $x^\star$ and $x'$
apart; this is essential, since $f_t$ takes different values at them.}
\label{fig:concentration}
\end{figure}

Intuitively (Figure~\ref{fig:concentration}), the clean signals consistent
with a noised $z$ differ by sliding along $\manifold$ rather than by
leaving it, so $f_t$ reads the along-manifold part of its input and is
insensitive to the normal part. Discarding the normal directions is precisely what the bottleneck does,
so it costs little. The resulting error is the product of two
quantities: how far off the manifold $z$ sits ($\delta$), and how
sensitive $f_t$ is in the discarded directions ($\eta$). The second is
small once the noise drops below the reach. A
linear $\pi$ cannot discard the normal directions exactly, since its
kernel generically carries a tangential component along which $f_t$
varies at the ordinary rate $O(1/t)$; part (b)'s $c_\pi/t$ term prices
this residual slippage, and it shrinks as the bottleneck widens. This
is one reason the trained $\dbot$ is set above the minimal $2\dM + 1$;
the second, posterior coverage, is discussed in
Appendix~\ref{app:extended:schedule}.

The proof makes this precise through a tangent--normal split of the
posterior covariance. Along the manifold the posterior carries the
ordinary spread of order $\sigma^2/t^2$, whereas reaching off the
manifold costs an extra factor of curvature and shrinks the normal
spread by $(\sigma/t\tau)^2$; the Jacobian
identity (right of~\eqref{eq:theory:mmse}) then transfers this gap to
$\nabla_z f_t$. Part (a) bounds the change of $f_t$ under the
nearest-point retraction, and part (b) prices the exchange of that
nonlinear retraction for the linear bottleneck projection $\pi$.

\begin{proof}
\emph{Step 0 (nearest point and unimodality).} Fix
$z \in \mathcal{T}_\delta$ with $\delta < t\tau$. Since
$\operatorname{dist}(z, t\manifold) \le \delta < t\tau$ and $t\manifold$
has reach $t\tau$, the point $z$ has a unique nearest point $tx^\star$ on
$t\manifold$, and the residual $w := z - tx^\star$ is \emph{normal} to
$\manifold$ at $x^\star$ (perpendicular to the surface), with
$\norm{w} \le \delta$. Any competing posterior mode is separated from
$x^\star$ by distance $\ge c\,\tau$ and so carries mass
$O(e^{-c^2 t^2\tau^2 / 2\sigma^2})$; for $\sigma \ll t\tau$ the posterior is
effectively unimodal about $x^\star$ and these tails are absorbed below.

\emph{Step 1 (local model and the Gaussian posterior).} Parametrize
$\manifold$ near $x^\star$ in \emph{tangent coordinates}
$\xi \in \mathbb{R}^{\dM}$, the displacement within the tangent plane
$T_{x^\star}\manifold$ recorded against an orthonormal tangent frame
$B \in \mathbb{R}^{D \times \dM}$ ($B^\top B = I$):
\[
x(\xi) = x^\star + B\xi + \tfrac{1}{2}\mathrm{II}(\xi, \xi) + O(\norm{\xi}^3).
\]
The linear term $B\xi$ moves in the tangent plane, and the \emph{second
fundamental form} $\mathrm{II}$, the symmetric, normal-valued quadratic
form measuring how $\manifold$ curves away from that plane
($\norm{\mathrm{II}} \le 1/\tau$), bends the point back onto the surface.
The posterior is
$p(x \mid z) \propto e^{-\norm{z - tx}^2 / 2\sigma^2}\,\rho(x)$, so its
negative logarithm, restricted to the surface and dropping the
$\xi$-independent normalizer, is
$\Phi(\xi) := \tfrac{1}{2\sigma^2}\norm{z - tx(\xi)}^2 - \log\rho(x(\xi))$.
Writing
$z - tx(\xi) = w - tB\xi - \tfrac{t}{2}\mathrm{II}(\xi,\xi) + O(\norm{\xi}^3)$
and squaring, the cross term $-2t\,w^\top B\xi$ vanishes because $w$ is
normal and $B\xi$ tangent ($w^\top B\xi = 0$), so $\xi = 0$ is a critical
point and the surviving quadratic is
\[
\Phi(\xi) = \mathrm{const} + \tfrac{1}{2\sigma^2}\,\xi^\top\big(
\underbrace{t^2 I}_{\text{tangent stiffness}}
- \underbrace{t\langle w, \mathrm{II}\rangle}_{\text{curvature}\,\times\,\text{offset}}
\big)\xi + O(\norm{\xi}^3),
\]
where $\langle w, \mathrm{II}\rangle$ is the \emph{shape operator} along
$w$, the symmetric form with
$\xi^\top \langle w, \mathrm{II}\rangle\,\xi = w^\top \mathrm{II}(\xi,\xi)$,
satisfying
$\norm{t\langle w, \mathrm{II}\rangle} \le t\norm{w}/\tau \le t\delta/\tau < t^2$.
The Hessian $\tfrac{1}{\sigma^2}(t^2 I - t\langle w, \mathrm{II}\rangle)$
is thus positive definite, of size $\sim t^2/\sigma^2$. Since this
Hessian is of order $\sigma^{-2}$ while the third derivatives of $x(\xi)$
and $\log\rho$ are $O(1)$, the cubic remainder is smaller than the
quadratic by a factor $\sim \sigma/(t\tau)$ on the bulk of the posterior
(where $\norm{\xi} \sim \sigma/t$), so by the Laplace approximation the
posterior over $\xi$ is Gaussian to leading order,
\[
\xi \sim \mathcal{N}(0, \Sigma_\xi), \qquad
\Sigma_\xi = \tfrac{\sigma^2}{t^2}\big(I + O(\delta/(t\tau))\big),
\]
the prior $-\log\rho$ contributing only at the subleading order
$\sigma^2$ absorbed into the correction.

\emph{Step 2 (tangent versus normal covariance).} Decompose
\[
x - \mathbb{E}_z[x] = B(\xi - \bar\xi)
+ \tfrac{1}{2}\big(\mathrm{II}(\xi,\xi) - \mathbb{E}\,\mathrm{II}(\xi,\xi)\big) + \cdots
\]
into its tangent (linear in $\xi$) and normal (quadratic in $\xi$) parts.
The tangent part has covariance $B\Sigma_\xi B^\top$, with eigenvalues
$\tfrac{\sigma^2}{t^2}(1 + O(\delta/(t\tau)))$ on the tangent space
$T_{x^\star}\manifold$. The normal part, being quadratic in $\xi$, has
covariance
$O\big(\norm{\mathrm{II}}^2 \operatorname{Var}\norm{\xi}^2\big)
= O\big(\sigma^4 / (t^4\tau^2)\big)$ on the normal space. The posterior
covariance is therefore of order $\sigma^2/t^2$ along the manifold but
smaller by the factor $(\sigma/t\tau)^2$ across it: the largest
eigenvalue of $\Cov_z(x)$ on directions normal to $\manifold$ is
$O(\sigma^4/(t^4\tau^2))$.

\emph{Step 3 (transfer to the Jacobian; part (a)).} By the Jacobian
identity (right of~\eqref{eq:theory:mmse}),
$\nabla_z f_t = \tfrac{t}{\sigma^2}\Cov_z(x)$, so its operator norm on
directions normal to $t\manifold$ is
$\tfrac{t}{\sigma^2}\cdot O(\sigma^4/(t^4\tau^2)) = O(\sigma^2/(t^3\tau^2))$,
and $O(1/t)$ on tangent directions. The segment from $z$ to its
retraction $r(z) = tx^\star$ has length $\norm{w} \le \delta$ and lies
in the normal direction, so
\[
\norm{f_t(z) - f_t\!\big(r(z)\big)}
\le \delta \cdot \sup_{z' \in \mathcal{T}_\delta}\big\|\nabla_z f_t(z')\big|_{T^\perp}\big\|
= O\!\Big(\tfrac{\sigma^2}{t^3\tau^2}\Big)\,\delta,
\]
which is the small-noise branch of part (a),
$\eta = O(\sigma^2/(t^3\tau^2))$ for $\sigma \ll t\tau$.

\emph{Step 3$'$ (the uniform branch).} The estimate of Steps 1--3 is a
small-noise expansion; a bound valid for every $t \in (0,1)$ with
$\delta < t\tau$ follows by localizing the posterior at scale
$\sigma/t$ without any Gaussian expansion. For a manifold point $x'$ at
chordal distance $d = \norm{x' - x^\star}$, the reach
inequality $\operatorname{dist}(x', T_{x^\star}\manifold) \le
d^2/(2\tau)$ \citep{federer1959curvature} and the normality of $w$ give
$|w^\top(x' - x^\star)| \le \delta\, d^2/(2\tau)$, so
\[
\norm{z - tx'}^2 - \norm{z - tx^\star}^2
\;=\; t^2 d^2 - 2t\, w^\top(x' - x^\star)
\;\ge\; t^2 d^2 \Big(1 - \tfrac{\delta}{t\tau}\Big),
\]
and for $\delta \le (1 - c_0)\, t\tau$ the posterior weight at distance
$d$ is at most $e^{-c_0 t^2 d^2 / 2\sigma^2}$ relative to the peak. The
posterior second moment about $x^\star$ is therefore $O((\sigma/t)^2)$
(the function $d^2 e^{-c_0 t^2 d^2 / 2\sigma^2}$ is maximized at
$d \asymp \sigma/t$), with constants depending on the density bounds
of $\rho_{\mathrm{data}}$ and the geometry of $\manifold$, and since
the covariance about the mean is dominated by the second moment about
any fixed point, $\Cov_z(x) = O((\sigma/t)^2)$ on the tube. By the
Jacobian identity (right of~\eqref{eq:theory:mmse}),
$\norm{\nabla_z f_t}
\le \tfrac{t}{\sigma^2} \cdot O\big((\sigma/t)^2\big) = O(1/t)$
in every direction, for every $t$. Combining with Step 3 yields the
form stated in the lemma,
$\eta(t, \sigma, \tau) = C \min\{ 1/t,\ \sigma^2/(t^3\tau^2) \}$.

\emph{Step 4 (linear compression; part (b)).} The retraction $r$ is not
a function of $\pi(z)$: $\pi$ is linear with kernel of dimension
$D - \dbot$, and two tube points on the same fiber $z + \ker\pi$
generally have different nearest points, so the assignment
$\pi(z) \mapsto r(z)$ is ill-defined. Construct $F$ instead through the
projected manifold. Take $\pi$ as in
Proposition~\ref{prop:manifold-approx}, bi-Lipschitz on $t\manifold$
with constant $L_\pi$. Its image
$\pi(t\manifold) = t\,\pi(\manifold)$ is a compact embedded submanifold of
$\mathbb{R}^{\dbot}$ with reach $t\tau_\pi > 0$. A point of
$\pi(\mathcal{T}_\delta)$ lies within $\norm{\pi}\,\delta \le t\tau_\pi/2$
of it, so the nearest-point projection
$\mathrm{np}(v) := \operatorname{arg\,min}_{y \in \pi(t\manifold)} \norm{v - y}$,
the point of $\pi(t\manifold)$ closest to $v$, is well-defined on
$\pi(\mathcal{T}_\delta)$ and Lipschitz there with constant at most $2$
\citep{federer1959curvature}. Set
\[
F(v) \;:=\; f_t\!\Big( \big(\pi|_{t\manifold}\big)^{-1}\!\big(\mathrm{np}(v)\big) \Big),
\qquad v \in \pi(\mathcal{T}_\delta).
\]
For $z \in \mathcal{T}_\delta$ write $\pi(z) = \pi(r(z)) + \pi(w)$ with
$\norm{\pi(w)} \le \norm{\pi}\,\delta$; then $\mathrm{np}(\pi(z))$
differs from $\pi(r(z))$ by $O(\delta)$, and its bi-Lipschitz lift
$t\tilde{x} := (\pi|_{t\manifold})^{-1}(\mathrm{np}(\pi(z)))$ differs
from $r(z)$ by at most $c_\pi\,\delta$ along $t\manifold$, where
$c_\pi$ collects $\norm{\pi}$, $L_\pi$, and the nearest-point Lipschitz
constant, all independent of $\sigma$, $\delta$, and $D$. Hence, using
part (a) and the tangential operator norm $O(1/t)$ from Step 3,
\[
\norm{f_t(z) - F(\pi(z))}
\;\le\; \norm{f_t(z) - f_t\!\big(r(z)\big)}
+ \norm{f_t\!\big(r(z)\big) - f_t(t\tilde{x})}
\;\le\; \eta\,\delta + O\!\Big(\tfrac{1}{t}\Big)\, c_\pi\,\delta,
\]
which is part (b) after absorbing constants into $c_\pi$. When
$\ker\pi$ is orthogonal to every tangent space of $t\manifold$, the
fiber and the slippage are purely normal and the tangential term
vanishes.
\end{proof}

\begin{remark}
\label{rem:cpi-sharp}
The $c_\pi/t$ term of part (b) is not an artifact of this construction.
Any $F$ is constant on the fibers of $\pi$; a fiber segment of length
$\sim \delta$ inside the tube generically carries a tangential component
(vectors of $\ker\pi$ meet the tangent spaces at a positive angle, but
not orthogonally), and along that component $f_t$ varies at the ordinary
tangential rate $O(1/t)$, independently of $\sigma$. Hence
$\sup_{\mathcal{T}_\delta} \norm{f_t - F \circ \pi} \gtrsim \delta/t$
for every $F$ unless $\ker\pi$ lies in the common normal space of
$t\manifold$, which requires $\dbot$ at least the dimension of the span
of the tangent spaces. Part (a) is therefore the sharp statement about
the denoiser itself, and part (b) the honest statement about linear
compression; the gap between them shrinks as the bottleneck widens,
since a larger $\dbot$ leaves a smaller kernel with less tangential
content.
\end{remark}

\begin{remark}[Worst case and typical case]
\label{rem:typical-slippage}
Part (b) is a supremum over the tube. For a noised input
$z = tx + \sigma\epsilon$ the tube radius is set by the component of
$\sigma\epsilon$ normal to $t\manifold$, whose length concentrates at
$\sigma\sqrt{D-\dM}$. Read at that radius, the compression term
of~\eqref{eq:theory:combined} is $O((D-\dM)\,\sigma^2/t^2)$, and the
tube hypothesis $\delta < t\tau$ requires $\sigma\sqrt{D-\dM} < t\tau$.
The supremum is pessimistic for the slippage term, however. Step~4 uses
the normal displacement $w = z - r(z)$ only through $\pi(w)$, bounding
$\norm{\pi(w)} \le \norm{\pi}\,\norm{w}$. To leading order in $\sigma$,
$w = \sigma P_N \epsilon$, with $P_N$ the orthogonal projection onto the
normal space of $t\manifold$ at $tx$, so for isotropic noise
\[
\mathbb{E}\,\norm{\pi(w)}^2
\;=\; \sigma^2 \operatorname{tr}\big(\pi P_N \pi^\top\big)
\;\le\; \sigma^2 \norm{\pi}_F^2
\;\le\; \dbot\, \norm{\pi}^2 \sigma^2,
\]
using that $\pi$ has rank at most $\dbot$. Since
$\epsilon \mapsto \norm{\pi P_N \epsilon}$ is $\norm{\pi}$-Lipschitz,
Gaussian concentration gives
$\norm{\pi(w)} \le \norm{\pi}\,\sigma\,(\sqrt{\dbot} + u)$ with
probability at least $1 - e^{-u^2/2}$. For a typical noised input the
slippage part of the compression cost is therefore of order
$c_\pi^2\, \dbot\, \sigma^2/t^2$, set by the bottleneck width rather than
by the ambient dimension. The normal-sensitivity part $\eta\,\norm{w}$
is a property of $f_t$ itself rather than of $\pi$, and this argument
does not reduce it. We expect it to depend on the curvature of
$\manifold$ rather than on $D$, since the posterior spreads in the normal
directions only through curvature, but we do not prove this.
\end{remark}

\begin{remark}[Where each branch applies]
\label{rem:lemma-branches}
Step~0, the Gaussian posterior of Step~1 and the small-noise branch of
$\eta$ are expansions in $\sigma/(t\tau)$ and hold only late in
sampling, once $\sigma \ll t\tau$. The uniform branch of Step~3$'$ needs
no expansion and holds at every $t$, but only on the tube
$\delta < t\tau$. At intermediate $t$, where $\sigma/t$ is comparable to
$\tau$, the two branches are both of order $1/t$, so for an input inside
the tube the compression term of~\eqref{eq:theory:combined} is at most
of order $\tau^2$, bounded but not small. A typical noised input,
however, lies at distance about $\sigma\sqrt{D-\dM}$ from $t\manifold$
(Remark~\ref{rem:typical-slippage}), which in this regime exceeds
$t\tau$ by a factor of about $\sqrt{D-\dM}$, so the lemma does not reach
it. Appendix~\ref{app:extended:schedule} therefore treats intermediate
$t$ without the lemma, through the width the bottleneck needs
(Remark~\ref{rem:dbot-width}) and the analyticity of $f_t$.
\end{remark}

\noindent\emph{Combined bound~\eqref{eq:theory:combined}.} For
$z \in \mathcal{T}_\delta$, the triangle inequality gives
$\norm{f_t(z) - P_K(\pi(z))} \le \norm{f_t(z) - F(\pi(z))} + \norm{F(\pi(z)) - P_K(\pi(z))}$.
The first term is $\le (\eta + c_\pi/t)\,\delta$ by
Lemma~\ref{lem:concentration}(b). For the second,
$F(\pi(z)) = g(\mathrm{np}(\pi(z)))$ with
$\mathrm{np}(\pi(z)) \in \pi(t\manifold)$,
and $P_K$ approximates $g$ on $\pi(t\manifold)$ to
$O(K^{-s})$ by Proposition~\ref{prop:manifold-approx}(b);
since $P_K$ is a fixed polynomial, hence Lipschitz on the compact
$\pi(\mathcal{T}_\delta)$, evaluating at $\pi(z)$ rather than at
$\mathrm{np}(\pi(z))$ adds an $O(\delta)$ term whose $\sigma$-free
constant joins the slippage inside the second bracket
of~\eqref{eq:theory:combined}, enlarging $c_\pi$. Squaring and using
$(a+b)^2 \le 2a^2 + 2b^2$ yields
\eqref{eq:theory:combined}. \hfill$\square$

\subsection{Attainment (Proposition~\ref{prop:separation}(i))}
\label{app:proof-attainment}

Throughout this subsection set $K := K_L = 2^L$, write
$\Pi_K(\mathbb{R}^{\dbot}; \mathbb{R})$ for the scalar polynomials of total
degree at most $K$, and let
$H := \dim \Pi_K(\mathbb{R}^{\dbot}; \mathbb{R}) = \binom{\dbot + K}{\dbot}$.

\begin{lemma}[Powers of affine forms span]
\label{lem:waring}
For every $K \geq 0$,
\begin{equation*}
\operatorname{span}\big\{\, (a^\top u + b)^K \;:\; a \in \mathbb{R}^{\dbot},\
b \in \mathbb{R} \,\big\} \;=\; \Pi_K(\mathbb{R}^{\dbot}; \mathbb{R}).
\end{equation*}
\end{lemma}

\begin{proof}
The inclusion $\subseteq$ is immediate. For $\supseteq$, fix
$a \in \mathbb{R}^{\dbot}$ and consider, for $\lambda \in \mathbb{R}$,
\begin{equation*}
\varphi_\lambda(u) \;:=\; (\lambda\, a^\top u + 1)^K
\;=\; \sum_{j=0}^{K} \binom{K}{j}\, \lambda^j\, (a^\top u)^j .
\end{equation*}
Evaluating at $K+1$ distinct values $\lambda_0, \dots, \lambda_K$ produces a
linear system for the functions
$\big(\binom{K}{j}(a^\top u)^j\big)_{j=0}^{K}$ whose coefficient matrix
$(\lambda_i^{\,j})_{i,j}$ is Vandermonde, hence invertible. Every pure
power $(a^\top u)^j$ with $j \leq K$ therefore lies in the span of the
family. It remains to verify that
$\{(a^\top u)^j : a \in \mathbb{R}^{\dbot}\}$ spans the homogeneous
polynomials of degree $j$, which is the classical polarization
identity: for any indices $i_1, \dots, i_j \in \{1, \dots, m\}$,
\begin{equation*}
u_{i_1} u_{i_2} \cdots u_{i_j} \;=\; \frac{1}{2^j\, j!}
\sum_{\varepsilon \in \{\pm 1\}^j} \varepsilon_1 \cdots \varepsilon_j\,
\big( \varepsilon_1 u_{i_1} + \dots + \varepsilon_j u_{i_j} \big)^j ,
\end{equation*}
expressing an arbitrary degree-$j$ monomial as a combination of $j$-th
powers of linear forms. Summing over $j \leq K$ completes the proof.
\end{proof}

\begin{lemma}[Exact realization by iterated multiplication]
\label{lem:realization}
Fix $R^\ast \geq 1$, affine forms $\ell_r(u) = q_r^\top u + c_r$ on
$\mathbb{R}^{\dbot}$, and a matrix $C \in \mathbb{R}^{D \times R^\ast}$. A
depth-$L$ stack of the mixers of Section~\ref{sec:architecture}, of
stream width $R^\ast$, channel-mixer hidden width $h = R^\ast$ and rank
$R' = 1$, with gated residual
connections, affine embedding, and linear head, computes the map
\begin{equation*}
u \;\longmapsto\; \sum_{r=1}^{R^\ast} C_{\cdot r}\; \ell_r(u)^{2^L}
\end{equation*}
exactly, with at most $c_0 (R^\ast)^2$ parameters per layer for an
absolute constant $c_0$.
\end{lemma}

\begin{proof}
The construction acts on each patch separately and uses only the channel
mixers. Every token mixer is switched off by setting its gate
$\alpha_{\mathrm{tok}} = 0$, an admissible parameter setting, so each
block reduces to its channel sublayer.
The affine embedding writes $w^{(0)}_r := \ell_r(u)$ into channel $r$
of the stream, $r = 1, \dots, R^\ast$ (the constants $c_r$ are supplied
by the embedding bias). Each subsequent channel mixer consists of a linear
branch and the quadratic primitive of~\eqref{eq:arch:eval},
wrapped in a gated residual connection: on stream state $w$ its output
is
\begin{equation*}
w \;+\; \operatorname{diag}(\alpha)\Big( W^{(1)} w
+ P\,\big[(Q^{(1)}_1 w) \odot (Q^{(2)}_1 w)\big] \Big).
\end{equation*}
Choose $Q^{(1)}_1 = Q^{(2)}_1 = P = I_{R^\ast}$ (the tied setting, which the
untied class contains), $W^{(1)} = -I_{R^\ast}$, and
$\alpha = \mathbf{1}$; the gates are zero-initialized but learnable, so
this is an admissible parameter setting. The output is then
$w + (-w + w \odot w) = w \odot w$: the linear branch cancels the
residual and the layer squares the stream elementwise. After $L$ such
layers channel $r$ holds $\ell_r(u)^{2^L}$, and the linear head applies
$C$. The per-layer parameter count is that of
$P, Q^{(1)}_1, Q^{(2)}_1, W^{(1)} \in \mathbb{R}^{R^\ast \times R^\ast}$ and the gates, in total
$O\big((R^\ast)^2\big)$.
\end{proof}

\begin{proof}[Proof of Proposition~\ref{prop:separation}(i)]
By Lemma~\ref{lem:waring} at degree $K_L$, the family
$\{\ell^{K_L} : \ell \text{ affine}\}$ spans
$\mathcal{P}_{K_L}(\mathbb{R}^{\dbot}; \mathbb{R})$ and therefore contains a
basis $\ell_1^{K_L}, \dots, \ell_H^{K_L}$ with
$H = \binom{\dbot + K_L}{\dbot}$. Any
$Q \in \mathcal{P}_{K_L}(\mathbb{R}^{\dbot}; \mathbb{R}^{D})$ thus has each
output coordinate a linear combination of the $\ell_r^{K_L}$, that is,
$Q(u) = C\,\big(\ell_r(u)^{K_L}\big)_{r=1}^{H}$ for some
$C \in \mathbb{R}^{D \times H}$. Lemma~\ref{lem:realization} with
$R^\ast = H$ realizes this map exactly with at most
$c_0 H^2 =: \Pbud_0(L, \dbot)$ parameters per layer, applied to the
input $\pi(z)$ supplied by the bottleneck embedding. This proves the
exact containment
$\{Q \circ \pi : Q \in \mathcal{P}_{K_L}\} \subseteq \mathcal{V}_{L, \Pbud}$
for $\Pbud \geq \Pbud_0$. The approximation bound follows by
instantiating $g = P_{K_L} \circ \pi$ with $P_{K_L}$ the polynomial of
Proposition~\ref{prop:manifold-approx}(b) at degree $K_L$, whose error
on $t\manifold$ is at most $C_1 K_L^{-2s}$; the tube extension is
~\eqref{eq:theory:combined}.
\end{proof}

\begin{remark}
The statement is representational rather than practical. The threshold
$\Pbud_0 \asymp H^2$ is astronomically large at the depths used in
practice, because the dense factor matrices charge $H$ parameters per
column to operations (carrying a channel, squaring a channel) that
touch one coordinate each, and no attempt has been made to optimize the
constant. A further reason for its size is that the construction
realizes the \emph{full} polynomial space $\mathcal{P}_{K_L}$ on
$\mathbb{R}^{\dbot}$, of dimension
$H = \binom{\dbot + K_L}{\dbot} \asymp K_L^{\dbot}$ (exponent
$\dbot = 2\dM + 1$), the same space in which
Proposition~\ref{prop:manifold-approx} finds its approximants, and it
charges dense factor matrices for every monomial of that space. Its role is to certify that
restricting from unconstrained polynomials to the architecture's
low-rank class loses nothing at the level of the rate: the degree form
$K_L^{-2s}$ of Proposition~\ref{prop:manifold-approx} is attained within
the architecture's own function class. The construction realizes the
target at the tied point $Q^{(1)}_q = Q^{(2)}_q$ of the primitive; untying
the factors, as in the trained architecture, enlarges the
parameterization without changing what is representable, so the threshold
$\Pbud_0$ is unaffected.
Normalization layers are excluded from $\mathcal{V}_{L, \Pbud}$, as
RMSNorm is not a polynomial map. What it changes in the trained network,
a degree-graded reweighting of the polynomial, and why we do not claim
the containment for the normalized class, is set out in
Appendix~\ref{app:extended:blocknorm}.
\end{remark}

\subsection{Polynomial content of softmax attention
(Proposition~\ref{prop:separation}(ii))}
\label{app:proof-attention}

\begin{lemma}[Independence of exponentials over polynomials]
\label{lem:exp-indep}
Let $q_1, \dots, q_n$ be real polynomials on $\mathbb{R}^\nu$ such that
$q_i - q_j$ is nonconstant for every $i \neq j$, and let
$p_1, \dots, p_n$ be polynomials. If
$\sum_{i=1}^{n} p_i(x)\, e^{q_i(x)} = 0$ for all $x$ in a nonempty open
set $U \subseteq \mathbb{R}^\nu$, then $p_i \equiv 0$ for every $i$.
\end{lemma}

\begin{proof}
Both sides are real-analytic, so the identity extends from $U$ to all
of $\mathbb{R}^\nu$. Choose a unit vector $v$ such that the top-degree
homogeneous part of each difference $q_i - q_j$ ($i \neq j$) does not
vanish at $v$; this is possible because finitely many nonzero
homogeneous polynomials vanish only on a closed set of measure zero on
the sphere. Then for every base point $x_0$, the univariate restriction
$\lambda \mapsto (q_i - q_j)(x_0 + \lambda v)$ is a polynomial of the same degree
as $q_i - q_j$ with leading coefficient determined by $v$ alone; in
particular it is nonconstant and tends to $+\infty$ or $-\infty$ as
$\lambda \to +\infty$, with the sign independent of $x_0$. The relation
$i \succ j$ iff $(q_i - q_j)(x_0 + \lambda v) \to +\infty$ is therefore a
strict total order on $\{1, \dots, n\}$, independent of $x_0$; let
$i^\ast$ be its maximum. Dividing the identity restricted to the line
by $e^{q_{i^\ast}(x_0 + \lambda v)}$ gives
\begin{equation*}
p_{i^\ast}(x_0 + \lambda v) \;=\; -\sum_{i \neq i^\ast}
p_i(x_0 + \lambda v)\; e^{(q_i - q_{i^\ast})(x_0 + \lambda v)} ,
\end{equation*}
and each term on the right tends to $0$ as $\lambda \to +\infty$, since the
exponent is a polynomial tending to $-\infty$ and the exponential decay
dominates polynomial growth. A univariate polynomial with limit $0$ at
$+\infty$ vanishes identically, so $p_{i^\ast}(x_0 + \lambda v) \equiv 0$ for
every $x_0$, hence $p_{i^\ast} \equiv 0$ on $\mathbb{R}^\nu$. Removing
the $i^\ast$ term and iterating completes the proof.
\end{proof}

\begin{lemma}[Affine polynomial content of a softmax attention layer]
\label{lem:affine-content}
For a token matrix $X = (x_1, \dots, x_N) \in \mathbb{R}^{N \times d}$,
let
\begin{equation*}
h_i(X) \;=\; x_i + \sum_{e=1}^{n_{\mathrm{h}}} W_O^e\,
\frac{\sum_{j=1}^{N} \exp\big(s^e_{ij}(X)\big)\, W_V^e\, x_j}
     {\sum_{j=1}^{N} \exp\big(s^e_{ij}(X)\big)},
\qquad
s^e_{ij}(X) \;=\; \tfrac{1}{\sqrt{d_h}}\,(W_Q^e x_i)^\top (W_K^e x_j),
\end{equation*}
be a softmax self-attention layer with $n_{\mathrm{h}}$ heads and residual
connection. If $h$ agrees with a polynomial map on a nonempty open set
of inputs, then $h$ is affine.
\end{lemma}

\begin{proof}
Both $h$ and the candidate polynomial are real-analytic, so agreement
on an open set implies agreement everywhere; write
$g_i := h_i - x_i$, a polynomial map, and fix the output token $i$.

Consider first a single head ($n_{\mathrm{h}} = 1$) and drop the head index.
Multiplying through by the (strictly positive) normalizer gives the
global identity
\begin{equation*}
\sum_{j=1}^{N} e^{s_{ij}(X)} \big( W_O W_V\, x_j - g_i(X) \big)
\;\equiv\; 0 .
\end{equation*}
Partition $\{1, \dots, N\}$ into classes by the equivalence
$j \sim j'$ iff $s_{ij} - s_{ij'}$ is constant as a polynomial in $X$,
and on each class $\mathcal{C}$ write $s_{ij} = s_{\mathcal{C}} +
\kappa_j$ with $\kappa_j \in \mathbb{R}$. Representatives of distinct
classes have nonconstant differences, so applying
Lemma~\ref{lem:exp-indep} coordinatewise (with polynomial coefficients
$\sum_{j \in \mathcal{C}} e^{\kappa_j}(W_O W_V x_j - g_i)$) forces, for
every class $\mathcal{C}$,
\begin{equation*}
\sum_{j \in \mathcal{C}} e^{\kappa_j} \big( W_O W_V\, x_j - g_i(X)
\big) \;\equiv\; 0,
\quad\text{i.e.}\quad
g_i(X) \;=\; \sum_{j \in \mathcal{C}} \omega_j\, W_O W_V\, x_j,
\quad
\omega_j = \frac{e^{\kappa_j}}{\sum_{j' \in \mathcal{C}}
e^{\kappa_{j'}}} ,
\end{equation*}
since the weights $e^{\kappa_j}$ are positive and sum to a nonzero
constant. The right-hand side is affine in $X$, so $g_i$ is affine and
$h_i = x_i + g_i$ is affine.

For $n_{\mathrm{h}} \geq 2$ heads, multiply through by the product of all
normalizers $D^e_i(X) := \sum_j e^{s^e_{ij}(X)}$:
\begin{equation*}
g_i(X) \prod_{e=1}^{n_{\mathrm{h}}} D^e_i(X) \;-\; \sum_{e=1}^{n_{\mathrm{h}}}
\Big( \sum_{j} e^{s^e_{ij}(X)}\, W_O^e W_V^e\, x_j \Big)
\prod_{e' \neq e} D^{e'}_i(X) \;\equiv\; 0 .
\end{equation*}
Expanding all products writes the left side as a finite sum
$\sum_\alpha p_\alpha(X)\, e^{\bar{s}_\alpha(X)}$ in which every exponent is
a sum $\bar{s}_\alpha = \sum_e s^e_{i j_e}$ over one score per head, and
every coefficient $p_\alpha$ is either $g_i$ (from the first product,
with positive constant weight) or an affine map $W_O^e W_V^e x_j$
(from the second, with negative constant weight). Group the exponents
into classes modulo additive constants and apply
Lemma~\ref{lem:exp-indep} as before: each class identity has the form
$c\, g_i(X) = (\text{affine map})$ where $c$ is a sum of the positive
constants attached to the $g_i$-terms in that class. Every term of the
expansion of $g_i \prod_e D^e_i$ lies in some class, and within its
class the $g_i$-coefficient $c$ is a nonempty sum of positive numbers,
hence nonzero. Therefore $g_i$ is affine, and so is $h$.
\end{proof}

\begin{lemma}[The optimal denoiser is not affine]
\label{lem:not-affine}
If $\rho_{\mathrm{data}}$ has compact support and is not a single point
mass, then $f_t$ is not affine, for any $t \in (0,1)$.
\end{lemma}

\begin{proof}
Suppose $f_t$ is affine. By the Jacobian
identity (right of~\eqref{eq:theory:mmse}),
$\Cov(x \mid z_t = z) = (\sigma^2 / t)\, \nabla_z f_t(z)$ is then a
constant matrix $\Sigma_0$. We claim $\Sigma_0 = 0$. The posterior of
$x$ given $z_t = \lambda v$ has density proportional to
$\exp\big( (\lambda t\, v^\top x - t^2 \norm{x}^2 / 2)/\sigma^2 \big)$
with respect to $\rho_{\mathrm{data}}$. Since the support function of
the convex hull of $\operatorname{supp} \rho_{\mathrm{data}}$ is
differentiable at almost every direction, for almost every unit vector
$v$ the linear functional $x \mapsto v^\top x$ has a unique maximizer
$x^\star$ over the hull; the maximizer is then an exposed point of the
hull and so belongs to $\operatorname{supp} \rho_{\mathrm{data}}$. Fix
such a $v$. For any $\zeta > 0$, the posterior weight of
$\{x : v^\top x \leq v^\top x^\star - \zeta\}$ relative to any
neighbourhood of $x^\star$ of positive $\rho_{\mathrm{data}}$-mass
decays like $e^{-\lambda t \zeta' / \sigma^2}$ as
$\lambda \to +\infty$, so the posterior converges weakly to
$\delta_{x^\star}$, and by boundedness of the support its second
moments converge as well; hence
$\Cov(x \mid z_t = \lambda v) \to 0$ and $\Sigma_0 = 0$. Zero posterior
covariance means the posterior is a Dirac mass for every $z$. But for
every $z$ the posterior is mutually absolutely continuous with
$\rho_{\mathrm{data}}$, the Gaussian weight being strictly positive, so
it is a Dirac mass only if $\rho_{\mathrm{data}}$ is one, contradicting
the assumption.
\end{proof}

\begin{proof}[Proof of Proposition~\ref{prop:separation}(ii)]
\emph{Domain.} Propositions~\ref{prop:manifold-approx}
and~\ref{prop:separation}(i) are stated for a single patch
$z \in \mathbb{R}^D$. Attention mixes patches, and on a single token it is
linear, since its softmax weight is $1$, so part (ii) has content only at
the level of an image. We therefore take the input to be an image
$Z = (z_1, \dots, z_N) \in \mathbb{R}^{ND}$, whose tokens $x_j$ are
obtained from $\pi(z_j)$ by the affine embedding of
Section~\ref{sec:arch:tokenization}, so that the token matrix $X = X(Z)$
is an affine function of $Z$. The per-patch class is compared with this
one by applying $Q \circ \pi$ to each patch, which is a polynomial map of
$Z$ of the same degree.

\emph{Argument.} The scores $s^e_{ij}(X(Z))$ are polynomials in $Z$ and
the values $W^e_V x_j$ are affine in $Z$, so clearing the normalizers as
in the proof of Lemma~\ref{lem:affine-content} again yields an identity
$\sum_\alpha p_\alpha(Z)\, e^{\bar{s}_\alpha(Z)} \equiv 0$ with polynomial
coefficients and exponents. Lemma~\ref{lem:exp-indep} applies in the $Z$
variables, with the exponents grouped into classes modulo constants as
polynomials in $Z$. This also handles the fact that $X(Z)$ sweeps an
affine subspace of token matrices rather than an open set. Agreement of
$h \circ X$ with a polynomial map of $Z$ on a nonempty open set of images
therefore forces $h \circ X$ to be affine in $Z$. Since $f_t$ is not
affine (Lemma~\ref{lem:not-affine}) while $\mathcal{V}_{L, \Pbud}$
contains $\{Q \circ \pi : Q \in \mathcal{P}_{K_L}\}$ exactly by part (i),
the two classes meet only in affine maps.
\end{proof}

\begin{remark}
Two boundaries of the argument are worth recording. First,
Lemma~\ref{lem:affine-content} concerns a single layer; a composition
of attention layers could in principle agree with a polynomial map
without any individual layer doing so, and extending the transcendence
argument through compositions is the open depth-$L$ part of
Conjecture~\ref{conj:rate-gap}. Second, the transcendence argument
itself is specific to the softmax, since it is the exponential form of
the weights that Lemma~\ref{lem:exp-indep} exploits. This does
not, however, confine the affine-content property to softmax attention:
Proposition~\ref{prop:normalized-attention} reaches the same conclusion
for every nonnegative normalized weighting, by an argument that uses no
algebraic structure at all
(Appendix~\ref{app:proof-normalized}). Linear attention under a
nonnegative feature map is therefore covered as well, and the only
attention-shaped operator left inside the polynomial class is the
unnormalized one, for which the convexity bound fails. Pointwise MLP
nonlinearities such as GELU are likewise non-polynomial maps, and their
inclusion does not place a Transformer block inside the polynomial
class, but we do not pursue a full block-level characterization here.
The feed-forward sublayer is discussed in Appendix~\ref{app:ffn}.
\end{remark}

\noindent The quantitative form of the separation, described informally
after Proposition~\ref{prop:separation} in Section~\ref{sec:theory}, is
the following.

\begin{conjecture}[Quantitative rate gap at matched budget]
\label{conj:rate-gap}
In the setting of Proposition~\ref{prop:separation}, let
$\mathcal{A}_{L, \Pbud}$ denote the class of depth-$L$ stacks of
softmax self-attention blocks with residual connections and at most
$\Pbud$ parameters per layer. For matched depth $L$ and every
$\Pbud \geq \Pbud_0(L, \dbot)$,
\begin{equation*}
\inf_{h \,\in\, \mathcal{A}_{L, \Pbud}}\; \sup_{z \in t\manifold}\;
\norm{f_t(z) - h(z)}^2 \;\geq\; C_2 \cdot K_L^{-2s}
\cdot \big(1 + \Delta(s, L)\big)
\end{equation*}
for a constant $C_2 > 0$ and a gap $\Delta(s, L) > 0$ nondecreasing in
the smoothness order $s$.
\end{conjecture}

\noindent It is open
for a specific reason: classical rational approximation theory bounds
the \emph{reverse} direction, with free-pole rationals approximating at
least as well as polynomials of matched degree and sometimes strictly
better \citep{newman1964rational,saff1987pade}, so no generic
degree-counting argument will do. A proof must exploit the structural
constraints of attention, positive normalized weights with input-tied
denominators, for which quantitative lower bounds are not currently
available. Extending
Proposition~\ref{prop:separation}(ii) itself from a single layer to
depth-$L$ stacks is likewise open, since a composition of
non-polynomial layers can in principle be polynomial, though we expect
the same transcendence mechanism to apply.

\subsection{Affine content of normalized attention
(Proposition~\ref{prop:normalized-attention})}
\label{app:proof-normalized}

\begin{proposition}[Affine content of normalized attention]
\label{prop:normalized-attention}
Let $h$ be a mixing layer with residual connection of the form
\begin{equation}
h_i(X) \;=\; x_i \;+\; \sum_{j=1}^{N} \omega_{ij}(X)\, g(x_j),
\qquad i = 1, \dots, N,
\label{eq:theory:normalized}
\end{equation}
where $g : \mathbb{R}^{d} \to \mathbb{R}^{d}$ is affine and the weights
satisfy, at every $X \in \mathbb{R}^{N \times d}$,
\begin{equation*}
\omega_{ij}(X) \;\geq\; 0
\qquad\text{and}\qquad
\sum_{j=1}^{N} \omega_{ij}(X) \;=\; 1 .
\end{equation*}
If $h$ coincides with a polynomial map on $\mathbb{R}^{N \times d}$,
then $h$ is affine. No regularity of any kind is assumed of
$\omega$: the conclusion holds whether the weights are exponential,
rational, piecewise linear, or discontinuous.
\end{proposition}

The scope is worth stating plainly, because it covers essentially every
attention mechanism in current use. Softmax weights are nonnegative and
normalized. So are the weights of linear attention under any nonnegative
feature map, including the ReLU variant used in high-resolution
diffusion backbones \citep{katharopoulos2020transformers,xie2024sana},
kernel-feature approximations of the softmax
\citep{choromanski2021rethinking}, and any convex mixing rule whatever.
None of these lies in the polynomial class except in its affine part.
Appendix~\ref{app:extended:boundary} treats the unnormalized boundary
where the conclusion fails, and how this proposition and
Proposition~\ref{prop:separation}(ii) divide the work.

Throughout, $\norm{X}$ denotes the Frobenius norm of
$X = (x_1, \dots, x_N) \in \mathbb{R}^{N \times d}$, so that
$\norm{x_j} \le \norm{X}$ for every $j$.

\begin{proof}
Write $g(x) = Gx + c$ with $G \in \mathbb{R}^{d \times d}$ and
$c \in \mathbb{R}^{d}$, and set
\[
S_i(X) \;:=\; h_i(X) - x_i \;=\; \sum_{j=1}^{N} \omega_{ij}(X)\, g(x_j).
\]

\emph{Step 1 (the convexity bound).} Because the weights are
nonnegative and sum to one, $S_i(X)$ is a convex combination of the
points $g(x_1), \dots, g(x_N)$, hence lies in their convex hull. The
norm of a convex combination is at most the largest norm among the
points combined, so for every $X$,
\begin{equation}
\norm{S_i(X)}
\;\leq\; \max_{1 \le j \le N} \norm{g(x_j)}
\;\leq\; \norm{G}_{\mathrm{op}} \max_j \norm{x_j} + \norm{c}
\;\leq\; a \norm{X} + b,
\label{eq:app:convexity-bound}
\end{equation}
with $a := \norm{G}_{\mathrm{op}}$ and $b := \norm{c}$ depending only on
the layer's parameters. This is the only place the hypotheses on
$\omega$ are used, and they are used only through nonnegativity and
normalization; no continuity, boundedness, or algebraic form is
required.

\emph{Step 2 (affine growth forces degree one).} Suppose $h$ coincides
with a polynomial map on $\mathbb{R}^{N \times d}$. Then each $S_i$ is
polynomial, since $x_i$ is. Decompose $S_i$ into homogeneous parts,
\[
S_i \;=\; \sum_{j=0}^{J} S_i^{[j]},
\qquad S_i^{[j]}(\lambda X) = \lambda^{j} S_i^{[j]}(X),
\qquad S_i^{[J]} \not\equiv 0,
\]
and suppose for contradiction that $J \geq 2$. Fix any
$X_0 \in \mathbb{R}^{N \times d}$ and evaluate along the ray $\lambda X_0$,
$\lambda > 0$:
\[
\frac{S_i(\lambda X_0)}{\lambda^{J}}
\;=\; S_i^{[J]}(X_0) \;+\; \sum_{j<J} \lambda^{j-J} S_i^{[j]}(X_0)
\;\xrightarrow[\lambda \to \infty]{}\; S_i^{[J]}(X_0),
\]
while~\eqref{eq:app:convexity-bound} gives
\[
\frac{\norm{S_i(\lambda X_0)}}{\lambda^{J}}
\;\leq\; \frac{a\, \lambda \norm{X_0} + b}{\lambda^{J}}
\;\xrightarrow[\lambda \to \infty]{}\; 0
\qquad\text{since } J \geq 2 .
\]
Hence $S_i^{[J]}(X_0) = 0$. As $X_0$ was arbitrary,
$S_i^{[J]} \equiv 0$, contradicting the choice of $J$. Therefore
$J \leq 1$, so $S_i$ is affine and
$h_i(X) = x_i + S_i(X)$ is affine. \qedhere
\end{proof}

\begin{remark}
The argument is insensitive to how the weights are produced, which is
what gives the proposition its reach: softmax, any nonnegative feature
map with normalization (including the piecewise-linear ReLU map, for
which no analytic continuation is available), and any convex mixing rule
are all covered by the same two steps. Multiple heads are covered as
well: a sum of $n_{\mathrm{h}}$ heads, each of the
form~\eqref{eq:theory:normalized}, obeys~\eqref{eq:app:convexity-bound}
with $a$ replaced by $\sum_e \norm{G_e}_{\mathrm{op}}$, and Step~2 is
unchanged.
\end{remark}

\begin{remark}[Sharpness]
Normalization cannot be dropped. For unnormalized linear attention with
the identity feature map (Appendix~\ref{app:extended:boundary}), the
weights $x_i^\top W_Q^\top W_K x_j$ are neither nonnegative nor
normalized and grow linearly in $\norm{X}$, so~\eqref{eq:app:convexity-bound}
fails, the layer is a cubic polynomial map, and the conclusion of the
proposition is false for it.
The dichotomy is therefore exact. An attention-shaped layer with a
data-dependent denominator has affine polynomial content; one without a
denominator can be polynomial, at the cost of the convex-hull property
that the same bound expresses.
\end{remark}

\section{Architecture Details}
\label{app:arch}

This appendix gives the derivations behind
Sections~\ref{sec:arch:primitive} and~\ref{sec:arch:mixers} in full.

\paragraph{The primitive as a Volterra term.}
Each projected coordinate
$(Q^{(m)}_q u)_c = \sum_{j=1}^{n} Q^{(m)}_{q,cj}\, u_j$ is a learned mixture of
\emph{all} inputs, so multiplying the $k$ projections of a group
produces every degree-$k$ monomial, cross-terms included:
$\prod_{m} (Q^{(m)}_q u)_c
= \sum_{j_1, \dots, j_k = 1}^{n} \big(\prod_{m} Q^{(m)}_{q,c j_m}\big)\, u_{j_1}\cdots u_{j_k}$.
Summing over the $R$ groups and mixing through $P$ collects these into the
order-$k$ Volterra term~\eqref{eq:bg:volterra},
$y_i = \sum_{j_1, \dots, j_k = 1}^{n} W^{(k)}_{i,\, j_1, \dots, j_k}\, u_{j_1}\cdots u_{j_k}$,
whose kernel is supplied directly in factored form,
\begin{equation}
W^{(k)}_{i,\, j_1, \dots, j_k} \;=\; \sum_{c=1}^{h} \sum_{q=1}^{R} P_{ic}
\prod_{m=1}^{k} Q^{(m)}_{q,\, c j_m}.
\end{equation}
This is a canonical-polyadic factorization with one rank-one term per
hidden unit and group $(c, q)$, so of rank at most $hR$, with an
independent factor per slot; the factorization constrains the kernel,
not the set of interactions it represents. We parameterize the factors
independently (untied), at $k = 2$.

Three properties of the primitive matter here. First, only the symmetric
part of the kernel is identifiable, since $u_{j_1}\cdots u_{j_k}$ is
unchanged by permuting its factors; untied factors therefore cost
nothing in what is represented, while giving the optimizer a larger
space to move in. Second, evaluation costs $\mathcal{O}(k\, nhR)$ per
kernel against $\mathcal{O}(n^{k+1})$ for a dense one. Third, a single
instance realizes an order-$k$ kernel through the $k$-fold product, so
polynomial order is a per-layer choice independent of depth. We use $k = 2$ throughout, since depth
already supplies degree $2^L$ (Proposition~\ref{prop:separation}(i)).

\paragraph{Token mixer.}
The token mixer is the primitive over the $N$ token positions with
$h = N$ and $P = I$, applied to each channel column $u_{:,a}$ with the
same maps on every channel,
\begin{equation}
y_{ia} = \sum_{j=1}^{N} W^{(1)}_{\mathrm{tok},ij}\, u_{ja}
+ \sum_{q=1}^{R} \big(Q^{(1)}_q u_{:,a}\big)_i \big(Q^{(2)}_q u_{:,a}\big)_i,
\qquad Q^{(1)}_q, Q^{(2)}_q \in \mathbb{R}^{N \times N},
\label{eq:token-factor}
\end{equation}
so that $W^{(2)}_{\mathrm{tok},ijj'} = \sum_{q=1}^{R} Q^{(1)}_{q,ij}\,
Q^{(2)}_{q,ij'}$, and for each output position $i$ the $N \times N$
slice of the kernel has rank at most $R$. All $2R + 1$ maps are dense
$N \times N$ matrices, so the mixer holds $(2R+1)N^2$ parameters per
block, independent of the width $d$, against $N^3$ for a dense
$W^{(2)}_{\mathrm{tok}}$ alone. The maps carry no channel index, so in
implementation all $2R$ quadratic maps are applied to the $N \times d$
token matrix at once and their outputs multiplied elementwise.
Evaluation costs $(2R+1)N^2 d$ multiply-adds for the maps and about
$2RNd$ for the products and their sum, so $\mathcal{O}(RN^2 d)$ in
total. Self-attention costs $\mathcal{O}(N^2 d + N d^2)$, the second
term for its four $d \times d$ projections, which the token mixer does
not have.

\paragraph{Channel mixer.}
The channel mixer is the primitive over the $d$ channels of one token,
with hidden width $h$, $R'$ pairs $Q^{(1)}_q, Q^{(2)}_q \in
\mathbb{R}^{h \times d}$ and $P \in \mathbb{R}^{d \times h}$, applied at
every position with the same weights. Its linear branch also passes
through the hidden layer, $W^{(1)}_{\mathrm{ch}} = P Q^{(0)}$ with
$Q^{(0)} \in \mathbb{R}^{h \times d}$, so the mixer holds $(2R'+2)dh$
parameters and costs as many multiply-adds per token. At $h = 2d$ and
$R' = 1$ (Section~\ref{sec:exp:setup}) this is $8d^2$, the parameter
count of the gated feed-forward block of \citet{li2026jit} at the same
width.

\section{Implementation Details}
\label{app:implementation}
\label{app:algorithms}

Algorithms~\ref{alg:train} and~\ref{alg:sample} give the training step
and the sampler. They follow Algorithms~1 and~2 of \citet{li2026jit},
since \ourmodel{} changes only the network $\mathrm{net}_\theta$, and use
the settings of Section~\ref{sec:exp:setup}. Images are scaled to
$[-1, 1]$, and $\varnothing$ denotes the null class label used for
classifier-free guidance \citep{ho2021cfg}.

\begin{algorithm}[!ht]
\caption{Training step.}
\label{alg:train}
\begin{algorithmic}[1]
\State \textbf{Input:} clean image $x$, class label $c$, network $\mathrm{net}_\theta$
\State with probability $0.1$, set $c \gets \varnothing$ \Comment{label dropout}
\State $t \sim \mathrm{logit\text{-}normal}(-0.8,\, 0.8)$ \Comment{location $-0.8$, scale $0.8$}
\State $\epsilon \sim \mathcal{N}(0, I)$
\State $z_t \gets t\, x + (1-t)\, \epsilon$
\State $\hat{x}_\theta \gets \mathrm{net}_\theta(z_t, t, c)$ \Comment{$x$-prediction}
\State $v \gets (x - z_t) / \max(1 - t,\, 0.05)$ \Comment{$= x - \epsilon$ for $t \le 0.95$}
\State $v_\theta \gets (\hat{x}_\theta - z_t) / \max(1 - t,\, 0.05)$
\State $\mathcal{L} \gets \norm{v_\theta - v}^2$
\State take a gradient step on $\mathcal{L}$ and update the EMA weights
\end{algorithmic}
\end{algorithm}

\begin{algorithm}[!ht]
\caption{Sampling with classifier-free guidance, Heun solver.}
\label{alg:sample}
\begin{algorithmic}[1]
\State \textbf{Input:} class label $c$, guidance scale $\omega$, guidance interval $(t_{\mathrm{lo}}, t_{\mathrm{hi}})$, number of steps $S = 50$; $\mathrm{net}_\theta$ uses the EMA weights
\Statex \textbf{Guided velocity} $\bar{v}(z, t)$:
\State \quad $v_c \gets (\mathrm{net}_\theta(z, t, c) - z) / \max(1 - t,\, 0.05)$
\State \quad $v_\varnothing \gets (\mathrm{net}_\theta(z, t, \varnothing) - z) / \max(1 - t,\, 0.05)$
\State \quad $w \gets \omega$ if $t_{\mathrm{lo}} < t < t_{\mathrm{hi}}$, else $w \gets 1$ \Comment{guidance interval}
\State \quad $\bar{v}(z, t) \gets v_\varnothing + w\,(v_c - v_\varnothing)$
\Statex \textbf{Integration} from $t = 0$ (noise) to $t = 1$ (data):
\State $z \sim \mathcal{N}(0, I)$, \ $t_i \gets i/S$ for $i = 0, \dots, S$ \Comment{linear time grid}
\For{$i = 0, \dots, S-2$}
\State $h \gets t_{i+1} - t_i$
\State $k_1 \gets \bar{v}(z, t_i)$, \ $k_2 \gets \bar{v}(z + h\,k_1,\, t_{i+1})$
\State $z \gets z + \tfrac{h}{2}\,(k_1 + k_2)$ \Comment{Heun step}
\EndFor
\State $z \gets z + (t_S - t_{S-1})\, \bar{v}(z, t_{S-1})$ \Comment{final step in Euler form}
\State \Return $z$
\end{algorithmic}
\end{algorithm}

\section{Results at \texorpdfstring{$512\times512$}{512 × 512}}
\label{app:results512}

Table~\ref{tab:main512} reports ImageNet $512\times512$ with patch size
$32$, so $N = 256$ as at $256\times256$ and the recipe is otherwise
unchanged. The patch dimension is now $\Dp = 3072$ and the bottleneck
compresses each patch by a much larger factor, which puts the manifold
argument of Section~\ref{sec:theory} under the most pressure. The
matched-scale ordering against JiT survives. \ourmodel{} lowers FID at
every scale at matched parameters and FLOPs, from $4.02$
to $3.73$ at B, $2.53$ to $2.31$ at L, $1.94$ to $1.75$ at H and $1.78$
to $1.55$ at G. These margins, $0.19$ to $0.29$, are close to those at
$256\times256$ ($0.22$ to $0.34$), but the largest gain is now at B
rather than G. Inception Score is higher than JiT at B and L and
slightly lower at H and G. Among pixel-space models without pretrained
components, only SiD2-UViT/2 reaches a lower FID ($1.48$), at $653$
GFLOPs against our $320$, and \ourmodel-H/32 ($1.75$ at $165$ GFLOPs)
improves on SiD2-UViT/4 ($2.19$ at $137$ GFLOPs).

\begin{table}[t]
\centering
\small
\setlength{\tabcolsep}{4pt}
\caption{Reference results on ImageNet $512\times512$. Notation as in
Table~\ref{tab:main256}. At this resolution the patch dimension is
$\Dp = 3072$, exceeding the hidden width of every model listed, so the
bottleneck embedding is compressing by a larger factor than at $256$.
Baseline numbers are as reported by \citet{li2026jit}.}
\label{tab:main512}
\resizebox{\textwidth}{!}{%
\begin{tabular}{lcccrrcc}
\toprule
& \multicolumn{3}{c}{pre-training} & & & & \\
\cmidrule(lr){2-4}
Method & token & perc. & self-sup. & params & GFLOPs & FID $\downarrow$ & IS $\uparrow$ \\
\midrule
\multicolumn{8}{l}{\emph{Latent-space diffusion}} \\
DiT-XL/2 \citep{peebles2023dit}    & SD-VAE & VGG & --     & 675+49M  & 525 & 3.04 & 240.8 \\
SiT-XL/2 \citep{ma2024sit}         & SD-VAE & VGG & --     & 675+49M  & 525 & 2.62 & 252.2 \\
REPA, SiT-XL/2 \citep{yu2025repa}  & SD-VAE & VGG & DINOv2 & 675+49M  & 525 & 2.08 & 274.6 \\
DDT-XL/2 \citep{wang2025ddt}       & SD-VAE & VGG & DINOv2 & 675+49M  & 525 & 1.28 & 305.1 \\
RAE, DiT$^{\text{DH}}$-XL/2 \citep{zheng2025rae} & RAE & VGG & DINOv2 & 839+415M & 642 & 1.13 & 259.6 \\
\midrule
\multicolumn{8}{l}{\emph{Pixel-space diffusion}} \\
ADM-G \citep{dhariwal2021adm}            & -- & -- & --     & 559M & 1983 & 7.72 & 172.7 \\
RIN \citep{jabri2023rin}                 & -- & -- & --     & 320M & 415  & 3.95 & 216.0 \\
SiD, UViT/4 \citep{hoogeboom2023simplediff} & -- & -- & --  & 2B   & 555  & 3.02 & 248.7 \\
VDM++, UViT/4 \citep{kingma2023vdmpp}    & -- & -- & --     & 2B   & 555  & 2.65 & 278.1 \\
SiD2, UViT/4 \citep{hoogeboom2025sid2}   & -- & -- & --     & N/A  & 137  & 2.19 & --    \\
SiD2, UViT/2 \citep{hoogeboom2025sid2}   & -- & -- & --     & N/A  & 653  & 1.48 & --    \\
PixNerd, XL/16 \citep{wang2025pixnerd}   & -- & -- & DINOv2 & 700M & 583  & 2.84 & 245.6 \\
JiT-B/32 \citep{li2026jit}               & -- & -- & --     & 133M & 26   & 4.02 & 271.0 \\
JiT-L/32 \citep{li2026jit}               & -- & -- & --     & 462M & 89   & 2.53 & 299.9 \\
JiT-H/32 \citep{li2026jit}               & -- & -- & --     & 956M & 183  & 1.94 & 309.1 \\
JiT-G/32 \citep{li2026jit}               & -- & -- & --     & 2B   & 384  & 1.78 & 306.8 \\
\midrule
\multicolumn{8}{l}{\emph{Pixel-space diffusion, polynomial (ours)}} \\
\ourmodel-B/32 & -- & -- & -- & 121M & 28  & $3.73$ & $279.1$ \\
\ourmodel-L/32 & -- & -- & -- & 394M & 87  & $2.31$ & $302.4$ \\
\ourmodel-H/32 & -- & -- & -- & 790M & 165 & $1.75$ & $307.3$ \\
\ourmodel-G/32 & -- & -- & -- & 1.6B & 320 & $1.55$ & $303.9$ \\
\bottomrule
\end{tabular}}

\end{table}

\clearpage
\section{Additional Qualitative Samples}
\label{app:samples}

\begin{figure}[p]
\centering
\includegraphics[width=\textwidth]{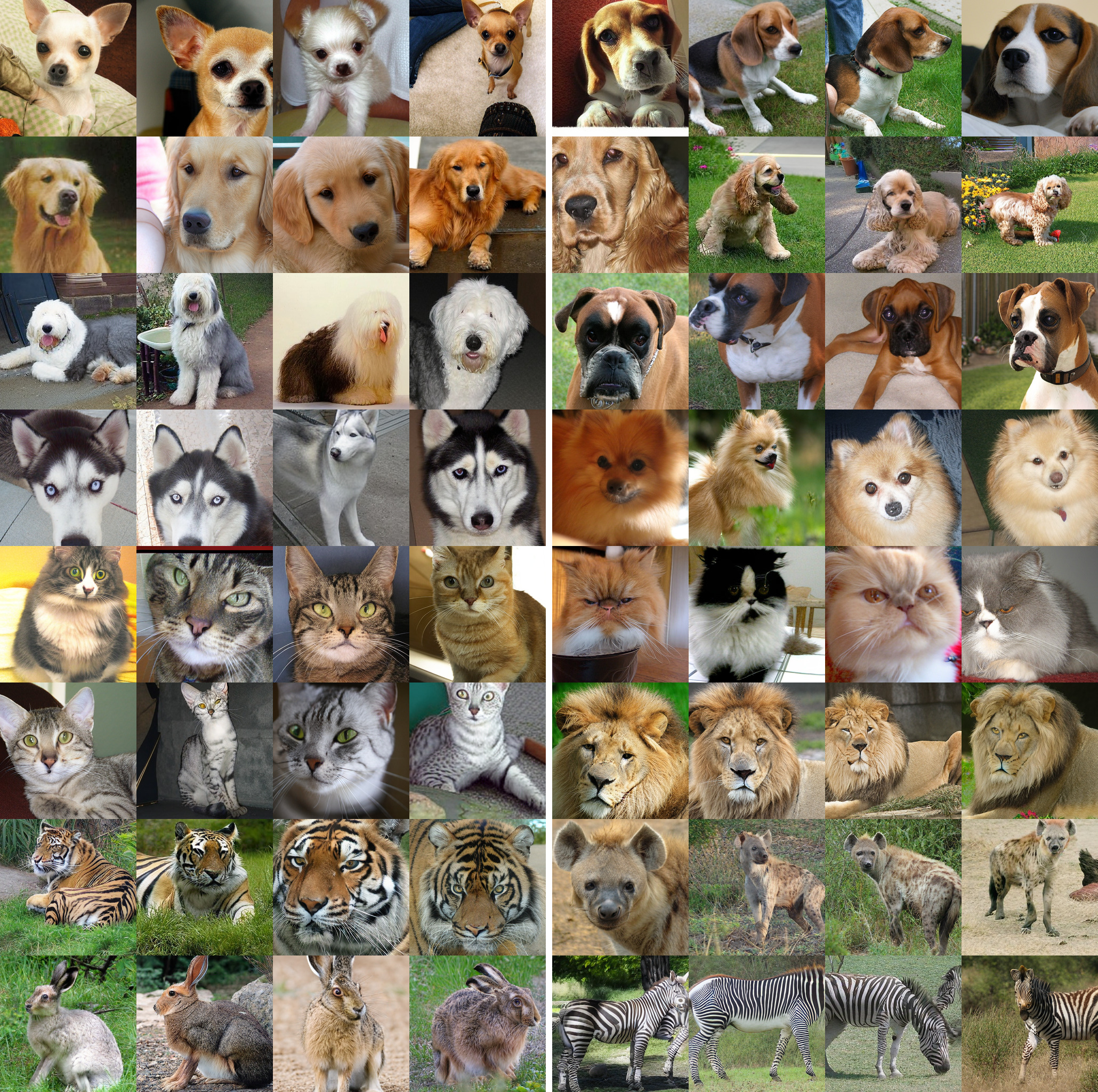}
\caption{\textbf{Uncurated samples from \ourmodel-G/16 on ImageNet $256\times256$, mammals.}
Each row shows two classes, four samples each, with the ImageNet class
index in parentheses. From the top, the rows show Chihuahua (151) and beagle (162); golden retriever (207) and cocker spaniel (219); Old English sheepdog (229) and boxer (242); Siberian husky (250) and Pomeranian (259); tabby (281) and Persian cat (283); Egyptian cat (285) and lion (291); tiger (292) and hyena (276); hare (331) and zebra (340).}
\label{fig:samples-app1}
\end{figure}

\begin{figure}[p]
\centering
\includegraphics[width=\textwidth]{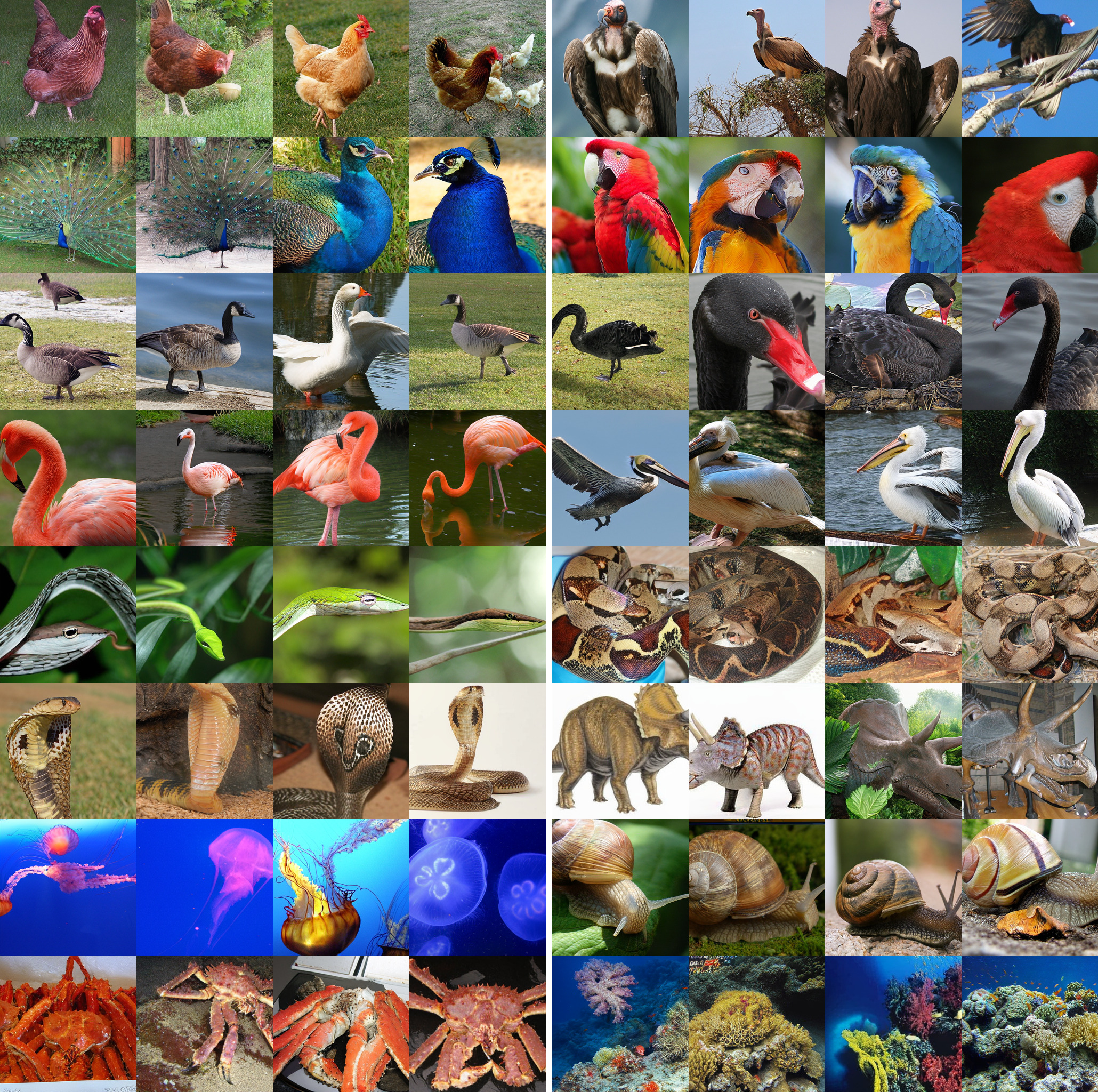}
\caption{\textbf{Uncurated samples from \ourmodel-G/16 on ImageNet $256\times256$, birds, reptiles and other animals.}
Each row shows two classes, four samples each, with the ImageNet class
index in parentheses. From the top, the rows show hen (8) and vulture (23); peacock (84) and macaw (88); goose (99) and black swan (100); flamingo (130) and pelican (144); vine snake (59) and boa constrictor (61); Indian cobra (63) and triceratops (51); jellyfish (107) and snail (113); king crab (121) and coral reef (973).}
\label{fig:samples-app2}
\end{figure}

\begin{figure}[p]
\centering
\includegraphics[width=\textwidth]{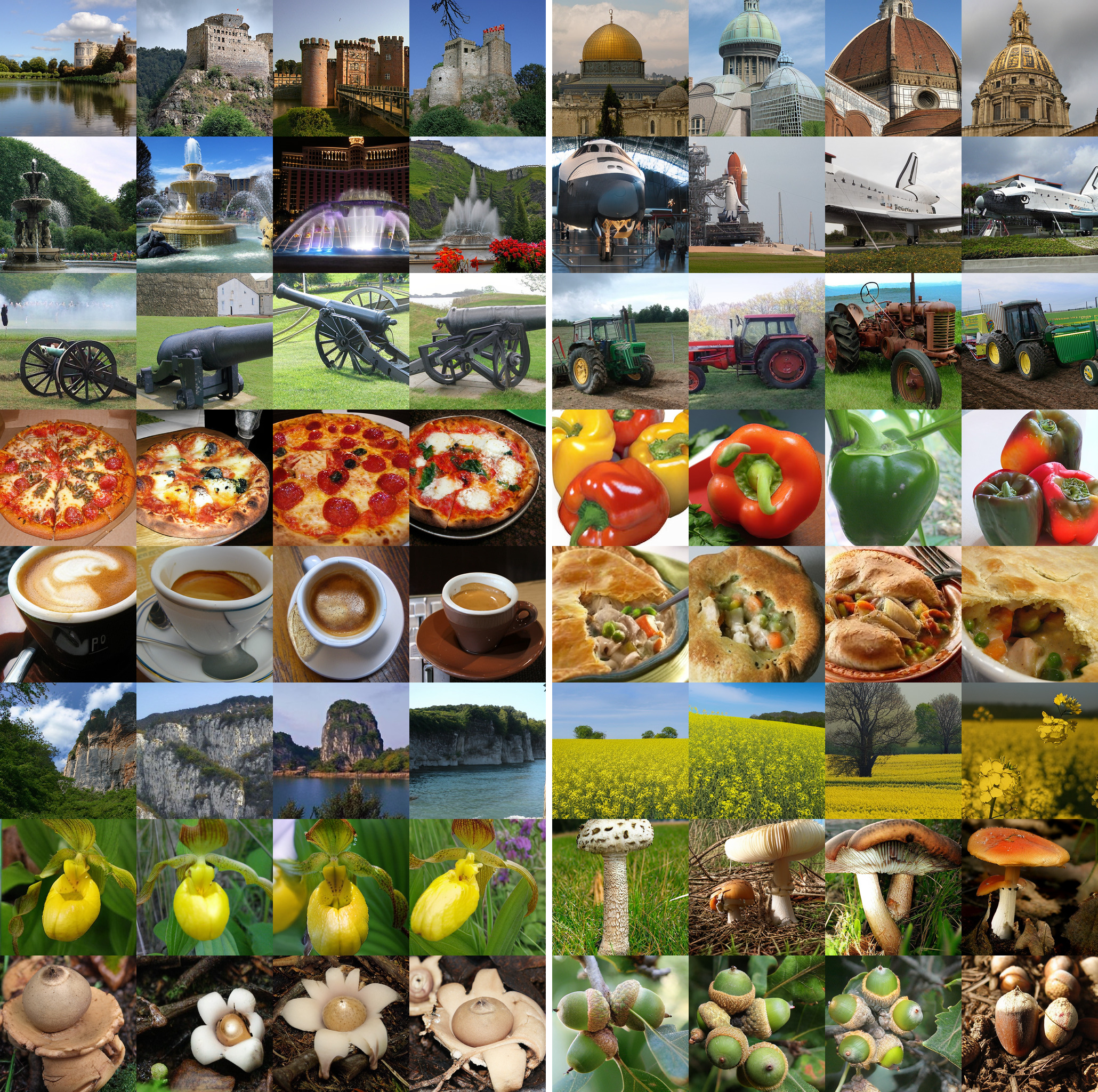}
\caption{\textbf{Uncurated samples from \ourmodel-G/16 on ImageNet $256\times256$, objects, food and nature.}
Each row shows two classes, four samples each, with the ImageNet class
index in parentheses. From the top, the rows show castle (483) and dome (538); fountain (562) and space shuttle (812); cannon (471) and tractor (866); pizza (963) and bell pepper (945); espresso (967) and potpie (964); cliff (972) and rapeseed (984); yellow lady's slipper (986) and mushroom (947); earthstar (995) and acorn (988).}
\label{fig:samples-app3}
\end{figure}

\begin{figure}[p]
\centering
\includegraphics[width=\textwidth]{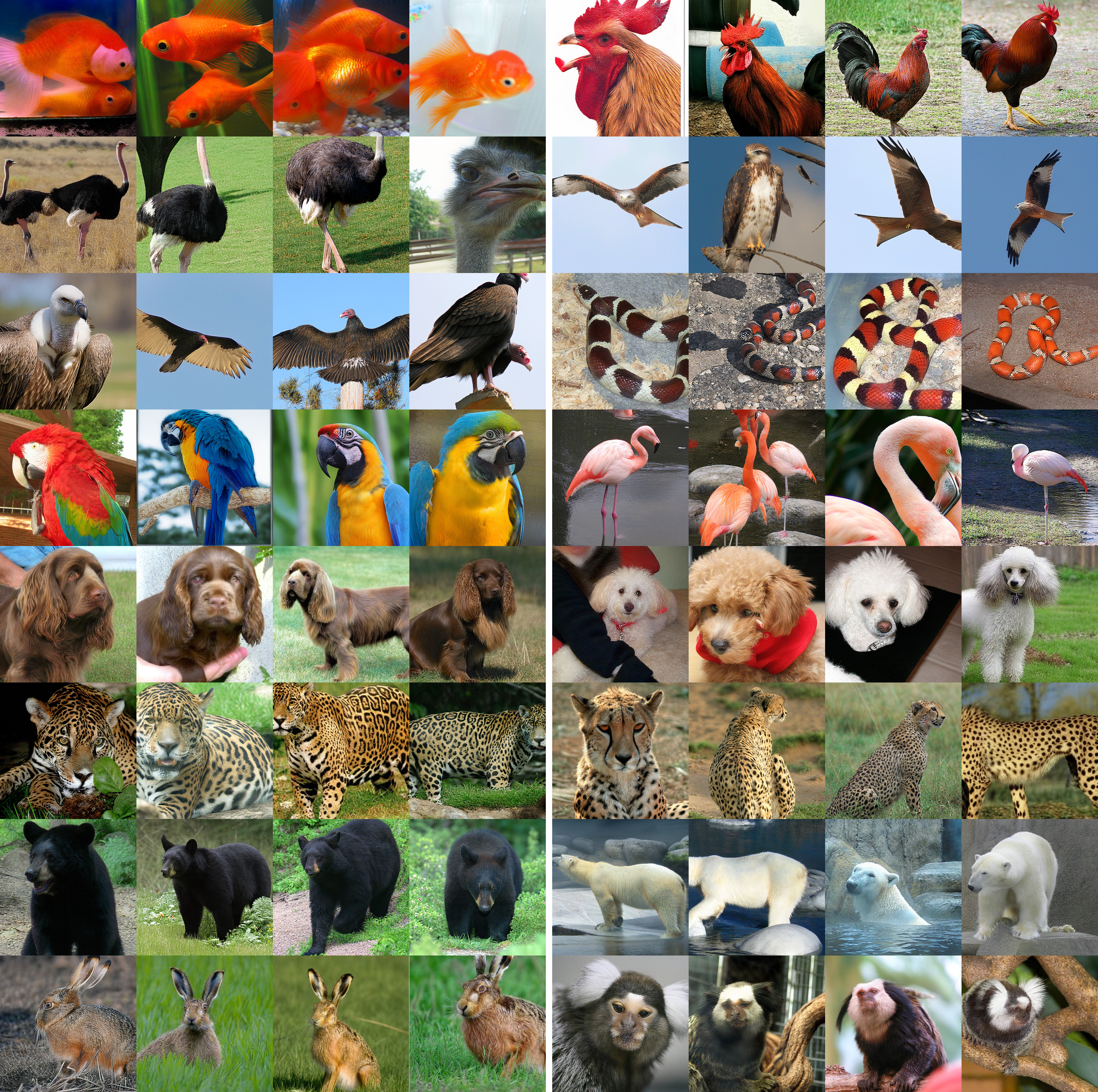}
\caption{\textbf{Uncurated samples from \ourmodel-G/32 on ImageNet $512\times512$, animals.}
Each row shows two classes, four samples each, with the ImageNet class
index in parentheses. From the top, the rows show goldfish (1) and cock (7); ostrich (9) and kite (21); vulture (23) and king snake (56); macaw (88) and flamingo (130); Sussex spaniel (220) and miniature poodle (266); jaguar (290) and cheetah (293); American black bear (295) and ice bear (296); hare (331) and marmoset (377).}
\label{fig:samples512-app1}
\end{figure}

\begin{figure}[p]
\centering
\includegraphics[width=\textwidth]{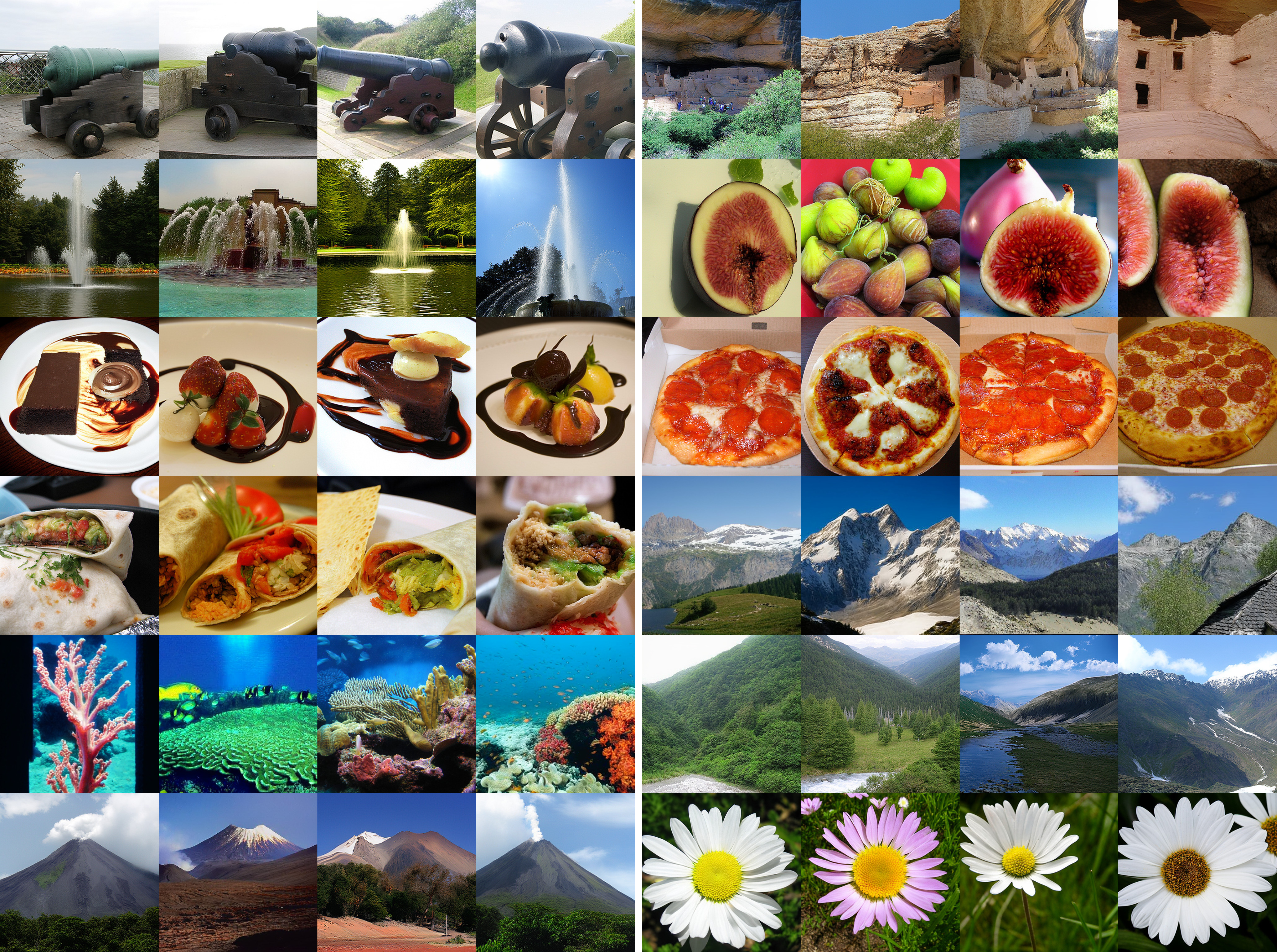}
\caption{\textbf{Uncurated samples from \ourmodel-G/32 on ImageNet $512\times512$, objects, food and scenes.}
Each row shows two classes, four samples each, with the ImageNet class
index in parentheses. From the top, the rows show cannon (471) and cliff dwelling (500); fountain (562) and fig (952); chocolate sauce (960) and pizza (963); burrito (965) and alp (970); coral reef (973) and valley (979); volcano (980) and daisy (985).}
\label{fig:samples512-app2}
\end{figure}

Figures~\ref{fig:samples-app1}--\ref{fig:samples-app3} show
class-conditional samples from \ourmodel-G/16 on ImageNet
$256\times256$, generated with classifier-free guidance and the 50-step
Heun sampler of Appendix~\ref{app:algorithms}. We drew 16 samples for
each of 56 classes and selected four per class for 48 of them, so the
grids are curated rather than random.
Figures~\ref{fig:samples512-app1} and~\ref{fig:samples512-app2} show
samples from \ourmodel-G/32 on ImageNet $512\times512$ with the same
sampler, selected in the same way (four of 16 samples per class, for 28
of 30 classes). The $512\times512$ samples in Figure~\ref{fig:samples}
are drawn from the same pool.

\end{document}